\documentclass{article}

\usepackage[eandd,final]{neurips_2026}

\usepackage[utf8]{inputenc} 
\usepackage[T1]{fontenc}    
\usepackage{hyperref}       
\usepackage{url}            
\usepackage{booktabs}       
\usepackage{amsfonts}       
\usepackage{nicefrac}       
\usepackage{microtype}      
\usepackage{xcolor}         

\usepackage[table]{xcolor}
\usepackage[utf8]{inputenc} 
\usepackage[T1]{fontenc}    
\usepackage{hyperref}       
\usepackage{url}   
\usepackage{kotex} 
\usepackage{booktabs}       
\usepackage{amsfonts}       
\usepackage{nicefrac}       
\usepackage{microtype}     
\usepackage[dvipsnames]{xcolor}         
\usepackage{subcaption}
\usepackage{float} 
\usepackage{booktabs}
\usepackage{multirow}
\usepackage{graphicx}
\usepackage{tabularx}
\usepackage{longtable}
\usepackage[export]{adjustbox}
\usepackage{graphicx}
\usepackage{longtable}
\usepackage{wrapfig}
\usepackage{mdframed}
\usepackage{xspace}
\usepackage[most]{tcolorbox}
\usepackage{keystroke}
\usepackage{menukeys}
\usepackage[shortlabels]{enumitem}
\usepackage{adjustbox}
\usepackage{amssymb}
\usepackage{pifont}

\newcommand{\framework}{{\textsc{BenchHub}}\xspace}

\def\eg{\emph{e.g}.,\xspace}
\def\ie{\emph{i.e}.,\xspace}
\newtheorem{observation}{Observation}
\title{Benchmarks Are Not Atomic: Composition-Aware LLM Evaluation using \framework}

\author{%
Eunsu Kim$^{1}$\thanks{Equal contribution.}\ \quad
Haneul Yoo$^{2}$\footnotemark[1]\ \quad
  Guijin Son$^{3,4}$ \quad 
  Hitesh Patel$^5$ \quad 
  Amit Agarwal$^5$ \quad 
  Alice Oh$^6$ \\
  $^1$Carnegie Mellon University \quad $^2$New York University \quad \\$^3$Seoul National University \quad $^4$OnelineAI \quad $^5$Oracle \quad $^6$KAIST
\\\texttt{eunsukim@andrew.cmu.edu \quad haneul.yoo@nyu.edu \quad alice.oh@kaist.edu} \\ 
}

\begin{document}

\maketitle

\addtocounter{footnote}{-1}
\begin{abstract}
LLM benchmarks are often treated as coherent measurement units, yet they are heterogeneous collections of instances spanning diverse domains, skills, formats, and contexts. As a result, aggregate benchmark scores can conflate model capability with benchmark composition, obscuring what existing benchmarks actually cover.
We introduce \framework{}, a composition-aware evaluation framework that represents benchmarks as distributions over instance-level attributes. \framework{} integrates 54 benchmarks comprising 839K samples across 10 languages. It enables researchers to inspect benchmark contents, uncover reusable coverage hidden behind benchmark names, transparently compare benchmarks, and construct controllable evaluation sets for application-aligned model selection.
Using \framework{}, we show that similarly motivated benchmarks can differ substantially in internal composition, existing benchmarks often contain reusable coverage beyond their stated purposes, models exhibit fine-grained category-level performance variation hidden by aggregate scores, and model rankings can shift under different reweighting and resampling configurations. Our results motivate evaluation practices that make benchmark composition explicit, inspectable, and controllable. 

\noindent\centering
\small
\begin{tabular}{@{}c l l@{}}
  \includegraphics[height=12pt,valign=c]{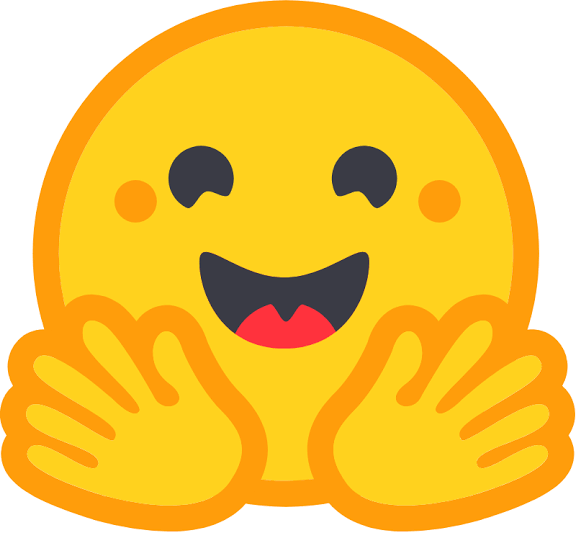} &
  \texttt{Dataset \& Website} &
  \href{https://huggingface.co/BenchHub}{\texttt{https://huggingface.co/BenchHub}} \\
  
  \includegraphics[height=12pt,valign=c]{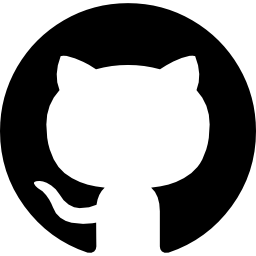} &
  \texttt{Code} &
  \href{https://github.com/rladmstn1714/BenchHub}{\texttt{https://github.com/rladmstn1714/BenchHub}} \\
\end{tabular}

\end{abstract}

\section{Introduction}
\label{sec:intro}
The rapid development and deployment of large language models (LLMs) have led to the emergence of benchmark datasets at an unprecedented rate.~\cite{hendrycks2021measuring, srivastava2023beyond, liang2023holistic}.
\footnote{For instance, NeurIPS Datasets and Benchmarks Track submissions increased from 447 in 2022 to 1,995 in 2025, corresponding to a 4.5$\times$ increase, which is nearly twice the growth rate of the main track~\citep{neurips2022factsheet,neurips2025dbtrack}.} 
Benchmark datasets serve as a primary instrument in assessing what LLMs can and cannot do and in shaping what capabilities the community expects LLMs to demonstrate~\cite{wang2024mmlupro, sokol2025benchmarkcards}. 
Accordingly, leaderboards and technical reports continually evaluate models on newly emerging benchmarks and aggregate their scores into compact indicators of model capability.

However, benchmark-level aggregate scores provide an incomplete, and sometimes misleading, view of evaluation~\cite{ribeiro-etal-2020-beyond, reuel2024betterbench}.
Benchmarks consist of heterogeneous instances that differ substantially in what they require, what they cover, and how they are presented~\cite{kim-etal-2025-benchmark, kim-etal-2025-biggen}.
Aggregating these instances into a single score can obscure these differences, making it harder to understand what aspects of model capability the score represents and which regions of the evaluation space remain underexplored.
It also limits robust model analysis: two models with identical aggregate scores may exhibit entirely disjoint failure modes, and comparative conclusions can flip when an evaluation set is subtly resampled, or swapped for a functionally similar benchmark~\cite{alzahrani-etal-2024-benchmarks, singh2025leaderboard, frick2025prompt}.
Thus, benchmarks should not be treated as atomic units; their internal composition is inherently part of the evaluation. 
What the community needs now is not merely more datasets, but better methodologies to understand and use the ones we already have.

\begin{figure}[t!]
    \centering
    \includegraphics[width=1\textwidth]{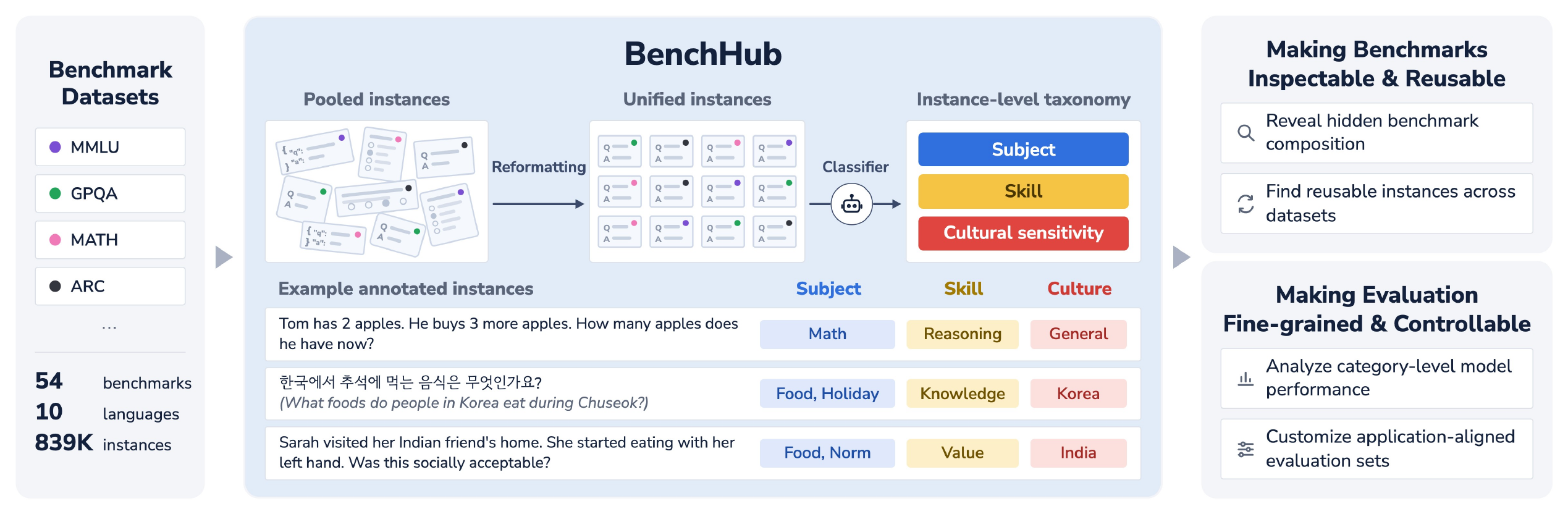}
\caption{
Conceptual overview of \framework.
\framework automatically merges existing benchmark datasets and annotates questions at the instance level with a unified taxonomy, making the contents of existing benchmarks explicit and reusable beyond their dataset names.
This enables users to construct objective-aligned test sets and conduct fine-grained, controllable model evaluation.
}
\label{fig:teaser}
\end{figure}
To make current benchmarks inspectable, reusable, and controllable, we introduce \framework{}, a dynamic infrastructure for transforming existing benchmarks into composition-aware evaluation resources. 
\framework{} integrates 54 benchmark datasets comprising 839K instances across 10 languages. 
Each instance is annotated with fine-grained attributes, including subject domain, required skill, input--output format, cultural specificity, and tool-use requirements, using an automated expansion pipeline backed by carefully designed taxonomies.
Unlike prior work on instance-level benchmark annotation~\citep{ye2024flask, zhou2025generalscalesunlockai}, which typically operates over fixed evaluation suites, \framework{} provides a unified instance-level representation that continuously integrates and reorganizes existing benchmark collections into searchable, analyzable, and reconfigurable evaluation data. This representation enables researchers to inspect evaluation composition, analyze attribute coverage, interpret model performance at a finer level, and construct or select custom evaluation sets aligned with specific operational goals.

Using \framework, we demonstrate the critical value of composition-aware evaluation in two areas.
\begin{itemize}[leftmargin=*]
    \item \textbf{Making benchmark datasets inspectable and reusable (\S\ref{sec:sq1}):} 
    We systematically reveal that existing benchmarks are internally heterogeneous and rarely compositionally interchangeable, even when designed for similar purposes. \framework exposes uneven coverage across attributes, and identifies reusable evaluation instances that remain hidden when practitioners select benchmarks by name alone.
    \item \textbf{Making model evaluation fine-grained and controllable (\S\ref{sec:sq2}):} 
    We empirically demonstrate that models exhibit structured performance disparities across composition dimensions that are entirely obscured by aggregate scores. \framework enables fine-grained, category-level analysis and shows how model rankings and deployment decisions can shift under plausible reweighting and resampling configurations.
\end{itemize}

Rather than diminishing the value of existing aggregate scores, which remain vital for high-level progress tracking, our findings complement them by offering a higher-resolution lens.
We argue that evaluation practice should more explicitly account for composition: which instances are included, how they are distributed, and how sensitive conclusions are to these distributions. 
\framework bridges this gap, enabling rigorous benchmark auditing, revealing underrepresented evaluation space, and paving the way for targeted, robust model assessment.

\section{\framework: Unified \& Composition-aware LLM Evaluation Framework}
\label{sec:benchhub}
\framework is a composition-aware evaluation framework that organizes existing LLM benchmarks into a shared instance-level representation. 
It integrates 54 benchmarks comprising 839k samples across 10 languages, with a primary focus on English and Korean through \framework-En and \framework-Ko, respectively. 
Rather than treating benchmarks as indivisible evaluation units, \framework represents them as distributions over compositional attributes, making benchmark composition explicit, inspectable, and controllable.

We design \framework around two core principles. 
First, it provides a fine-grained, multi-dimensional taxonomy for describing the composition of evaluation instances. 
Second, it uses an automated expansion pipeline to reformat, annotate, and integrate newly released datasets into the unified benchmark space. 
In this section, we describe the taxonomy design (\S\ref{sec:method-taxonomy}), the contained dataset (\S\ref{sec:method-Datasets}), and the automated pipeline (\S\ref{sec:method-implement}).

To improve usability, \framework also provides a web-based platform with interactive tools and code utilities for data selection, customized evaluation-set construction, multilingual extensions, and automated model evaluation.
Implementation details are provided in Appendix~\ref{appendix:benchhub_web}.

\subsection{Taxonomy}
\label{sec:method-taxonomy}
We annotate benchmarks using six complementary dimensions designed to capture the major factors that shape benchmark composition and evaluation behavior. 
These dimensions characterize \textit{what} capability is evaluated, \textit{how} responses are produced and assessed, and \textit{which} knowledge or cultural context is required. 
Specifically, we define three dataset-level attributes---\textbf{task, answer format, and tool usage}---and three sample-level attributes---\textbf{skill, subject, and cultural-specificity}. 
Together, these dimensions define the composition space used throughout \framework to inspect benchmark contents, compare evaluation mixtures, and construct targeted evaluation sets. 
The full taxonomy is illustrated in Appendix~\ref{appendix:method-taxonomy}.

\textbf{Dataset-level attributes:}
\begin{enumerate}[leftmargin=*, itemsep=1pt, topsep=0pt, parsep=0pt]
    \item \textbf{Task} refers to the high-level family defined by the dataset authors (\eg mathematical reasoning, code generation, cultural understanding). This provides a general understanding of a dataset's purpose. We assign it automatically from the dataset’s abstract or description using LLM inference.
    \item \textbf{Answer format} specifies the expected response format: binary, multiple-choice QA (MCQA), short-form, free-form, and comparison~(\eg determining which response is better between A and B). This is crucial for selecting appropriate evaluation prompts and formats.
    \item \textbf{Tool usage} indicates whether a task requires language capabilities only (\emph{language-only}) or interaction with external tools, such as code interpreters, web browsers, and calculators (\emph{requires external tools}). This dimension supports agentic evaluation, where models must decide when and how to invoke external resources.
\end{enumerate}

\textbf{Instance-level attributes:}
\begin{enumerate}[leftmargin=*, itemsep=1pt, topsep=0pt, parsep=0pt]
\setcounter{enumi}{3}
  \item \textbf{Subject} denotes the knowledge domain. We define six coarse-grained categories---\textit{Science}, \textit{Technology}, \textit{Humanities and Social Science (HASS)}, \textit{Art \& Sports}, \textit{Culture}, and \textit{Social Intelligence}---along with 64 sub-categories, by integrating various knowledge classification systems (Appendix~\ref{appendix:method-taxonomy}). Each sample may have multiple subject labels. 
    \item \textbf{Skill} captures the required ability to answer the question (\ie knowledge, reasoning, and value/alignment). These categories reflect widely used capability abstractions in LLM evaluation settings.
  
    \item \textbf{Cultural-specificity} represents the cultural or geographical focus. Culturally agnostic items are labeled as \emph{General}; otherwise, we assign a \emph{Local} tag. This dimension is important because model responses and their appropriateness may vary substantially across cultural contexts~\citep{singh-etal-2025-global}.
\end{enumerate}

\subsection{Datasets}
\label{sec:method-Datasets}

\begin{figure}[htb!]
    \centering

    \includegraphics[width=\textwidth]{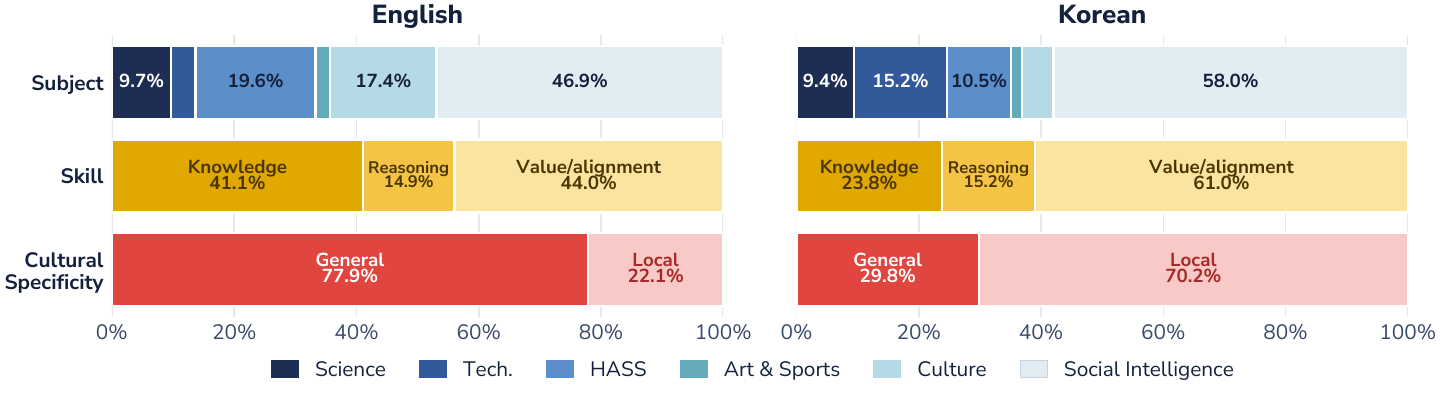}
 \label{fig:stacked_barh_data_all_en}

    \caption{Data distribution of all datasets used in \framework by coarse-grained subjects, targets, and tasks. The English and Korean data include 183k and 137k questions, respectively.}
    \label{fig:stacked_barh_data_dist}

\end{figure}

We apply this taxonomy to 54 benchmarks in 10 languages, mainly covering English and Korean and totaling over 839k instances.
Figures \ref{fig:stacked_barh_data_dist} show the overall statistics of English and Korean datasets included in our benchmark. For English and Korean, we include 31 English and 12 Korean language benchmarks with a total of 41 datasets.~\footnote{We count the multilingual datasets---BLEnD~\citep{myung2024blend} and CaLMQA~\citep{arora-etal-2025-calmqa}---in both.}
We curate (1) general-purpose (\ie culturally agnostic) datasets commonly used by holistic evaluation benchmarks~\citep{ye2024flask, ni2024mixeval} and (2) culture-specific datasets.
We select English datasets spanning multiple cultures drawn from a recent survey~\citep{pawar-etal-2024-generate}, curating over 300 papers and datasets regarding LLM cultural awareness. For Korean, we include most datasets released after 2022. Tables~\ref{tab:benchmarks_stats}--\ref{tab:benchmarks_stats_multilingual} in Appendix provide a complete list of the datasets.

\subsection{Automated and Dynamic Expansion} 
\label{sec:method-implement}
Since benchmark datasets are emerging at a rapid pace, \framework is designed to support continuous expansion.
To integrate newly introduced datasets, we automate the dataset integration process using an LLM-assisted pipeline that reformats datasets into the \framework schema and assigns taxonomy labels to each instance.
The pipeline consists of four steps:
\begin{enumerate}[leftmargin=*, itemsep=1pt, topsep=1pt, parsep=0pt]
    \item \textbf{Reformatting:} We first automatically parse, reformat, and map a new dataset to the standardized \framework scheme using an LLM-guided rule-based approach. If the dataset does not adhere to our predefined schema, an LLM agent (\eg GPT-4o or Gemini) is employed to map keys to the correct format.
    \item \textbf{Metadata assignment:} The LLM agent extracts the meta-task description and infers the task, answer format, and tool usage from the dataset documentation (\eg abstract).
    \item \textbf{Sample-level categorization:} We then assign sample-level attributes (\ie skill, subject, and cultural-specificity) using a fine-tuned Qwen-2.5-7B-instruct model (\texttt{BenchHub-Cat-7B}).\thinspace\footnote{Details on the training and validation of \texttt{BenchHub-Cat-7B} are provided in Appendix~\ref{appendix:categorizer_mono}. \texttt{BenchHub-Cat-7B} is publicly available via huggingface: \href{https://huggingface.co/BenchHub/BenchHub-Cat-7b}{BenchHub/BenchHub-Cat-7b}.} Table~\ref{tab:cat_acc} in Appendix reports the accuracy of the categorizer model for sample-level categorization. 
    \item \textbf{Merging:} The processed and annotated dataset is seamlessly merged into the main collections, thereby producing the next \framework release.
\end{enumerate}

This automated pipeline enables \framework{} to continuously expand as new datasets emerge, supporting more comprehensive evaluations and analyses over time. 
Although LLM-based expansion is inherently imperfect, we empirically validate the categorization process: on human/metadata-aligned validation sets, \texttt{BenchHub-Cat-7B} achieves 0.871 accuracy for subject, 0.967 for skill, and 0.986 for cultural-specificity. 
We further examine the reliability, scalability, and robustness of the expansion process in Appendix~\ref{appendix:implementation}. 

\textbf{Responsible reuse.}
Because \framework{} aggregates existing benchmark datasets, it may inherit limitations or constraints from the original source datasets, such as contamination and dataset-specific reuse constraints, such as licensing and citation requirements.
To support responsible reuse, \framework{} preserves source-dataset attribution for each instance and provides citation and license-reporting utilities that summarize source datasets, licenses, instance counts, citation guidelines, and BibTeX entries for customized evaluation sets. 
We further examine the possible effect of contaminated instances through a controlled simulation in Appendix~\ref{sec:data_contamination}.

\section{Auditing Benchmark Composition and Reusability}
\label{sec:sq1}
In this section, we use \framework to audit existing benchmarks as compositional resources rather than indivisible named datasets. 
Specifically, we examine the evaluation coverage hidden behind benchmark names and the extent to which it can be reused beyond their original purposes (Obs.~\ref{obs:obs1}).
Second, we validate whether benchmarks designed for similar high-level goals actually instantiate comparable mixtures of domains, skills, and cultural contexts (Obs.~\ref{obs:obs2}).

\begin{observation}\label{obs:obs1}
Benchmark Names Underdescribe Reusable Evaluation Coverage.
\end{observation}
\begin{figure}[htb!]
    \centering

    \begin{subfigure}[t]{0.64\textwidth}
        \centering
        \includegraphics[height=4.6cm]{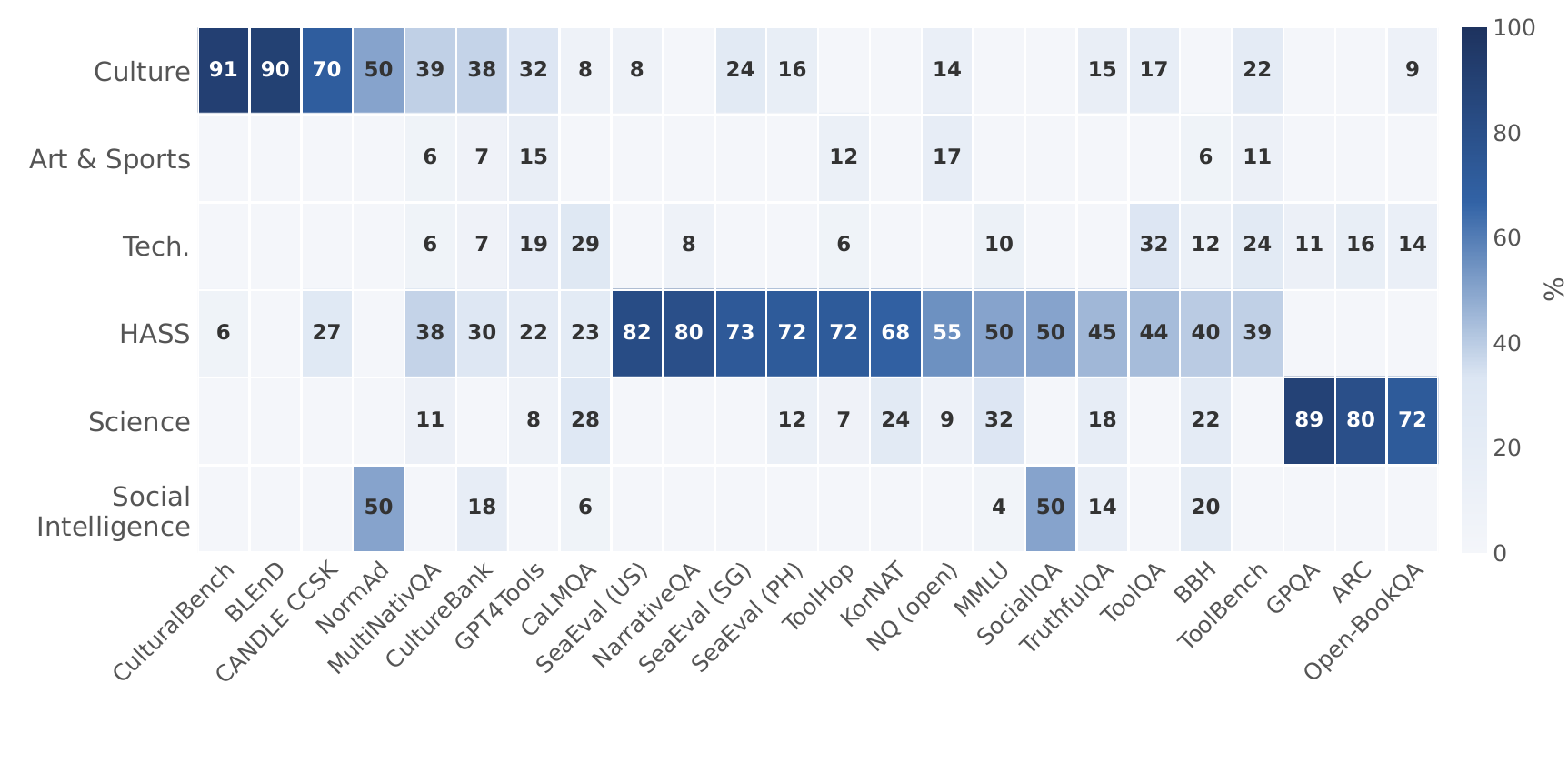}
        \caption{Subject distribution by benchmarks (\%, $p(\text{subject} | \text{benchmark})$)}
        \label{fig:en_bh_coverage}
    \end{subfigure}
    \hfill
    \begin{subfigure}[t]{0.34\textwidth}
        \centering
        \includegraphics[height=4.5cm]{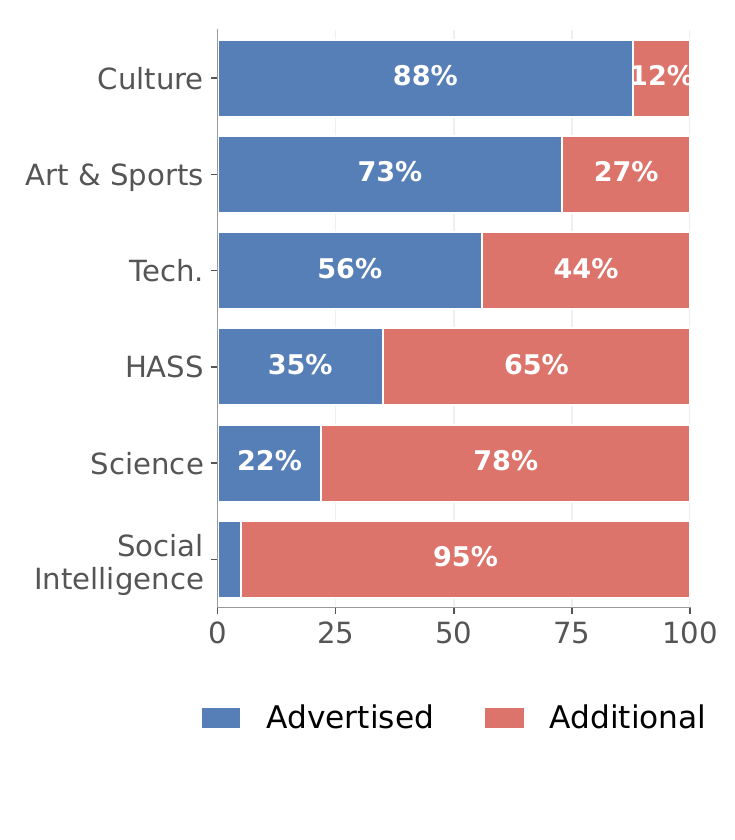}
        \caption{Advertised vs. Additional}
        \label{fig:proportion_unintended_cats}
    \end{subfigure}

    \caption{Subject coverage distribution and alignment in English benchmarks.}
    \label{fig:subject_coverage_and_intention}
\end{figure}
Figure~\ref{fig:en_bh_coverage} shows the subject distribution of each benchmark under the \framework taxonomy.
While most benchmarks have a dominant advertised focus, many contain substantial secondary coverage that is not apparent from their
names or high-level descriptions. 
For example, \texttt{NQ-open} includes culture-related questions about Indian currency and Chinese spice powder, while \texttt{ToolBench} includes \texttt{Art \& Sports}-tagged instances requiring historical NBA roster knowledge. 
Such examples show that existing benchmarks often contain reusable evaluation material that would be missed by name-based dataset selection.

We quantify this mismatch by comparing each benchmark's advertised scope with its instance-level \framework label.
We operationalize each benchmark's advertised scope based on the subject areas explicitly stated or clearly intended by the benchmark's original authors.
\begin{enumerate}[itemsep=1pt, topsep=0pt, partopsep=-10pt, parsep=0pt]
  \item One author first review the benchmark paper's abstract and the corresponding Huggingface README and assigns all subject labels supported by these sources.
  \item When these materials do not provide sufficient information, the author refer to the main paper and inspects dataset instances from Huggingface.
  \item If no subject category can be reliably inferred, the benchmark is left unmapped.
\end{enumerate}

A sample is counted as \emph{advertised} coverage if its \framework label matches this scope; otherwise, it is counted as \emph{additional} coverage exposed by \framework. To assess the reliability of this procedure, one more author independently repeats the same annotation process. Two annotators agree on 91.7\% of benchmarks at the coarse level.

In Figure~\ref{fig:proportion_unintended_cats}, 65.1\% of English samples fall outside their benchmarks' advertised subject categories.
This indicates that existing benchmarks contain substantial reusable coverage beyond their name.
For underrepresented domains, \framework provides a practical way to retrieve relevant instances that have already been embedded in existing benchmarks before constructing new ones from scratch.

\textbf{Takeaway.}
Benchmark names or their advertised scopes are incomplete interfaces to evaluation data.
\framework exposes hidden instance-level, fine-grained coverage, making existing benchmarks more visible, auditable, and reusable for domain-specific evaluation.

\begin{observation}\label{obs:obs2}
Similarly Motivated Benchmarks Encode Different Composition.
\end{observation}
\begin{figure}[htb!]
    \vspace{-3mm}
    \centering
    \includegraphics[width=\linewidth]{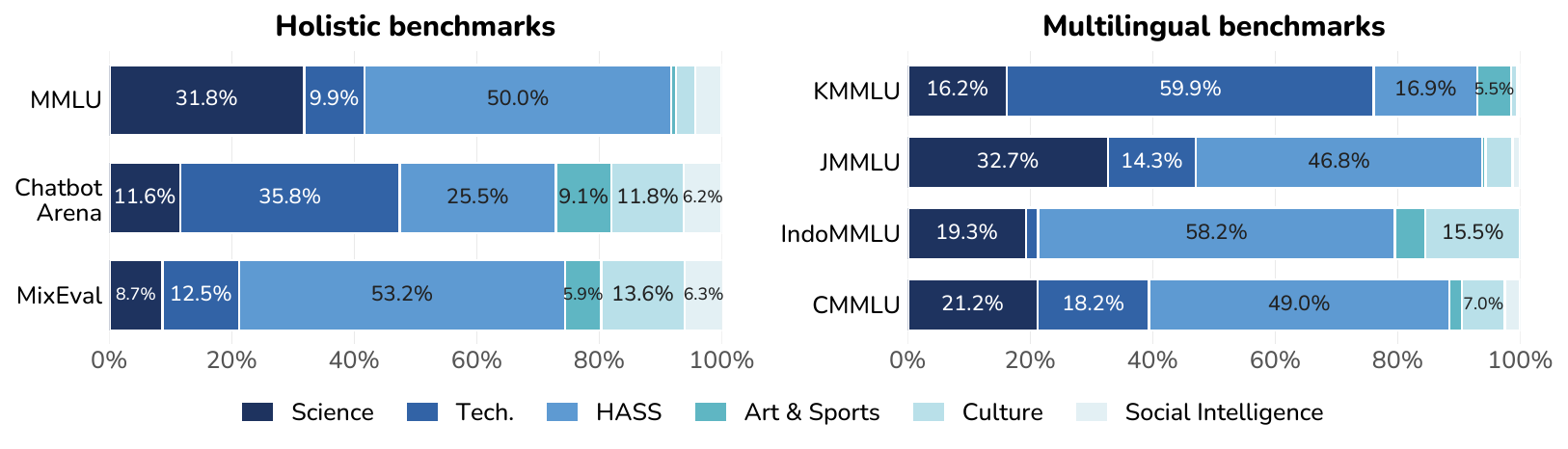}
    \vspace{-2mm}
    \caption{Subject distributions of existing evaluation benchmarks.}
    \label{fig:data_dist}
\end{figure}

In contrast to Obs.~\ref{obs:obs1}, benchmarks with similar high-level motivations may define different evaluation targets.
Here, we compare two benchmarks: (1) holistic benchmarks in English for general-purpose LLMs and (2) MMLU-style benchmarks in multiple languages.

\textbf{Holistic benchmarks operationalize general capability differently.} 
Figure~\ref{fig:data_dist} compares Chatbot Arena~\citep{chiang2024chatbotarena}, MixEval~\citep{ni2024mixeval}, and MMLU~\citep{hendrycks2021measuring} using our fine-tuned classifiers (\S\ref{sec:method-implement}).
While they are all used to summarize general model capability, their subject distributions differ substantially.
For example, Chatbot Arena contains only 25.5\% of Humanities and Social Science (HASS) questions, whereas both MixEval and MMLU contain more than half HASS questions.
Their skill distributions also diverge: MixEval includes fewer than 0.30\% value/alignment tasks and is heavily skewed toward knowledge evaluation (Figure~\ref{fig:mmlu_dist_bar_additional}).
These disparities indicate that holistic benchmarks do not instantiate distinct evaluation mixtures despite similar high-level positioning.

\textbf{Multilingual benchmarks introduce composition shifts across languages.} 
Composition differences are further amplified in multilingual settings. 
Figures~\ref{fig:data_dist} and \ref{fig:mmlu_dist} show that MMLU-style benchmarks vary substantially across languages.
For example, English MMLU emphasizes HASS, whereas Korean MMLU (KMMLU)~\cite{son-etal-2025-kmmlu} comprises 76.1\% STEM questions.
Thus, cross-lingual performance differences on such benchmarks may reflect shifts in subject distributions and mislead decision-making.
Composition-aware reporting is therefore essential to reliably interpret multilingual evaluation results.

\textbf{Takeaway.}
Similarly motivated benchmarks are not necessarily compositionally interchangeable.
Benchmark scores should be interpreted as performance on a specific mixture of domains, skills, formats, and languages, rather than as direct measurements of a single well-defined capability.

\section{Composition-Aware Model Evaluation and Selection}
\label{sec:sq2}
In this section, we examine how the hidden and uneven mixtures of existing benchmarks shape model evaluation. 
Using \framework, we move from benchmark-level evaluation to composition-aware evaluation.
First, we construct category-level evaluation sets across English and Korean benchmarks and show that model rankings vary substantially across different regions of the \framework taxonomy (Obs.~\ref{obs:obs3}). 
Second, we demonstrate that rankings are sensitive to the distribution of evaluation instances, revealing benchmark composition as a hidden variable in model comparison (Obs.~\ref{obs:obs4}). 
Finally, we show that making composition explicit enables use case-specific evaluation sets, leading to model-selection decisions that differ from those produced by general-purpose benchmark aggregation (Obs.~\ref{obs:obs5}).

\subsection{Evaluation Setup}
\label{sec:eval_setup}

\textbf{Dataset.} We select  32{,}954 English and 21{,}541 instances for English and Korean, respectively.
To manage the large number of fine-grained categories, we sample up to 150 examples per category, per problem format, fully including categories with 100--150 samples and merging categories with fewer than 80 samples into a miscellaneous group within the same coarse-grained category.

\textbf{Models.}
We use the following 15 open-weight models for the experiments:
Qwen~3 
(Qwen3-4B, Qwen3-8B, Qwen3-14B, Qwen3-32B, 
Qwen3-30B-A3B-Instruct-2507, Qwen3-Next-80B-A3B-Instruct);
Gemma3 
(gemma-3-4b-it, gemma-3-12b-it);
Llama~3
(Meta-Llama-3-70B-Instruct);
Mistral/Mixtral 
(Ministral-8B-Instruct-2410, Mistral-Small-3.2-24B-Instruct-2506, 
Mixtral-8x7B-Instruct-v0.);
and Olmo~3 
(Olmo-3-1025-7B, Olmo-3-1125-32B, Olmo-3.1-32B-Instruct). 

Evaluation details, including inference prompts and evaluation methods, are provided in Appendix~\ref{appendix:eval-details}.

\subsection{Evaluation Results of Open-weight Models using \framework}
\label{sec:eval_results}

\begin{observation}\label{obs:obs3}
Aggregate Scores Collapse Distinct Model Profiles.
\end{observation}

\begin{figure}[ht!]
    \vspace{-3mm}
    \centering
    \includegraphics[width=\linewidth]{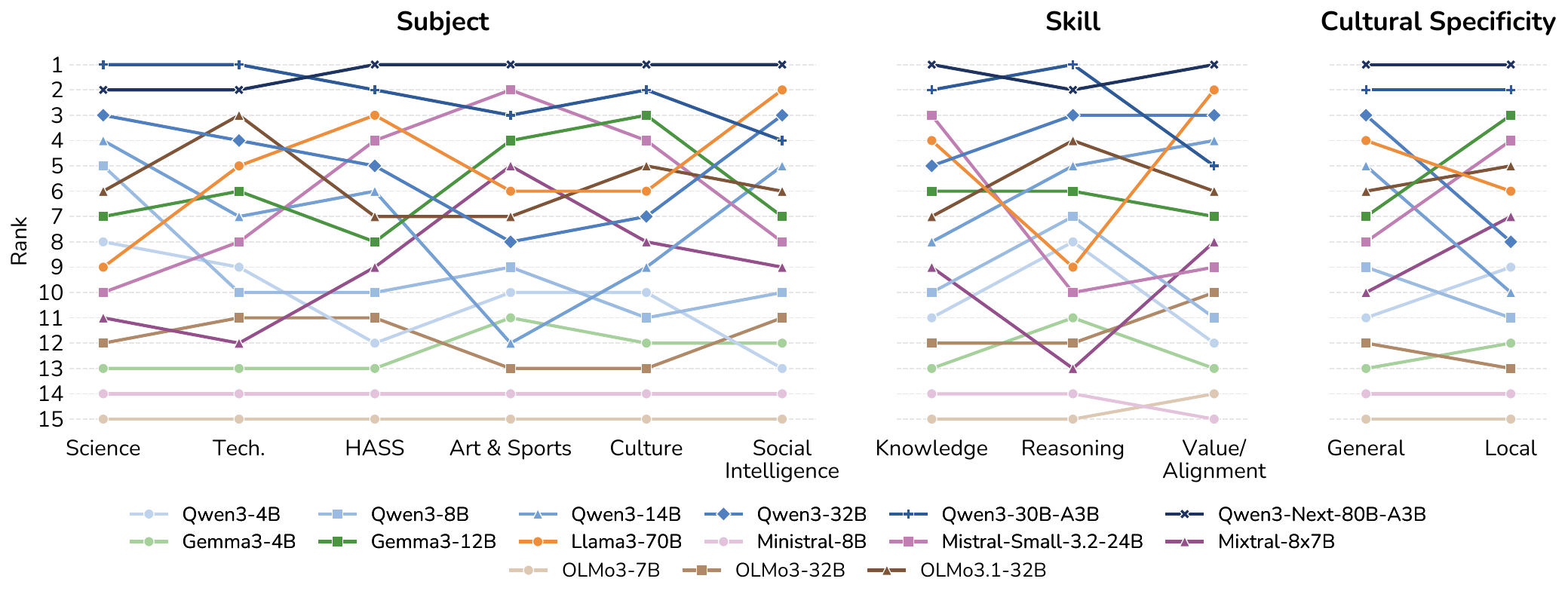}
    \vspace{-2mm}

    \caption{LLM evaluation rankings using \framework (English).}
    \label{fig:benchhub_ranking_en}
\end{figure}

\begin{figure}[ht!]
    \centering
    \includegraphics[width=0.95\textwidth]{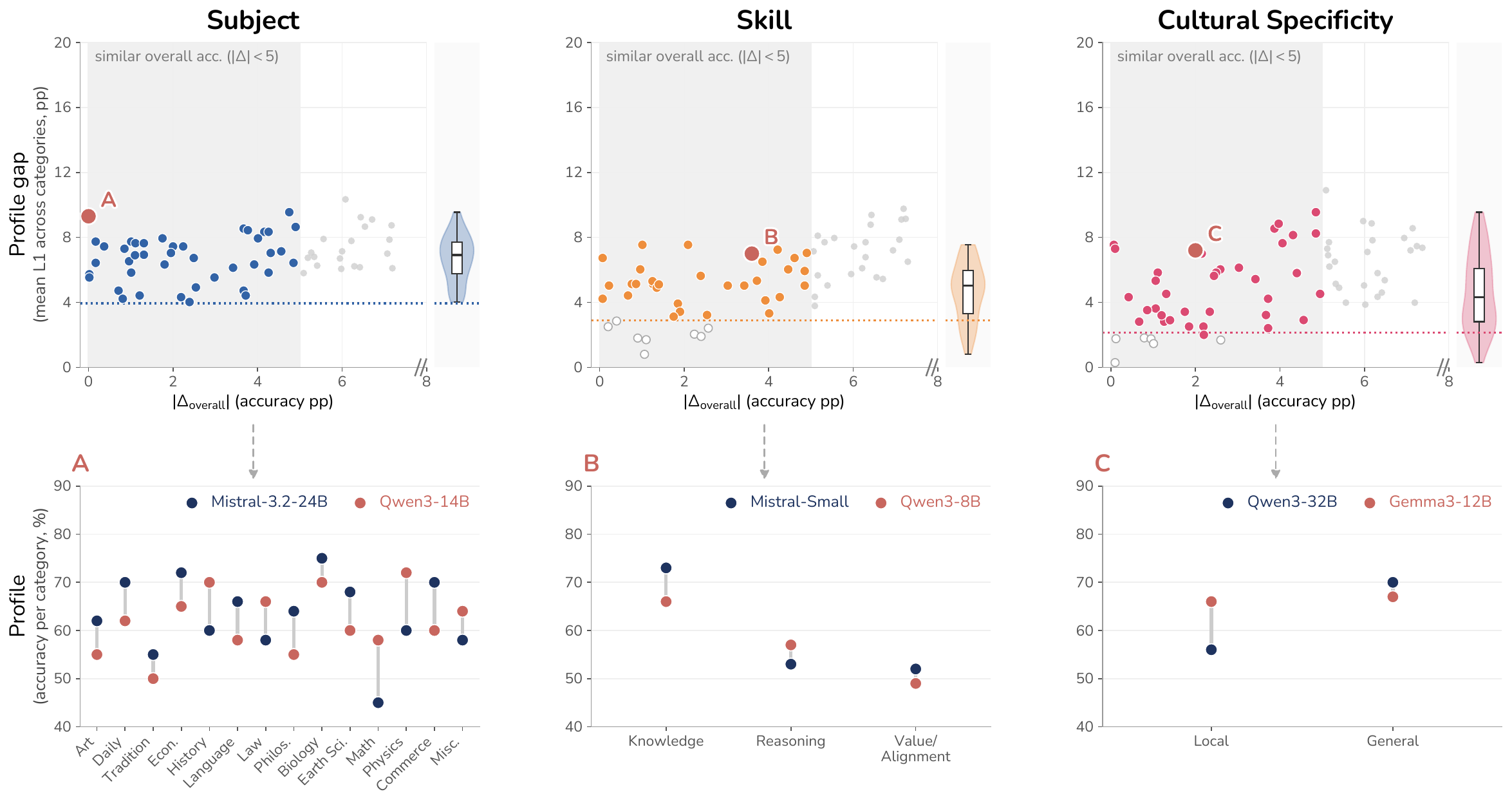}
\caption{
Model pairs with similar aggregate accuracy can show distinct category-level profiles.
Each point is an unordered model pair.
The shaded region marks pairs within 5 \%p differences in overall accuracy; red highlighted points (points A, B, C) indicate examples shown in the profile plots.}
    \label{fig:model_profile}
\end{figure}

\textbf{Rankings vary across composition dimensions.}
We first examine whether model rankings are stable across different categories of the \framework taxonomy.
As illustrated in Figure~\ref{fig:benchhub_ranking_en}, model rankings substantially fluctuate across subject, skill, and cultural-specificity categories.
In English, the median model-level rank range, defined as the difference between a model's worst and best category-wise rank within a dimension, is 5 rank positions for Subject, 3 for Skill, and 2 for Cultural-specificity.
Furthermore, pairwise model preference frequently flip: 29.5\% of model pairs reverse their ordering between Science and Art \& Sports.
We observe a similar trend of rank instability across Korean benchmarks, as shown in Figure~\ref{fig:benchhub_ranking_ko}.
These findings indicate that aggregate rankings collapse category-dependent model behavior into a single global order.

\textbf{Similar aggregate scores hide different strengths and weaknesses.}
To further demonstrate the masking effect of aggregation, we analyze model pairs whose overall accuracies differ by at most 5 percentage points. 
Figure~\ref{fig:model_profile} illustrates how \framework provides a higher-resolution view of model performance beyond aggregate scores. 
For two models and a taxonomy dimension, we define their \emph{profile gap} as the mean absolute accuracy difference across categories in that dimension. 
While near-score pairs in English exhibit a median overall gap of only 1.24 pp, their category-level profile gaps remain substantially larger: a median of 4.70 pp for Subject, 5.02 pp for Skill, and 4.49 pp for Cultural-specificity. 
In extreme cases, the maximum profile gap reaches up to 8.46 pp for Subject and 7.86 pp for Skill.
This suggests that models appearing functionally equivalent under aggregate scores can exhibit entirely disjoint strengths and weaknesses.
By decomposing performance along fine-grained taxonomy dimensions, \framework helps uncover these hidden differences and improves the interpretability of model evaluation.

\textbf{Takeaway.}
Model capability is not a single scalar.
Aggregate scores inherently collapse these profiles, destroying the nuance required for rigorous model selection.
When viewed through the \framework taxonomy, models reveal structured, interpretable performance profiles across subjects, skills, and cultural contexts.

\label{sec:sq3}
\begin{observation}\label{obs:obs4}
Benchmark Composition Shapes Model Rankings.
\end{observation}

We investigate whether evaluation conclusions, such as model rankings and model selection decisions, are robust to changes in benchmark composition.
While benchmark scores are often treated as stable indicators of model capability, this implicitly assumes that evaluation outcomes are invariant to the distribution of evaluation instances.

To quantify how evaluation composition affects model comparison, we repeat 100 simulations to construct 1k-instance evaluation sets under four sampling strategies: (1) random sampling, (2) stratified sampling using the subject categories in \framework, and sampling following the distributions of (3) MixEval and (4) Chatbot Arena.
These strategies induce substantially different subject mixtures.

\begin{figure*}[t]
    \centering
    \includegraphics[width=\linewidth]{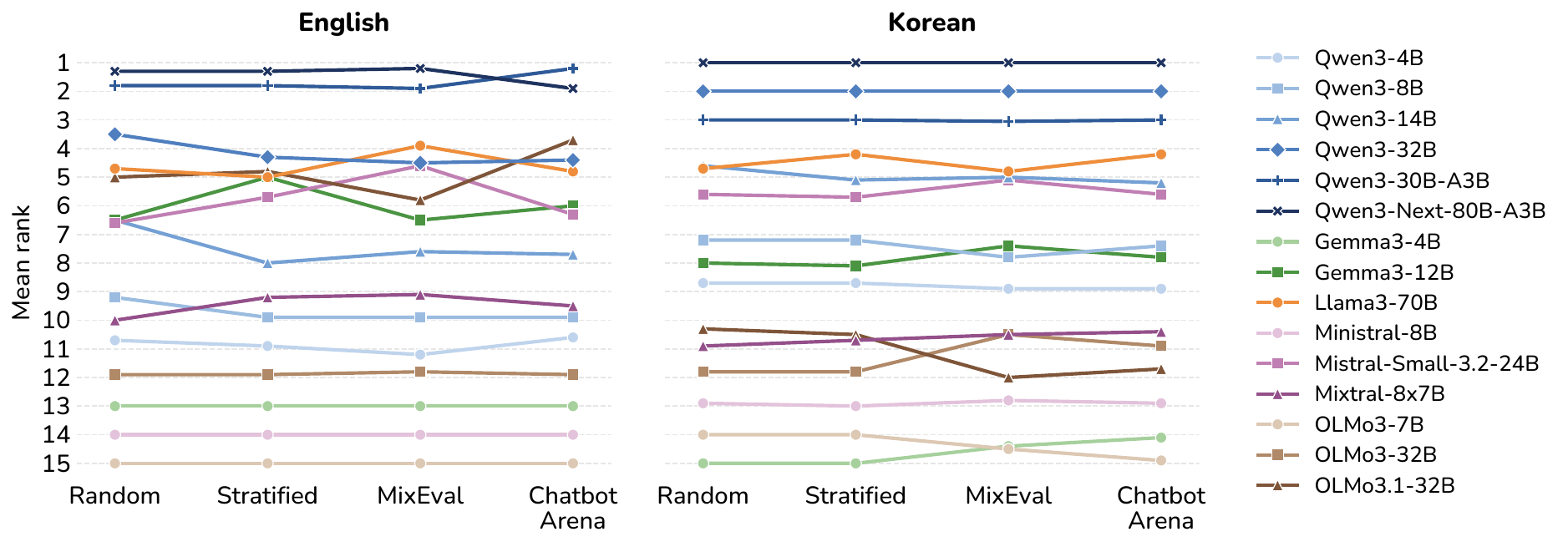}
    \caption{LLM ranking according to four sampling methods.}
    \label{fig:simulation_ranking}
\end{figure*}

These composition shifts lead to systematic changes in model scores and rankings.
Across sampling strategies, the average model-level accuracy range is 3.39 pp in English and 4.95 pp in Korean, with maximum shifts of 4.95 and 7.25 pp, respectively.
Rank distributions also shift broadly: after Benjamini--Hochberg correction, 53.3\% model-by-method-pair comparisons in English and 54.4\% in Korean show significant rank-distribution differences tested by Mann--Whitney U test ($p < 0.01$).

These results demonstrate that benchmark composition acts as a hidden intervention on model evaluation: even when global leaderboard structure is preserved, the evaluation distribution can shift scores, alter model-specific ranks, reverse pairwise preference, and change selected models.

\textbf{Takeaway.}
Model rankings are not invariant properties of models alone.
They also depend on how the evaluation set is constructed.
Benchmark composition therefore acts as a hidden variable in model comparison: it shapes which capabilities are emphasized, which models appear strongest, and how leaderboard rankings should be interpreted.

\begin{observation}\label{obs:obs5}
Application-aligned Evaluation Can Change Model Selection.
\end{observation}

Since model rankings depend on benchmark composition, evaluation sets should be strongly aligned with the intended deployment context rather than inherited from a generic benchmark distribution. 
\framework supports this by allowing users to specify an explicit target composition over language, subject, skill, answer format, and cultural-specificity dimensions. 
The resulting score is computed over instances matching this declared profile, making the evaluation target transparent and reproducible.

We initiate this workflow with five application scenarios, including an English legal chatbot and a Korean docent agent.
Full profile definitions with additional scenarios are provided in Appendix~\ref{appendix:customize_benchhub_ex}.

\begin{enumerate}[leftmargin=*, nosep, label=(\alph*)]    
    \item Legal chatbot with advanced reasoning: we construct an English test set combining legal domain knowledge, along with reasoning and free-form generation capabilities.
    \item Docent agent for Korean traditional arts: we construct a Korean test set focusing on expert knowledge in Korean traditional arts.
\end{enumerate}

\begin{table}[htb!]
\centering

\caption{Top-5 LLMs evaluated by \framework in real-world application scenarios.}
\label{tab:customized_result_top_n}

\resizebox{\linewidth}{!}{
\begin{tabular}{@{}c|cc|cc@{}}
\toprule
\multirow{2}{*}{Rank} & \multicolumn{2}{c|}{\begin{tabular}[c]{@{}c@{}}(a) Legal chatbot with advanced reasoning (EN)\end{tabular}} & \multicolumn{2}{c}{\begin{tabular}[c]{@{}c@{}}(b) Docent agent for Korean traditional art (KO)\end{tabular}} \\ \cmidrule(l){2-5} 
 & Customized & Stratified & Customized & Stratified \\ \midrule
1 & Qwen3-30B-A3B-Instruct-2507 & Qwen3-Next-80B-A3B-Instruct & Qwen3-Next-80B-A3B-Instruct & Qwen3-Next-80B-A3B-Instruct \\
2 & Qwen3-Next-80B-A3B-Instruct & Qwen3-30B-A3B-Instruct-2507 & Qwen3-32B & Qwen3-32B \\
3 & Olmo-3.1-32B-Instruct & Qwen3-32B & Mistral-Small-3.2-24B-Instruct-2506 & Qwen3-30B-A3B-Instruct-2507 \\
4 & gemma-3-12b-it & Olmo-3.1-32B-Instruct & Meta-Llama-3-70B-Instruct & Meta-Llama-3-70B-Instruct \\
5 & Meta-Llama-3-70B-Instruct & Meta-Llama-3-70B-Instruct & gemma-3-12b-it & Qwen3-14B \\ \bottomrule
\end{tabular}
}
\end{table}

Compared with a generic stratified baseline, customized evaluation changes the top-5 candidate set in every case.
For the English legal chatbot, the top model changes from {Qwen3-Next-80B-A3B-Instruct} under the generic baseline to {Qwen3-30B-A3B-Instruct-2507} under the customized profile.
Even when the top model remains stable, the candidate set can shift: in the Korean docent scenario, {Mistral-Small-3.2-24B-Instruct} rises into the top-3 and Gemma3-12B enters the top-5, while {Qwen3-30B-A3B} drops out.

These results show that application-aligned evaluation is not merely a narrower version of generic benchmarking.
Changing the target composition changes which capabilities are emphasized, which in turn can change the model selected for deployment. 
\framework therefore makes model selection a transparent composition-design problem rather than an implicit consequence of benchmark choice.

\textbf{Takeaway.} 
Generic benchmark rankings do not necessarily identify the best model for a specific use case.
By making the target distribution explicit, \framework shows when application requirements change the selected model or candidate set.

\section{Related Work}
\label{sec:rel-work}
As LLMs become integral to real-world generative AI systems, the focus on benchmarks and leaderboards has matured into evaluation \emph{science}~\citep{weidinger2025toward,saxon2024benchmarks}.
Our work contributes to this evolution by examining benchmark composition as both a property of the benchmark ecosystem and a factor that shapes model rankings and selection decisions.

\textbf{Benchmark Auditing.}
Recent work has begun to audit benchmarks themselves, asking whether benchmark can provide reliable measurement signals. 
This includes studies of data contamination, benchmark validity, discriminative power, item quality, and benchmark documentation~\citep{kim-etal-2025-benchmark,sokol2025benchmarkcardsstandardizeddocumentationlarge,bean2025measuringmattersconstructvalidity,zhou2025lost}. 
However, they do not characterize the broader benchmark ecosystem in terms of covered, missing, or redundantly reused content across datasets. 
In contrast, \framework{} provides an ecosystem-level view by mapping benchmark instances into a shared taxonomy, revealing composition skew, hidden coverage beyond dataset names, and cross-benchmark redundancy for more composition-aware analysis and reuse.

\textbf{Benchmark aggregation and instance-level annotation.}
Prior work has expanded LLM evaluation by aggregating diverse benchmarks into broad evaluation suites. 
HELM evaluates models across multiple scenarios and metrics to capture trade-offs beyond accuracy~\citep{liang2023holistic}, while MixEval constructs benchmark mixtures aligned with real-world user queries~\citep{ni2024mixeval}. 
Although aggregation improves coverage, it can obscure the internal composition of evaluation sets. 
A complementary line of work addresses this through instance-level annotations. 
FLASK provides human annotations of fine-grained skills, domains, and difficulty levels, improving interpretability but limiting scalability~\citep{ye2024flask}. 
ADeLe instead uses automated demand-level annotations to improve the explanatory and predictive power of evaluation~\citep{zhou2025generalscalesunlockai}. 
\framework{} extends this direction with a comprehensive, automated, and dynamically extensible framework for composition-aware evaluation, making benchmark contents labelable, inspectable, and reusable at scale.

\textbf{Controllable model evaluation.}
Recent work shows that model rankings and selection decisions are sensitive to benchmark choice and evaluation-set composition~\citep{siska-etal-2024-examining,garbacea-etal-2026-personalized,mishra2021robustmodelrankingsleaderboard}. 
Motivated by this, prior studies have proposed controllable evaluation frameworks, including user-adjustable leaderboards, instance reweighting, and prompt-specific ranking methods~\citep{mishra2021robustmodelrankingsleaderboard,jung2026definesbestinteractiveuserdefined,frick2025prompttoleaderboard}. 
\framework{} advances this line of work by making benchmark composition explicit and controllable at the instance level. 
Rather than relying only on benchmark- or prompt-level customization, \framework{} maps instances into a shared taxonomy, enabling users to construct and rebalance evaluation sets for specific domains, skills, cultural contexts, and deployment needs.

\section{Conclusion}
\label{sec:discussion}
We introduce \framework{}, a composition-aware evaluation framework that organizes existing LLM benchmarks into a shared instance-level representation. \framework{} integrates 54 benchmarks comprising 839K samples across 10 languages, with a primary focus on English and Korean.
By making implicit benchmark composition explicit, \framework{} shows that benchmark-level usage can obscure the reusable structure of evaluation data. This reveals that existing benchmarks may support broader evaluation purposes than their names or high-level descriptions suggest. \framework{} also enables transparent comparisons among benchmarks designed for similar purposes by exposing differences in their internal composition.
Beyond benchmark analysis, \framework{} provides a higher-resolution view of model performance through fine-grained instance-level categories, while also supporting the construction of controllable evaluation sets tailored to specific application needs. Together, our findings suggest that treating benchmarks as compositional resources, rather than fixed atomic units, can make LLM evaluation more interpretable, reusable, and application-aligned. 
We envision \framework{} and its accompanying tools supporting both researchers and practitioners as they explore existing benchmarks, construct evaluation sets for new purposes, and evaluate and select models that better align with their specific needs.
\section*{Limitations}
\label{appen:limitations}
By acknowledging these limitations, we aim to continuously improve \framework and encourage contributions from the community.

\textbf{Incomplete English Dataset Coverage.}
Due to the vast amount of English-language evaluation data, we could not include all relevant datasets in this version of \framework. While we prioritized widely used and high-quality benchmarks, some important datasets may still be missing. Future iterations will expand coverage.

\textbf{Categorization Bias from LLMs.}
\framework's automated categorization relies on Qwen-2.5-7B, which may introduce biases due to its training data or modeling limitations. Although we validate robustness to categorization errors up to 1.5\% (Appendix~\ref{appendix:reliabilty_cats}), future work will explore human-in-the-loop methods and ensemble approaches.

\textbf{Potential Data Contamination.}
\framework aggregates existing benchmarks and thus inherits any contamination present in its sources. While \framework itself does not introduce new contamination risk, results may be influenced if evaluated models were exposed to included benchmark samples. We provide a controlled contamination simulation in Appendix~\ref{sec:data_contamination}.
\section*{Acknowledgments}
This work was supported by Institute of Information \& communications Technology Planning \& Evaluation(IITP) grant funded by the Korea government(MSIT) (No. RS-2024-00509258 and No. RS-2024-00469482, Global AI Frontier Lab). This work was supported by Artificial intelligence industrial convergence cluster development project funded by the Ministry of Science and ICT(MSIT, Korea)\&Gwangju Metropolitan City.
\clearpage
\bibliographystyle{plain}
\bibliography{reference/custom,reference/anthology-1, reference/anthology-2}


\clearpage
\section*{Appendix}
\appendix

\section{Broader Impact Statement}
\label{sec:broader_impact}

\framework{} aims to make LLM evaluation more transparent, reusable, and application-aligned by exposing the instance-level composition of existing benchmarks. Its potential positive impacts include helping researchers and practitioners better understand what benchmarks actually contain, identify underrepresented subjects or skills, and construct evaluation sets that better match their intended applications. Its positive impacts can further extend to multilingual evaluation by enabling comparisons of benchmark coverage across languages and helping identify underrepresented languages and cultures in existing evaluation resources. By moving beyond benchmark names and aggregate scores, \framework{} can also support more interpretable model comparison and reduce the risk of selecting models based on incomplete or misleading benchmark-level summaries.

At the same time, composition-aware evaluation can introduce potential risks if used uncritically. First, users may over-rely on automatically assigned instance-level labels, even though such labels can be incomplete or noisy. This could lead to inappropriate conclusions about model capabilities or benchmark coverage. Second, the ability to construct customized evaluation sets may be misused to selectively report favorable model rankings or to overfit evaluation choices to a desired conclusion. Third, because \framework{} reveals reusable contents within existing benchmarks, it could encourage benchmark reuse in contexts for which the original data were not designed.

To mitigate these risks, we provide validation analyses of the automated categorization process and encourage users to inspect sampled instances, account for label uncertainty, and clearly document the criteria used to construct customized evaluation sets. To support transparent and reproducible evaluation, we provide a code utility that reports both aggregate and fine-grained results, along with the composition of the selected evaluation data.

\clearpage
\section{Interactive Platform and Utilities of \framework}
\label{appendix:benchhub_web}
We provide a web interface (Appendix~\ref{appendix:benchhub_UI}), additional code utilities for data loading and evaluation (Appendix~\ref{appendix:benchhub_code_utility}).

\begin{itemize}[leftmargin=*]
    \item Our interface is available as a Hugging Face Space: \url{https://huggingface.co/spaces/BenchHub/BenchHub}.
    \item Our code utilities are available on AnonymousGitHub: \url{https://anonymous.4open.science/r/BenchHub-56B8/}.\footnote{We will make the repository public upon acceptance.}
\end{itemize}

\subsection{\framework Web Interface}
\label{appendix:benchhub_UI}
\begin{figure}[htbp]
    \centering

    \begin{subfigure}[b]{0.74\textwidth}
        \centering
        \includegraphics[width=\linewidth]{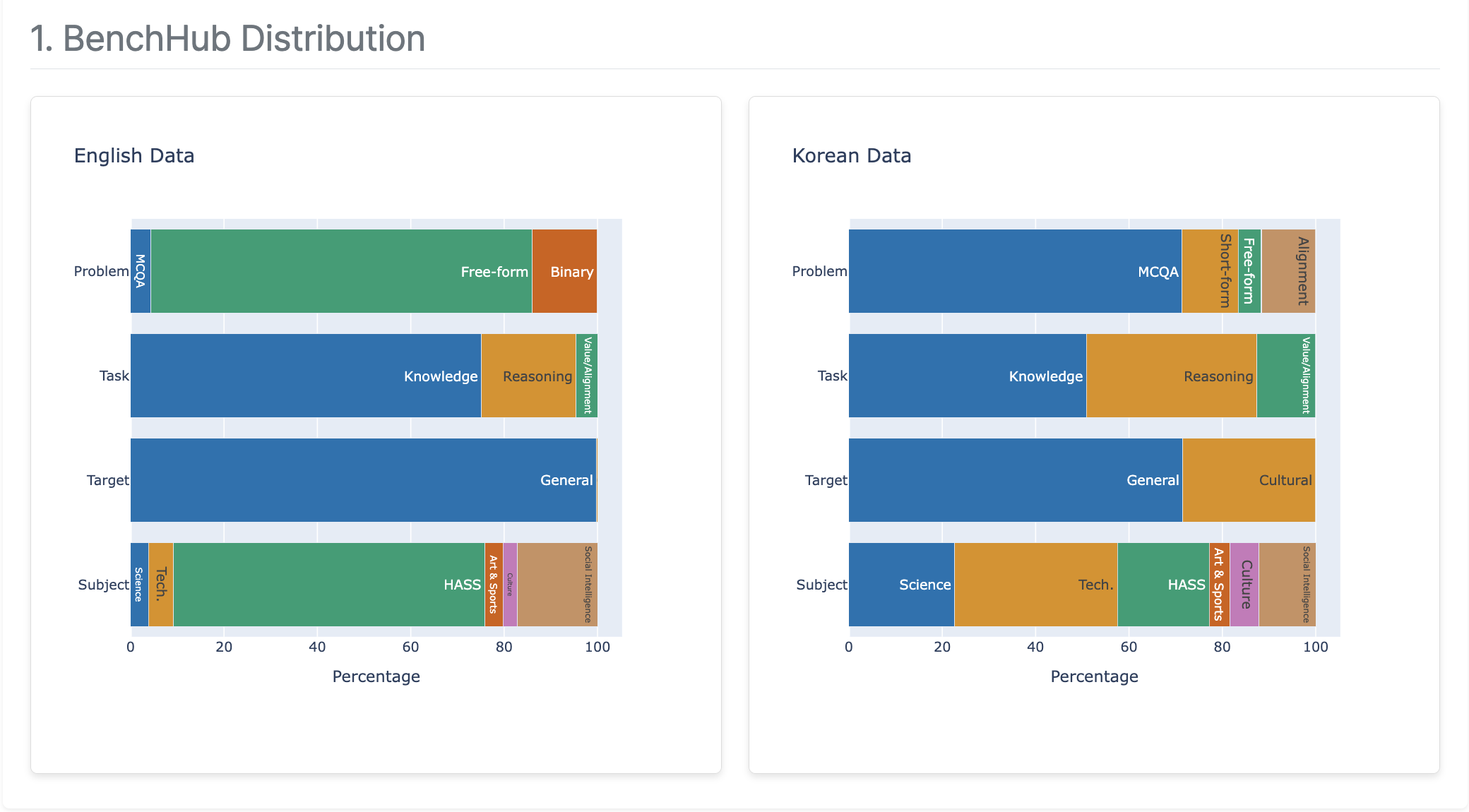}
        \caption{\framework Distribution}\label{fig:ui_1}
    \end{subfigure}

    \begin{subfigure}[b]{0.74\textwidth}
        \centering
        \includegraphics[width=\linewidth]{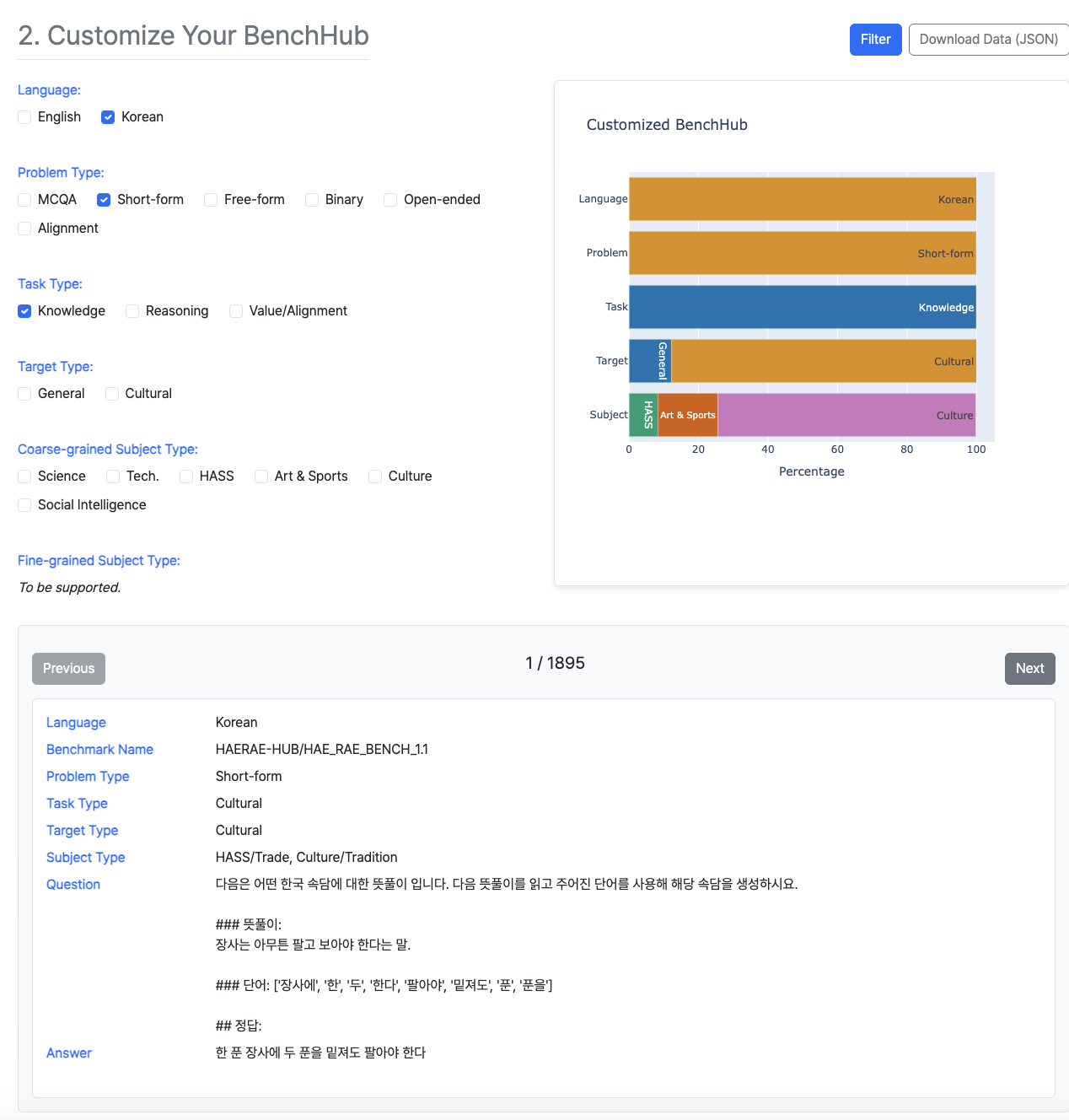}
        \caption{Dataset Selection}\label{fig:ui_2}
    \end{subfigure}
    
    \begin{subfigure}[b]{0.74\textwidth}
        \centering
        \includegraphics[width=\linewidth]{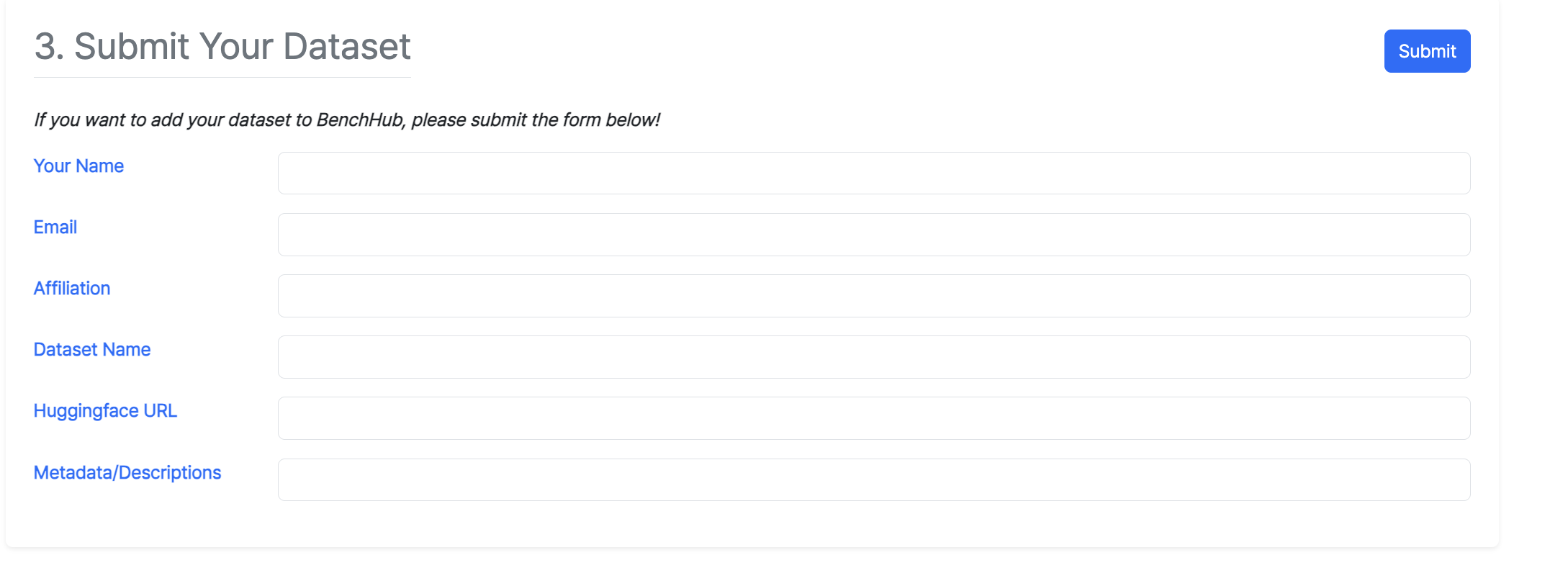}
        \caption{New Dataset Submission}\label{fig:ui_3}
    \end{subfigure}
    
    \caption{User Interface of \framework Web Demo.}
    \label{fig:web_demo}
    \label{fig:ui}
\end{figure}
We provide the \framework web interface (Figure~\ref{fig:ui}) to enable users to interactively explore available datasets and identify those that best suit their needs. It also supports the continuous addition and management of new data. Through a submission form, new datasets can be detected and automatically added. To achieve these, we provide three main functions.

\textbf{(1) \framework Distribution} (Figure~\ref{fig:ui_1})
This feature offers comprehensive statistics of all datasets we have. Users can interactively explore the overall data distribution they are interested in. Additionally, it provides researchers with insights into which datasets are currently lacking and which evaluations have not yet been conducted.

\textbf{(2) Customizing \framework} (Figure~\ref{fig:ui_2})
This allows users to access sample lists and statistics for selected categories. By reviewing samples, users can verify whether the dataset matches their needs and explore datasets suitable for their purposes. Users can also download the entire set corresponding to the samples.\footnote{Additional customizing features, such as fine-grained category adjustments and interactive control of category proportions via the platform (\eg adjusting the ratio between reasoning and knowledge questions), are to be developed.}

\textbf{(3) Submitting New Dataset} (Figure~\ref{fig:ui_3})
To facilitate the addition of new datasets, we provide a submission section to input the Dataset Name, Huggingface URL, and Metadata/Descriptions. Based on this information, the author decides whether to add the dataset to \framework.
\subsection{\framework Code Utilities}
\label{appendix:benchhub_code_utility}

To proliferate our structured data into actionable insights for researchers and practitioners, we release code utilities.
The web demo allows users to filter out datasets by any category combinations, inspect statistics, download their customized subsets, and propose new datasets via pull requests.
The code utilities offer two main features:

\subsubsection{Dataset Loader} 
The dataset loader filters the aggregated benchmark pool according to the categories selected by the user. 
It provides two loading options: 
(1) returning all instances that match the specified categories, or 
(2) returning a deduplicated version in which overlapping entries, including near-duplicates, are removed. 
The latter is useful because multiple aggregated benchmarks may contain overlapping or semantically similar samples.

\textbf{Duplicate filtering method.} 
To perform deduplication, we implement an embedding-based sampling method inspired by MixEval~\citep{ni2024mixeval}. 
The process consists of two steps: 
(1) computing query embeddings using \texttt{mpnet-base-v2} from SentenceTransformers and projecting them into a two-dimensional space via t-SNE, and 
(2) uniformly sampling instances in this reduced space. 
Because queries on similar topics tend to cluster in localized regions of the embedding map, this procedure helps reduce redundant samples during dataset construction.

\textbf{Empirical validation.} 
To validate this approach, we conduct a controlled duplicate-injection experiment. 
We first extract 7,715 English \framework samples categorized under mathematics. 
We then introduce 60 synthetic duplicates by prompting \texttt{gemini-2.5-flash} to generate identical copies and five near-duplicates for 10 randomly selected questions through paraphrasing or numerical perturbation. 
Finally, we apply the embedding-based projection and uniform sampling procedure described above.

\begin{figure}[htb!]
    \centering
    \includegraphics[width=0.5\linewidth]{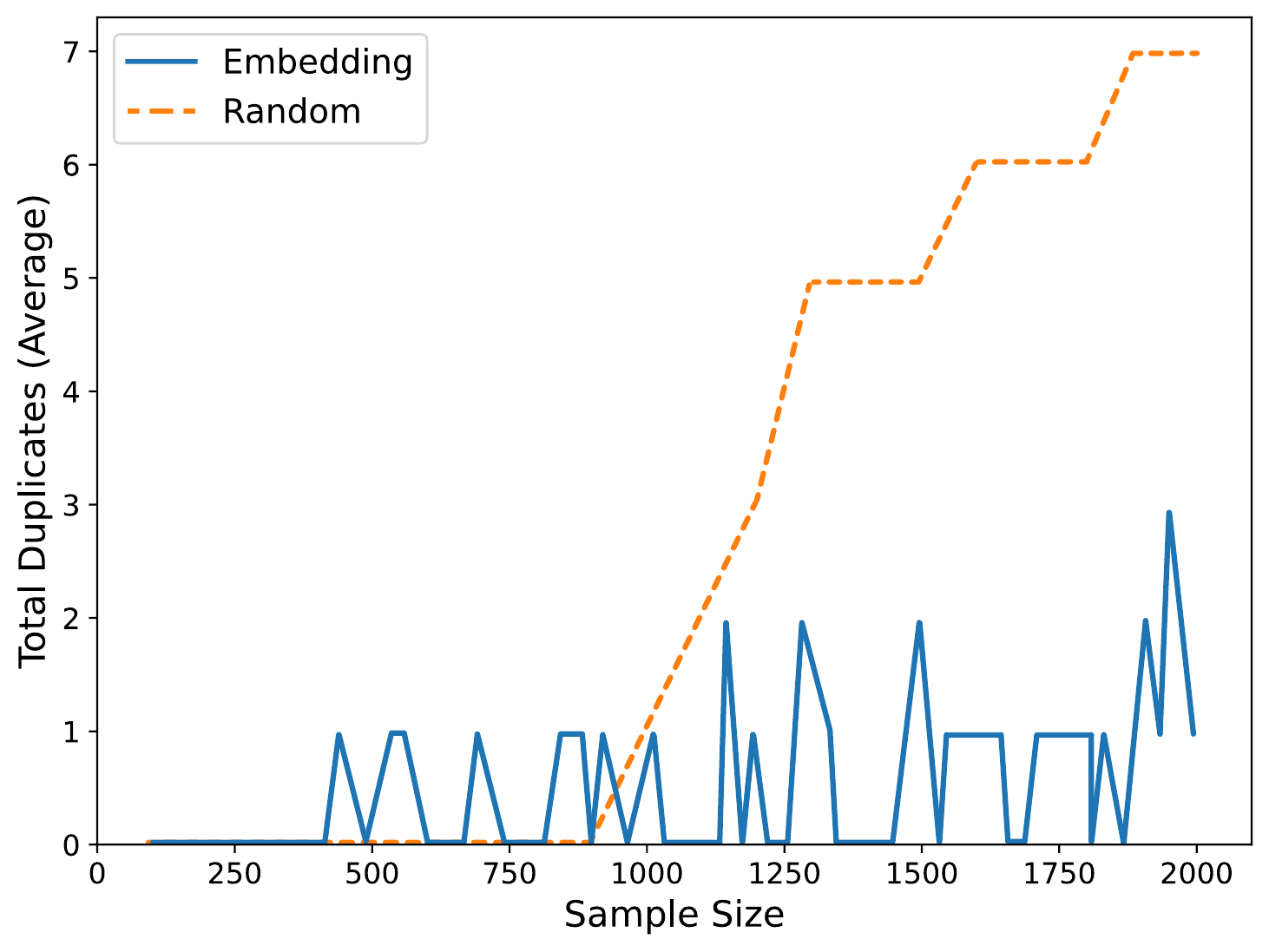}
    \caption{Average number of duplicates included in the sampling size when using the embedding-based method (Blue) and random sampling (Orange).}
    \label{fig:sampling_compare}
\end{figure}

We find that embedding-based sampling consistently limits duplicates to at most 0--1 per batch, even at large sample sizes. 
In contrast, random sampling frequently yields more than five duplicates once the sample size exceeds 1,250. 
Detailed results are shown in Figure~\ref{fig:sampling_compare}. 

\subsubsection{Citation Report Generator} 
For each customized dataset, \framework generates a La\TeX{} report that summarizes source datasets, licenses, instance counts, and citation guidelines. 
It also provides Bib\TeX{} entries to help users directly incorporate the selected dataset into their projects while ensuring proper credit to the original dataset authors.

\begin{tcolorbox}[title=Example of Citation Guidelines, breakable, enhanced, top=1pt] 

\begin{verbatim}
The evaluation dataset are sampled using BenchHub
~\cite{benchhub}. 

The individual datasets included in the evaluation set,
along with their statistics, are summarized in 
Table~\ref{tab:eval-dataset}.

% Please add the following required packages to your document
preamble:
% \usepackage{booktabs}
\begin{table}[h]
\centering
\begin{tabular}{@{}lll@{}}
\toprule
\textbf{Dataset} & \textbf{Number of Samples} 
& \textbf{License}\\ \midrule
{table_content}
\bottomrule
\end{tabular}
\caption{Breakdown of datasets included in the evaluation set.}
\label{tab:eval-dataset}
\end{table}

% --- BibTeX Entries ---
@inproceedings{...}
@inproceedings{...}
\end{verbatim}
\end{tcolorbox}

\subsubsection{Automatic Evaluation Pipeline}
For better reproducibility, we adopt HRET~\citep{lee-etal-2026-redefining}\thinspace\footnote{HRET is an evaluation toolkit supporting multiple datasets, including \framework.}, enabling direct evaluations on \framework.

We manage all code, datasets, models, and demo via Huggingface. 
In this repository, we release:
(1) the complete datasets,
(2) useful codes (\eg load and preprocess dataset),
(3) the interactive web interface, and
(4) our categorizer model.

\section{Implementation of \framework}
\label{appendix:implementation}
\framework follows three stages: (1) reformatting, (2) metadata assignment, and (3) sample-level categorization. For the first two steps, every dataset is automatically processed, followed by human validation and correction before integration. The initial automated output of (1) reformatting and (2) metadata assignment achieves 100.0\% and 96.4\% agreement with human annotations, respectively.

\subsection{Automated Categorization Process}
\label{appendix:categorizer_mono}
Here, we provide a detailed description of sample-level categorization and its validation in the following section.

\subsubsection{Training Categorizer for English and Korean Language}

\begin{wraptable}{r}{0.45\textwidth}
\vspace{-4.5mm}
\caption{SFT configuration details.}
\label{tab:sft_config}

 \begin{tabular}{lr}
    \toprule
    \textbf{Hyperparameter} & \textbf{Value} \\ \midrule
    Sequence Length & 8,192 \\
    Learning Rate & \(2 \times 10^{-5}\) \\
    Global Batch (Effective) & 256 \\
    Learning Rate Scheduler & Cosine Decay \\
    Warmup Ratio & 0.05 \\
    Training Epochs & 3 \\ \bottomrule
    \end{tabular}
\end{wraptable}

We fine-tune the Qwen-2.5-7B-instruct models
\footnote{This model is publicly available via huggingface: \href{https://huggingface.co/BenchHub/BenchHub-Cat-7B}{BenchHub/BenchHub-Cat-7B}} 
to automatically categorize the skill, subject, and cultural specificity type of a given sample. 
Table~\ref{tab:sft_config} shows the SFT configs of the categorizer model.
In Table~\ref{tab:cat_acc}, we report accuracy by comparing model predictions against human-annotated labels.
Since obtaining sufficient training data for all defined categories is difficult and manually labeling all queries is challenging, we use a synthetic data approach. Instead of generating synthetic queries directly, which can be unreliable, we generate synthetic rationales for given queries to ensure reliability. The process is as follows: first, we create all possible combinations of our three categories---skill, subject, and target. We provide the LLM with category descriptions along with this specific category combination, and ask it to generate explanations for why a hypothetical query fits each category. We use GPT-4o~\citep{openai2024gpt4ocard} as a synthetic rationale generator. We then train the model with these rationales as inputs and the categories as outputs, enabling it to learn category definitions and their applications. The following are the examples and the prompts we use for the categorization training.

For the target category, we adopt a binary classification scheme consistent with prior work such as Global MMLU~\cite{singh-etal-2025-global}. After extracting the raw ``target'' label (\eg South Korea) from the categorizer’s output, we further refine it into two subcategories: \textit{Local}, if the model specifies a particular cultural or local context, and \textit{General}, if the model determines that the query is culturally independent.

\begin{tcolorbox}[title=Example of Rationale, breakable, enhanced, top=1pt]
example = "The query is asking about the cause of symptoms (vomiting and diarrhea) in a 6-year-old boy who ate kimbap at kindergarten and later experienced these symptoms along with three other children. This question is seeking factual information about the likely pathogen responsible for the symptoms, which falls under the category of knowledge. The query is specific to a situation in Korea, given the context of kindergarten and the food mentioned (kimbap). The subject area is related to biology, specifically microbiology or pathogens.
\end{tcolorbox}
\begin{tcolorbox}[title=Prompt for Rationale Generation of Given Query,breakable, enhanced, top=1pt, left=1pt, right=1pt, bottom=1pt]
I want to assign three categories to the following query, but before doing this, you should create a description of the given query. Explain the query first (\eg what the question is asking about (\ie subject type), the type of ability needed to solve it (\ie skill type), whether it’s a question about a specific culture or a general question (\ie cultural specificity type), etc.). Refer to the definition of each label and the output format. \\
Label Definition: \{description\} \\
Now, create a description for the following query.
\end{tcolorbox}

\begin{tcolorbox}[title=Prompt for Synthetic Rationale Generation, breakable, enhanced, top=1pt, 
title = Prompt for Synthetic Rationale Generation] 
The following are the categories of one query, with an explanation for each category provided below. Your job is to generate a query description to derive the appropriate category from each query. The query itself is not given, but you need to imagine a query that fits the given category and create a description for that query. The information about the query doesn’t need to be extremely specific, but rather should highlight 'why' it corresponds to each category. Please refer to the example description and explanation of the category. \\
Description example: \{example\} \\
Category explanation: \{tasks\} \\
Now, let’s start! \\
Given category: \{category\} \\
Your Description:

\end{tcolorbox}

\begin{tcolorbox}[title=Prompt for Synthetic Rationale Generation, breakable, enhanced, top=1pt, 
title = Prompt for Category Generation] 
    **You are an agent tasked with assigning three categories—`subject\_type`, `task\_type`, and `target\_type`—to describe what is required to answer the following prompt.**

* **subject\_type**: What domain of knowledge or skill is needed?
* **task\_type**: What type of cognitive process or reasoning is involved?
* **target\_type**: Is the required knowledge or skill specific to a particular country or culture?

Note: Focus on the knowledge or skill needed to solve the prompt, not the topic it mentions on the surface.  
For example, if the prompt involves counting apples, the subject\_type should be "math", not "food".

The following text is a meta data of a certain prompt. Based on this data, assign three labels to the following data. Refer to the description of each label and the output format. Present the output in the following format: {'task\_type' : str,'target\_type' : str,'subject\_type' : LIST[str]}

Please refer the following information:
\#\#\# **skill type Description**  
- **task\_type** indicates the type of task the query belongs to. Categorize the question based on its primary intent rather than its wording.  

\#\#\#\# **Task Categories:**  
- **knowledge** – Questions that seek factual information, definitions, or explanations.Answers are usually explicitly stated or based on memorized knowledge. 
    - Example: *"What is the capital of France?"*  
    - Example: *"What is the pythagorean theorem?"*
- **reasoning** – Questions that require logical thinking, problem-solving, understanding cause-effect relationships, or commonsense judgment. Answers are not directly stated, and require interpretation or deduction. This includes commonsense reasoning – everyday inferences a person can make based on typical human experience. 
    - Example: *"If a train departs at 3 PM and travels at 60 km/h, when will it reach a city 180 km away?"*  
- **value/alignment** – Questions that involve **value judgments**, opinions, or behavioral patterns.  
    - Example: *"Is it ethical to use AI in hiring decisions?"*  
    - Example: *"What are the social impacts of remote work?"*  

\#\#\# **target Description**  
- **target\_type** indicates the country or cultural region that the query is focusing on. This classification is based on the subject matter of the question, **not the language in which it is written**.  
- Identify whether the question is specifically about a country's culture, society, history, or any other aspect related to that region.  
- If there is no corresponding value, you can add it.  

\#\#\#\# **target Options:**  
- **general** – A general target without a specific cultural or national focus.  
- **ko** – Targeting **Korea**.  
- **us** – Targeting **the United States**.  
- (중략)  

- subject\_type represents the knowledge domain or reasoning field needed to answer the prompt. Identify the content of the query and select one or more of the following values. If there is no matching category, respond with 'misc'.  
- Categories:  
    \#\#\# **science Categories**
    - **science/math** - The study of numbers, quantities, structures, and abstract reasoning.
    - **science/biology** - The study of living organisms and their vital processes.
    - (중략)
    - **science/microbiology** - The study of microorganisms and pathogens. (가정된 세부 카테고리)

Now, present the corresponding categories of following data in json format.  
Data:
{
  "query": "What causes vomiting and diarrhea in a child after eating kimbap?",
  "answer": "Likely bacterial infection such as Salmonella or E. coli.",
  "category": null
}

---

{
  "subject\_type": ["science/biology", "science/microbiology"],
  "task\_type": "knowledge",
  "target\_type": "ko"
}

\end{tcolorbox}

\subsection{Reliability of Automated Categorization}
\label{appendix:reliabilty_cats}
\subsubsection{Categorization Accuracy}
\label{appendix:reliability_cats_acc}
\begin{wraptable}{r}{0.4\textwidth}
\vspace{-4.5mm}
\caption{Accuracy of fine-tuned categorizer on Qwen-2.5-7b.}
\label{tab:cat_acc}

    \begin{tabular}{lr}
    \toprule
    {\textbf{Sample-level Attribute}} & \textbf{Accuracy} \\ \midrule
    Subject & 0.871    \\
    Skill & 0.967 \\
    {Cultural-specificity} & 0.986 \\ \bottomrule
    \end{tabular}
    \end{wraptable}

Table~\ref{tab:cat_acc} shows the performance of the categorizer across three classification types. To evaluate categorization, we use existing datasets that already have fine-grained categories, which are similar to or more confined than our taxonomy (7 benchmarks and 307 instances in English/Korean). 
For the subject, we consider it correct if the designated main category is included. This criterion is necessary because source datasets typically provide only one canonical label, whereas BenchHub permits multiple applicable labels. 
Since the existing datasets often do not contain all three taxonomies we've defined, we compare the available category types in the evaluation. For example, MMLU defines only the subject category, while MATH datasets fall under the 'reasoning' (skill type). Both subject and skill show high performance, while the target (to determine which culture/country the query is related to) scores lower. The target has a large number of possible culture options (all countries), making it difficult to train for all options. Therefore, instead of segmenting the target by country/culture, we enhance reliability by separating it into two labels: general and cultural, based on the initial label from categorize. 

\paragraph{Human re-validation of the main evaluation instances.}
We conduct an additional human validation using instances sampled from the actual English and Korean evaluation pools used in our main analyses. To deliberately evaluate a more challenging setting than our original validation, we focus on instances assigned two or more subject labels and sample 232 instances in total (En: 116, Ko: 106). An annotator independently identifies both incorrect assigned labels and missing applicable labels. Against these human-validated annotations, the categorizer achieves F1 scores of 0.835 at the coarse level and 0.754 at the fine-grained level.
 
\paragraph{Observed systematic error patterns.}
In English, false negatives frequently involve life sciences and atmospheric sciences, while false positives often involve daily life and IT; the most common confusions are Culture $\rightarrow$ HASS and Culture $\rightarrow$ Science. In Korean, false negatives frequently involve language, administration, and economics, while false positives often involve work life and tradition; the dominant confusions are Technology $\rightarrow$ HASS and Culture $\rightarrow$ HASS.

\subsubsection{{Influence of Categorization Accuracy on Model Evaluation}}
We examine and discuss the influence of categorization accuracy on model evaluation outcomes in \framework.
To quantify and simulate the categorizing errors, we conduct an ablation study in which the categorization error rate is systematically varied and controlled.
Following the experimental setups described in \S\ref{sec:sq3}, we employ a stratified sampling strategy to preserve dataset-level balance across categories.
We introduce a controlled \textit{corruption rate}, which denotes the proportion of misclassified samples in the test set.
We increment the corruption rate from 0.0\% to 10.0\% in 0.5\% steps.
For each corruption level, we perform 50 independent simulation runs to ensure statistical robustness.
We compare the model rankings obtained from the corrupted test sets to the baseline rankings derived from the original, uncorrupted set.

We demonstrate that categorization errors up to 1.5\% yield negligible disruption to model rankings, confirmed by Spearman's rank correlation coefficient and Wilcoxon Signed-Rank test.
This finding suggests a notable resilience of the evaluation framework to minor categorization inaccuracies.
It is noteworthy that this robustness extends beyond simple misclassification scenarios to dynamic, real-world settings tailored for users.
Introducing a small fraction of samples comprising undefined categories is less likely to cause significant shifts in model rankings.
Moreover, the categorizer can be incrementally updated and improved through continual learning, ensuring ongoing adaptation and maintenance of \framework pipeline among evolving benchmarks.

\paragraph{Robustness to random and systematic categorization errors.}
Based on the categorizer error patterns we observe in Appendix~\ref{appendix:reliability_cats_acc}, we simulate both (1) random label noise and (2) language-specific systematic errors reflecting the observed false-positive, false-negative, and category-confusion biases. We vary the perturbation rate from 0\% to 30\% in five-percentage-point increments and average the results across random seeds. Using these, we evaluate the robustness of Observations 1--3, which rely primarily on the subject-type labels.
 
\paragraph{Obs. 1: Advertised scope.}
The proportion of instances outside the advertised scope remains close to the original estimate of 65.1\% and is more stable under systematic errors than under random errors, reaching approximately 69\% even at a 30\% perturbation rate. Thus, both the direction and the overall magnitude of the finding are preserved.
 
\begin{table}[ht!]
\centering
\small
\caption{Proportion (\%) of instances outside the advertised scope under random and systematic label noise.}
\label{tab:noise_obs1}
\setlength{\tabcolsep}{4pt}
\begin{tabular}{lccccccc}
\toprule
Noise rate & 0\% & 5\% & 10\% & 15\% & 20\% & 25\% & 30\% \\
\midrule
Random     & 65.07 & 66.07{\scriptsize$\pm$0.02} & 67.11{\scriptsize$\pm$0.02} & 68.11{\scriptsize$\pm$0.01} & 69.11{\scriptsize$\pm$0.01} & 70.10{\scriptsize$\pm$0.03} & 71.08{\scriptsize$\pm$0.02} \\
Systematic & 65.07 & 65.70{\scriptsize$\pm$0.03} & 66.33{\scriptsize$\pm$0.02} & 67.01{\scriptsize$\pm$0.04} & 67.65{\scriptsize$\pm$0.04} & 68.37{\scriptsize$\pm$0.03} & 68.97{\scriptsize$\pm$0.08} \\
\bottomrule
\end{tabular}

\end{table}
 
\paragraph{Obs. 2: Composition of holistic benchmarks and multilingual benchmarks.}
The dominant subject domain remains unchanged for all seven holistic or multilingual benchmarks under both random and systematic noise, even at a 30\% perturbation rate. When comparing the relative ordering of all subject-domain pairs, 92.3\% of the pairwise orderings are still preserved at the 30\% perturbation rate. The few changes involve only one or two lower-ranked domains whose original proportions are already very similar. For example, Science and HASS in KMMLU differ by only about 1\%p before perturbation. Overall, categorization noise makes the subject distributions more even but does not change the main composition pattern of the benchmarks.
 
\paragraph{Obs. 3: Model ranking instability across dimensions.}
Ranking stability under varying error rates. We measure ranking stability using Spearman's rank correlation coefficient ($\rho$), where a value of 1 indicates identical model rankings and lower values indicate larger ranking changes. Averaged across the six subject domains, ($\rho$) remains high as the perturbation rate increases. Even at 30\% noise, the average correlation is 0.956 under random perturbations and 0.960 under systematic perturbations, indicating that the overall model rankings across subject categories are largely preserved (Table~\ref{tab:noise_obs3}).
 
Pair-reversal rate (\%) under random and systematic subject-label perturbations. Although the rate decreases as the perturbation rate increases, substantial pair reversals remain at 30\% noise (15.6\% for random $\sim$ 18.8\% for systematic).
 
\begin{table}[ht!]
\centering
\small
\caption{Spearman's $\rho$ between original and perturbed model rankings, averaged across the six subject domains.}
\label{tab:noise_obs3}
\begin{tabular}{lccccccc}
\toprule
Noise type & 0\% & 5\% & 10\% & 15\% & 20\% & 25\% & 30\% \\
\midrule
Random     & 1.000 & 0.995 & 0.991 & 0.979 & 0.975 & 0.964 & 0.956 \\
Systematic & 1.000 & 0.997 & 0.988 & 0.980 & 0.974 & 0.970 & 0.960 \\
\bottomrule
\end{tabular}

\end{table}

\subsubsection{Enhancing Categorization Robustness}
\label{appendix:enhancing_cats}
The classification process of \framework currently relies on a model trained with Qwen-2.5-7B-instruct, which may introduce potential model-specific bias when relying on a single classifier. As a possible direction for improving categorization, we additionally train classifiers using Llama-3.1-8B and Mistral-7B-v0.1 with the same training data and procedure. We then construct a multi-agent classification system in which the predictions from all three models (Qwen, Llama, and Mistral) are aggregated via majority voting. This system achieves a 2.4\%p increase in agreement with human labels compared to Qwen-2.5-7B-instruct alone. Among the individual classifiers, Qwen-2.5-7B achieves the best standalone performance, and we expect that leveraging larger foundation models will further amplify the benefits of majority voting.  

While majority voting improves robustness, it also triples the computational cost for training and inference. As an alternative, we implement a confidence-based hybrid approach: majority voting is invoked only when the classifier's confidence (measured by average logit probability) falls below a threshold of -0.04. This method enhances agreement by 1.4\%p while substantially reducing the additional cost, thereby offering a practical trade-off between robustness and efficiency.

\subsubsection{Expanding Categorizer to New Category during Inference}

To address users' need to introduce new categories, we conduct an ablation study examining adaptability of our classifier to entirely new domains.
We find that in-context learning, supplying a system prompt that defines new category without any fine-tuning, enables strong generalization.
As a case study, we introduce a hypothetical domain \emph{Magic (supernatural)}, consisting of one coarse-grained and eight fine-grained categories.
Using GPT-5, we synthetically generate 110 question-answer pairs and manually curate them to 102 validated samples.
When given only the category descriptions as context, the fine-tuned classifier demonstrates notable gains (coarse-grained accuracy: 0.000\textrightarrow0.941; fine-grained accuracy: 0.000\textrightarrow0.823).
An example instance from this domain is shown below:
\begin{tcolorbox}[breakable, enhanced, top=1pt, left=1pt, right=1pt, bottom=1pt]
Q: What does the spell Lumora Spiralis do? \\
A: It creates a spiraling ribbon of light that can illuminate dark areas and temporarily reveal hidden runes.
\end{tcolorbox}

\subsubsection{Categorizing Open-ended User Intent}

\framework provides a flexible, intent-driven evaluation framework that operates without requiring additional configuration from users.
To support open-ended evaluation scenarios (\eg ``\textit{I want to build and evaluate AI assistant used in Korean math class.}'', \framework incorporates an automated intent interpretation module based on GPT-4, which translates free-form natural language instructions into the corresponding categories within our taxonomy.
Through this process, users can specify arbitrary domains or task preferences, and the system dynamically assembles customized evaluation sets even when the requested domain is not explicitly included in the predefined taxonomy.

\subsection{Multilingual Expansion of \framework}
\label{appendix:benchhub-multilingual}
While we focus on two languages (\ie Korean and English), we highlight that \framework is a language-agnostic, flexible framework that can be easily extended to other languages.
To empirically guide this extension, we present \texttt{BenchHub-Multi-Cat-7B}, a multilingual categorizer supporting 10 languages---English (En); 3 high-resource (Arabic (Ar), German (De), Dutch (Nl)); 3 mid-resource (Indonesian (Id), Korean (Ko), Ukrainian (Uk)); 3 low-resource (Swahili (Sw), Nepali (Ne), Kyrgyz (Ky)).
Our multilingual categorizer achieves an average accuracy of 77.5\% on fine-grained subject categorizations for unseen, out-of-domain data.
Furthermore, we introduce \framework-multilingual, which extends our benchmark suite to a total of 10 languages consisting of 13 datasets and 444,402 samples.
We hope \framework-multilingual will serve as a foundational step for reliable LLM evaluations in non-English languages.
The details of the training procedure and the datasets for each language are provided in Appendix~\ref{appendix:multilingual_expansion}.

\label{appendix:multilingual_expansion}
\subsubsection{Multilingual Categorizer}
\begin{table}[t]
\vspace{-4.5mm}
\centering
\small
\caption{Categorizer accuracy on G-MMLU and M-MMLU.}
\label{tab:m_cat_acc}
\begin{tabular}{lcc}
\toprule
\textbf{Language} & \textbf{G-MMLU} & \textbf{M-MMLU} \\
\midrule
Arabic     & 0.765 & 0.767 \\
German     & 0.789 & 0.833 \\
Indonesian & 0.800 & 0.808 \\
Kyrgyz     & 0.681 & --    \\
Nepali     & 0.709 & --    \\
Dutch      & 0.804 & --    \\
Swahili    & 0.614 & 0.653 \\
Ukrainian  & 0.765 & --    \\
\bottomrule
\end{tabular}
\vspace{-4mm}
\end{table}
We fine-tune Qwen-2.5-7B-instruct on ten languages (English; three high-resource languages: Arabic, German, Dutch; three mid-resource languages: Indonesian, Korean, Ukrainian; and three low-resource languages: Swahili, Nepali, Kyrgyz). For the training dataset, we use 20,000 samples from Global MMLU~\citep{singh-etal-2025-global}, with 2,000 samples per language. Since Global MMLU provides human-validated fine-grained subject categories, we adopt these categories while mapping them to our taxonomy. The training method and configurations follow those used in the categorizer for Korean and English (Appendix~\ref{appendix:categorizer_mono}).

We validate the categorizer on 2,850 Global MMLU samples (285 samples per language) that were not used during fine-tuning (in-domain), and on 1,225 Multilingual MMLU samples (245 samples per language) from outside the training distribution (out-of-domain). Our model achieves 75.3\% accuracy in-domain and 77.5\% accuracy out-of-domain for fine-grained subject categorization. Table~\ref{tab:m_cat_acc} reports detailed results for both evaluation settings. Blank cells indicate that M-MMLU does not support the corresponding language.

\section{Additional Dataset Analysis}
\subsection{Lists of datasets Used}
Table~\ref{tab:benchmarks_stats} summarizes the English and Korean datasets included in \framework, while Table~\ref{tab:benchmarks_stats_multilingual} reports the additional multilingual datasets included in \framework.

\subsection{Additional Statistics}

\begin{figure}[htb!]
    \centering
    \begin{subfigure}[b]{\linewidth}
        \centering\small\includegraphics[width=1.0\linewidth]{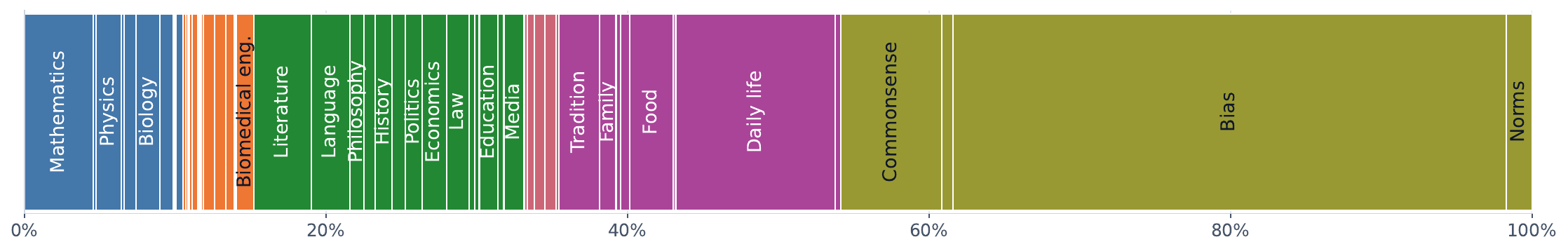}
        \caption{English}\label{fig:pie_data_all_en}
    \end{subfigure}
    
    \begin{subfigure}[b]{\linewidth}
        \centering\small\includegraphics[width=1.0\linewidth]{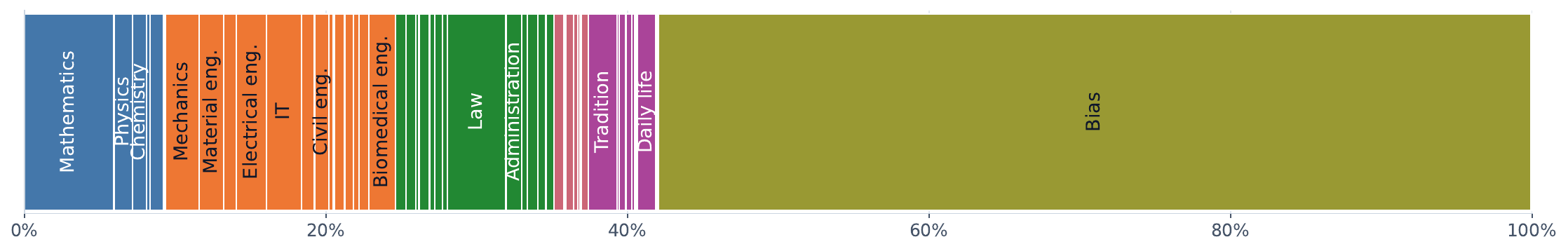}
        \caption{Korean}\label{fig:pie_data_all_ko}
    \end{subfigure}

    \caption{Fine-grained subject distribution of all datasets used in \framework.}
    \label{fig:pie_data_dist}
\end{figure}

\begin{figure}[t]
    \centering

    \begin{subfigure}[t]{\textwidth}
        \centering
        \includegraphics[width=\linewidth]{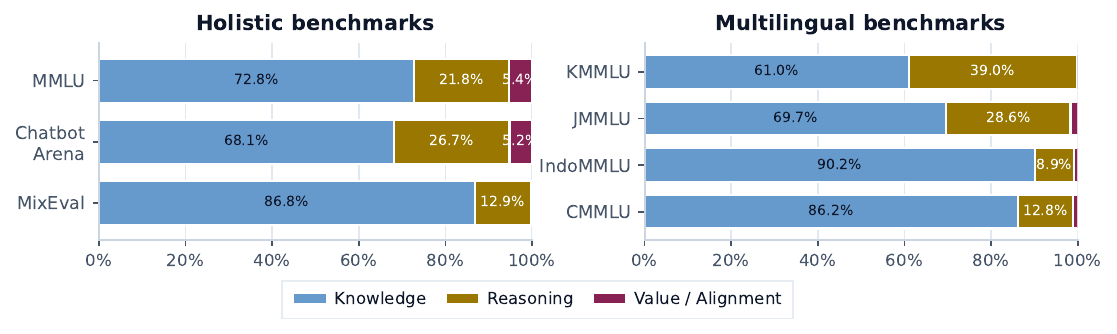}
        \caption{Skill}
        \label{fig:mmlu_dist_skill}
    \end{subfigure}
    \hfill
    \begin{subfigure}[t]{\textwidth}
        \centering
        \includegraphics[width=\linewidth]{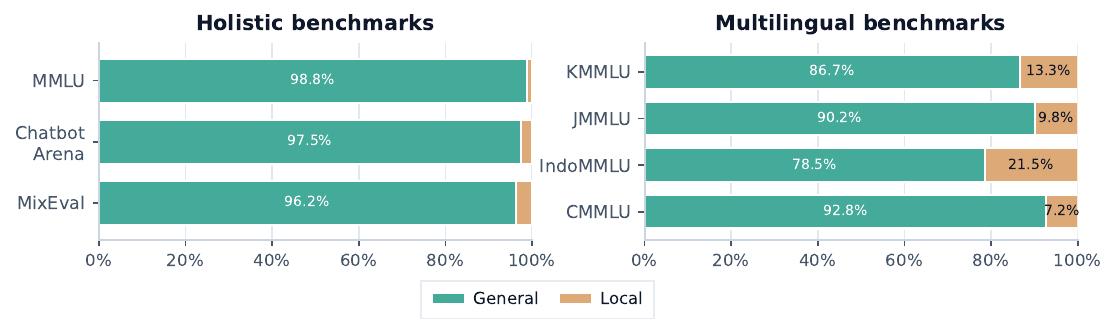}
        \caption{Cultural specificity}
        \label{fig:mmlu_dist_target}
    \end{subfigure}

    \caption{Skill and cultural specificity distributions of the Holistic benchmarks and MMLU series in English, Korean, Japanese, Indonesian, and Chinese.}
    \label{fig:mmlu_dist_bar_additional}
\end{figure}

\begin{figure}[htb!]
    \centering

        \includegraphics[height=4.6cm]{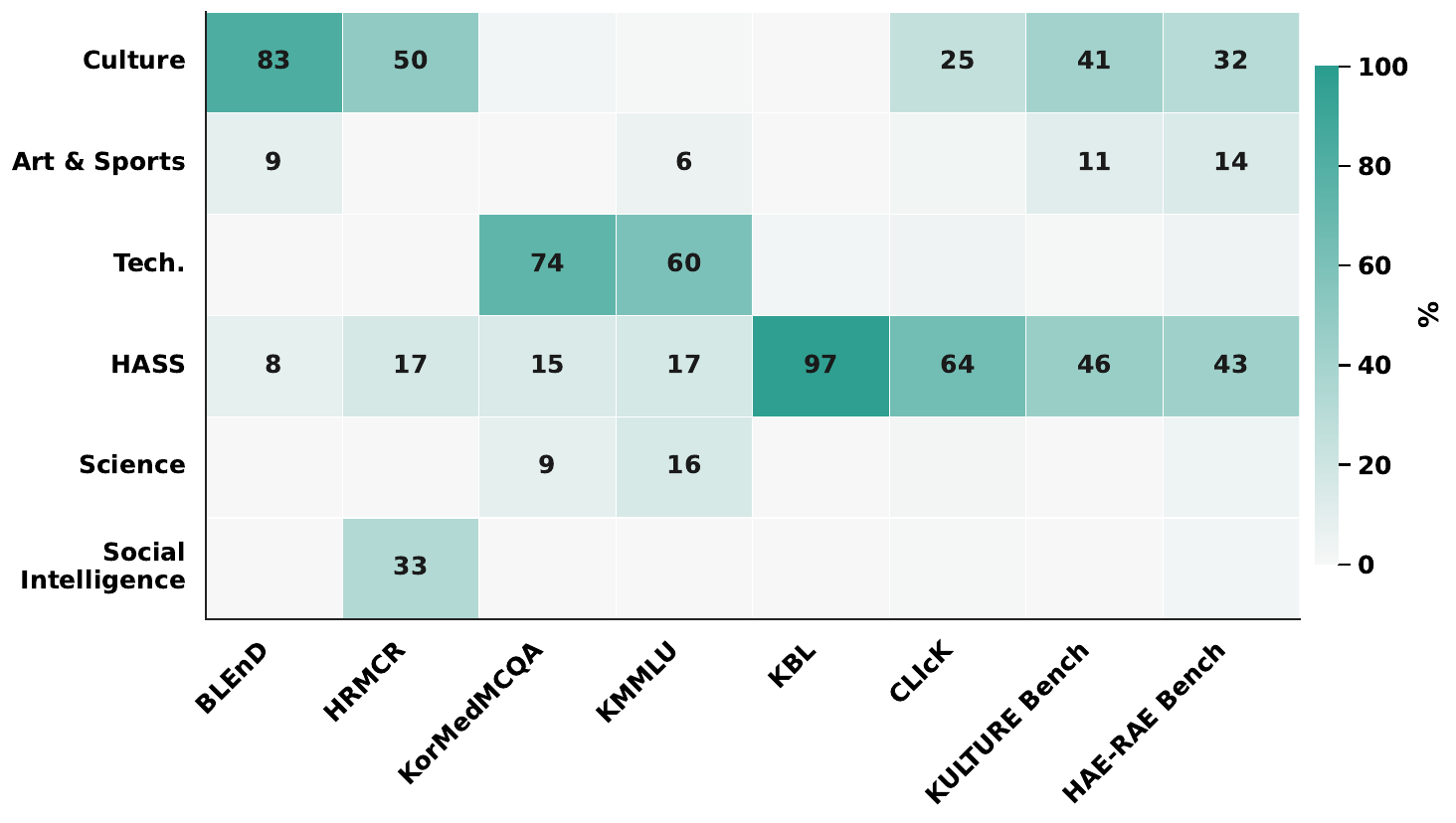}
        \caption{Subject distribution by benchmarks in korean benchmarks (\%, $p(\text{subject} | \text{benchmark})$).}
        \label{fig:ko_bh_coverage}
   
\end{figure}

Figure~\ref{fig:pie_data_dist} shows the fine-grained subject-level distribution in \framework. Figure~\ref{fig:ko_bh_coverage} presents the subject distribution analysis of Korean benchmarks for Obs.~\ref{obs:obs1}.
Figure~\ref{fig:mmlu_dist_bar_additional} presents the skill and target-category distributions of the holistic multilingual benchmarks analyzed in Obs.~\ref{obs:obs2}.

\subsection{Cross-Benchmark Overlap}
\label{appendix:duplicates}

\begin{figure}[t]
    \centering

    \begin{subfigure}[t]{\textwidth}
        \centering
        \includegraphics[width=\linewidth]{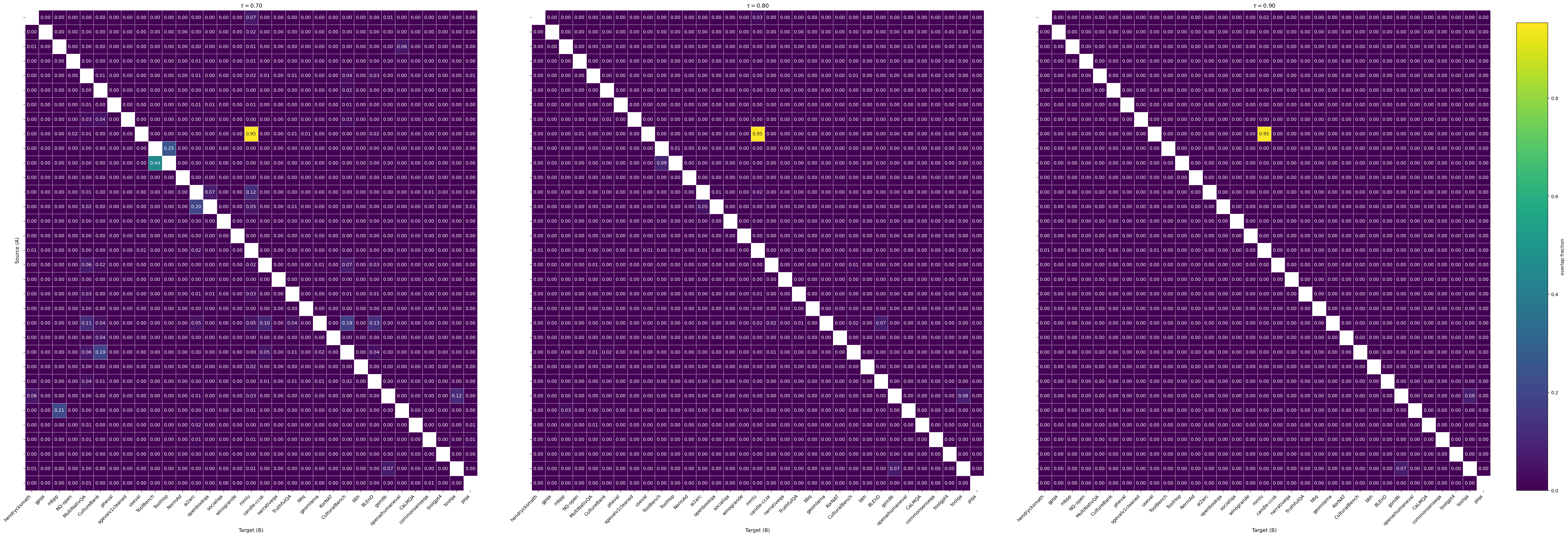}
        \caption{English}
        \label{fig:heatmap_en}
    \end{subfigure}
    \hfill
    \begin{subfigure}[t]{\textwidth}
        \centering
        \includegraphics[width=\linewidth]{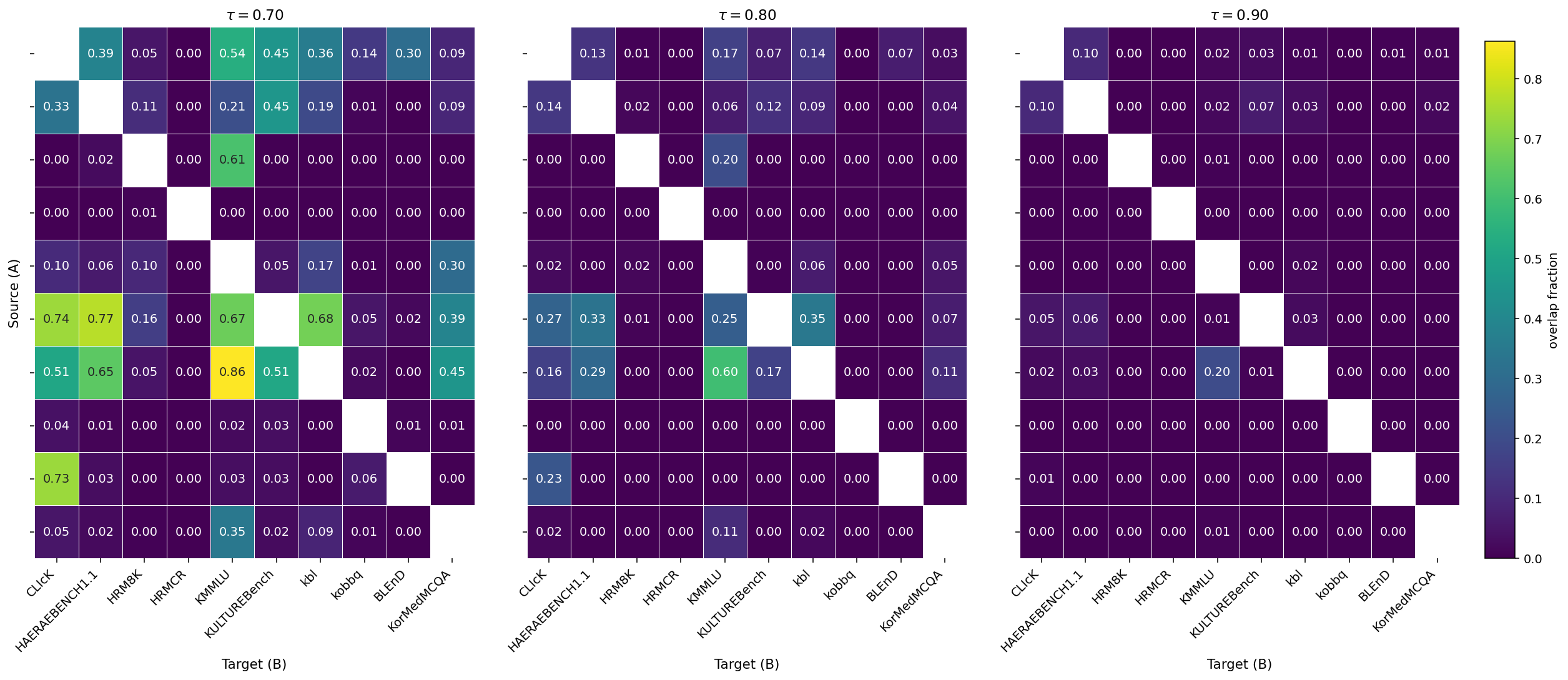}
        \caption{Korean}
        \label{fig:heatmap_ko}
    \end{subfigure}

    \caption{Cross-benchmark overlap at cosine thresholds $\tau \in \{0{.}70,\, 0{.}80,\, 0{.}90\}$.
    Rows index the source benchmark $A$; columns index the target benchmark $B$.
    Cell $(A,B)$ reports the fraction of items in $A$ whose nearest neighbor in $B$
    has cosine similarity at least $\tau$.}
    \label{fig:overlap_heatmaps}
\end{figure}

We embed every \framework{} instance with a multilingual sentence
encoder---\texttt{all-mpnet-base-v2} for English and
\texttt{paraphrase-multilingual-mpnet-base-v2} for Korean---and use the
\texttt{subject\_type} taxonomy as a topic label, projected to its
top-level domain $D$ (Science, HASS, Culture, Tech., Art~\&~Sports,
Social~Intelligence). For every ordered benchmark pair $(A,B)$ and
threshold $\tau\!\in\!\{0.7,0.8,0.9\}$ we compute the
\emph{topic-conditioned twin rate}
\[
\mathrm{leak}_{A,B}(\tau,D)\;=\;
\frac{\bigl|\{i\in A\cap D:\;\max_{j\in B\cap D}\cos(e_i,e_j)\geq \tau\}\bigr|}
     {|A\cap D|},
\]
\ie the fraction of $A$'s domain-$D$ instances that have a
near-duplicate inside $B$'s domain-$D$ instances. Restricting to the
same domain isolates content overlap from the trivial signal that two
benchmarks happen to belong to the same broad area. 
We report results at  $\tau\in\{0.7, 0.8, 0.9\}$ in Figure~\ref{fig:overlap_heatmaps}.

A surprising portion of the cross-benchmark leak is not paraphrased
overlap but byte-identical re-use. Counting unique cross-benchmark
instance pairs at $\cos\geq 0.9$ (a threshold at which embeddings can
be treated as nearly duplicates), English BenchHub contains $\mathbf{285}$
such pairs; the count is essentially flat from $\cos\geq 0.98$ down
to $\cos\geq 1.00$ (exact textual match), confirming these are not
near-paraphrases. They are concentrated in three benchmark pairs:
$\sim\!\!100$ pairs between \texttt{cais/mmlu} and
\texttt{SeaEval/us\_eval} (US civics/knowledge questions),
$\sim\!\!100$ between \texttt{openai/gsm8k} and \texttt{toolqa} (math
word problems), and $\sim\!\!80$ between
\texttt{EleutherAI/hendrycks\_math} and \texttt{cais/mmlu}'s
mathematics partition (\eg \emph{``If $a{=}7$ and $b{=}13$, what is
the number of even positive integers less than $ab$?''} appears
verbatim in both). Korean BenchHub shows the same effect at the same
order of magnitude---$103$ unique pairs at $\cos\geq 0.99$, dominated
by \texttt{lbox/kbl} $\leftrightarrow$ \texttt{HAERAE-HUB/KMMLU}
($\sim\!\!90$ pairs). Whatever these pairs measure, scoring them
twice is double-counting; for any topic-level analysis they should be
deduplicated, not aggregated.

\clearpage
\begin{table}[htb!]
\centering
\caption{Benchmarks Included in \framework.}
\label{tab:benchmarks_stats}
\resizebox{\linewidth}{!}{
\begin{tabular}{llclrl}
\toprule
\textbf{Dataset} & \textbf{Reference} & \textbf{Cultural Specificity} & \textbf{Lang.} & \textbf{\# of Samples} & \textbf{License} \\
\midrule
ARC & \cite{clark2018think} & \cellcolor{blue!10}General & EN & 3,548 & cc-by-sa 4.0 \\
SocialIQA & \cite{sap-etal-2019-social} & \cellcolor{blue!10}General & EN & 1,954 & CC-BY-4.0 \\
WinoGrande & \cite{winogrande2021sakaguchi} & \cellcolor{blue!10}General & EN & 1,767 & Apache-2.0 \\
Natural Questions (open) & \cite{kwiatkowski-etal-2019-natural} & \cellcolor{blue!10}General & EN & 1,769 & Apache-2.0 \\
NarrativeQA & \cite{kocisky-etal-2018-narrativeqa} & \cellcolor{blue!10}General & EN & 10,557 & Apache-2.0 \\
TruthfulQA & \cite{lin-etal-2022-truthfulqa} & \cellcolor{blue!10}General & EN & 817 & Apache-2.0 \\
Open-BookQA & \cite{mihaylov-etal-2018-suit} & \cellcolor{blue!10}General & EN & 1,000 & Apache-2.0 \\
MMLU & \cite{hendrycks2021measuring} & \cellcolor{blue!10}General & EN & 14,042 & MIT \\
BBQ & \cite{parrish-etal-2022-bbq} & \cellcolor{blue!10}General & EN & 58,492 & cc-by-4.0 \\
PIQA & \cite{bist2020piqa} & \cellcolor{blue!10}General & EN & 3,084 & Apache-2.0 \\
CommonsenseQA & \cite{talmor-etal-2019-commonsenseqa} & \cellcolor{blue!10}General & EN & 1,140 & MIT \\
BBH & \cite{suzgun-etal-2023-challenging} & \cellcolor{blue!10}General & EN & 6,261 & MIT \\
MATH & \cite{hendrycks2021math} & \cellcolor{blue!10}General & EN & 4,521 & MIT \\
HumanEval & \cite{chen2021evaluating} & \cellcolor{blue!10}General & EN & 164 & MIT \\
MBPP & \cite{austin2021program} & \cellcolor{blue!10}General & EN & 974 & cc-by-4.0 \\
GSM8k & \cite{cobbe2021training} & \cellcolor{blue!10}General & EN & 1,319 & MIT \\
GPQA & \cite{rein2024gpqa} & \cellcolor{blue!10}General & EN & 1,191 & cc-by-4.0 \\
ToolHop & \cite{ye-etal-2025-toolhop} & \cellcolor{blue!10}General & EN &996  & cc-by-4.0 \\
ToolQA & \cite{zhuang-etal-2023-toolqa} & \cellcolor{blue!10}General & EN & 1,545 & Apache-2.0 \\
ToolBench& \cite{qin-etal-2023-toolllm} & \cellcolor{blue!10}General & EN &  77,120 & Apache-2.0 \\
GPT4Tools & \cite{yang-etal-2023-gpt4tools} & \cellcolor{blue!10}General & EN & 13,070 & Apache-2.0 \\
MultiNativQA & \cite{hasan-etal-2025-nativqa} & \cellcolor{orange!15}Local & EN & 3,435 & cc-by-nc-sa-4.0 \\
CulturalBench & \cite{chiu-etal-2025-culturalbench} & \cellcolor{orange!15}Local & EN & 6,134 & cc-by-4.0 \\
SeaEval & \cite{wang-etal-2024-seaeval} & \cellcolor{orange!15}Local & EN & 275 & cc-by-nc-4.0 \\
CANDLE CCSK & \cite{nguyen2023extracting} & \cellcolor{orange!15}Local & EN & 500 & cc-by-4.0 \\
GeoMLAMA & \cite{yin-etal-2022-geomlama} & \cellcolor{orange!15}Local & EN & 124 & unknown \\
NormAd & \cite{rao-etal-2025-normad} & \cellcolor{orange!15}Local & EN & 7,899 & cc-by-4.0 \\
CultureBank & \cite{shi-etal-2024-culturebank} & \cellcolor{orange!15}Local & EN & 22,990 & MIT \\
CaLMQA & \cite{arora-etal-2025-calmqa} & \cellcolor{orange!15}Local & EN, KO & 96 & MIT \\
BLEnD & \cite{myung2024blend} & \cellcolor{orange!15}Local & EN & 4,132 & cc-by-sa-4.0 \\
BLEnD & \cite{myung2024blend} & \cellcolor{orange!15}Local & KO & 1,000 & cc-by-sa-4.0 \\
KorNAT & \cite{lee-etal-2024-kornat} & \cellcolor{orange!15}Local & EN & 24 & cc-by-nc-2.0 \\
KBL & \cite{kimyeeun-etal-2024-developing} & \cellcolor{blue!10}General & KO & 3,304 & cc-by-nc-4.0 \\
KorMedMCQA & \cite{kweon2024kormedmcqa} & \cellcolor{blue!10}General & KO & 3,009 & cc-by-nc-2.0 \\
KMMLU & \cite{son-etal-2025-kmmlu} & \cellcolor{blue!10}General & KO & 30,499 & cc-by-nd-4.0 \\
HRM8K & \cite{ko-etal-2025-understand} & \cellcolor{blue!10}General & KO & 8,011 & MIT \\
KoBBQ & \cite{jin-etal-2024-kobbq} & \cellcolor{orange!15}Local & KO & 81,128 & MIT \\
KULTURE Bench & \cite{wang-etal-2024-kulture} & \cellcolor{orange!15}Local & KO & 3,584 & Apache-2.0 \\
HAE-RAE Bench & \cite{son-etal-2024-hae} & \cellcolor{orange!15}Local & KO & 4,900 & cc-by-nc-nd-4.0 \\
CLIcK & \cite{kim-etal-2024-click} & \cellcolor{orange!15}Local & KO & 1,995 & cc-by-nd-4.0 \\
HRMCR & \cite{son-etal-2025-multi} & \cellcolor{orange!15}Local & KO & 100 & Apache-2.0 \\
KoSBi & \cite{lee-etal-2023-kosbi} & \cellcolor{orange!15}Local & KO & 6,801 & MIT \\
\bottomrule
\end{tabular}
}
\end{table}

\begin{table}[htb!]
\small
\centering

\caption{Benchmarks Included in \framework-multilingual.}
\label{tab:benchmarks_stats_multilingual}
\resizebox{\linewidth}{!}{
\begin{tabular}{llcrl}
\toprule
\textbf{Dataset} & \textbf{Reference} & \textbf{Cultural Specificity} & \textbf{\# of Samples} & \textbf{License} \\
\midrule

\multicolumn{5}{l}{\textbf{Language: AR}} \\
G-MMLU & \cite{singh-etal-2025-global} & General/Local & 14,042 & apache-2.0 \\
ArabLegalEval & \cite{hijazi-etal-2024-arablegaleval} & Local & 15,311 & - \\
ArabicMMLU & \cite{koto-etal-2024-arabicmmlu} & General/Local & 14,455 & cc-by-nc-sa-4.0 \\ \midrule

\multicolumn{5}{l}{\textbf{Language: DE}} \\
G-MMLU & \cite{singh-etal-2025-global} & General/Local & 14,042 & apache-2.0 \\
GermanQUAD & \cite{pfister-hotho-2024-supergleber} & General & 2,204 & cc-by-4.0 \\
MLQA & \cite{pfister-hotho-2024-supergleber} & General & 4,517 & cc-by-sa3.0 \\ \midrule

\multicolumn{5}{l}{\textbf{Language: NL}} \\
G-MMLU & \cite{singh-etal-2025-global} & General/Local & 14,042 & apache-2.0 \\ \midrule

\multicolumn{5}{l}{\textbf{Language: ID}} \\
G-MMLU & \cite{singh-etal-2025-global} & General/Local & 14,042 & apache-2.0 \\
Eli5-indo & \href{https://huggingface.co/datasets/indonesian-nlp/eli5_id}{indonesian-nlp/eli5\_id} & General & 245,274 & - \\
facQA & \cite{lovenia-etal-2024-seacrowd} & General & 1,564 & cc-by-sa-4.0 \\
idkmrc & \cite{putri-oh-2022-idk} & Local & 1,198 & cc-by-sa4.0. \\
QASiNa & \cite{rizqullah-purwarianti-aji-2023-qasina} & Local & 133 & MIT. \\
TyDi QA & \cite{lovenia-etal-2024-seacrowd} & General & 4,276 & Apache-2.0 \\
xcopa & \cite{ponti-etal-2020-xcopa} & Local & 4,001 & cc-by-4.0 \\ \midrule

\multicolumn{5}{l}{\textbf{Language: UK}} \\
G-MMLU & \cite{singh-etal-2025-global} & General/Local & 14,042 & apache-2.0 \\
UA-CBT (Eval-UA-tion 1.0) & \cite{hamotskyi-etal-2024-eval} & Local & 2,129 & cc-by-4.0 \\ \midrule

\multicolumn{5}{l}{\textbf{Language: Sw}} \\
G-MMLU & \cite{singh-etal-2025-global} & General/Local & 14,042 & apache-2.0 \\ \midrule

\multicolumn{5}{l}{\textbf{Language: Ne}} \\
G-MMLU & \cite{singh-etal-2025-global} & General/Local & 14,042 & apache-2.0 \\
Winogrande-Nepali & \cite{nyachhyon-etal-2025-consolidating} & General & 8,135 & MIT \\ \midrule

\multicolumn{5}{l}{\textbf{Language: Ky}} \\
G-MMLU & \cite{singh-etal-2025-global} & General/Local & 14,042 & apache-2.0 \\
TUMLU & \cite{isbarov-etal-2025-tumlu} & Local & 785 & - \\
\bottomrule
\end{tabular}
}
\end{table}

\begin{figure}[htb!]
    \centering
    \begin{subfigure}[b]{0.45\textwidth}
        \centering
        \includegraphics[width=\linewidth]{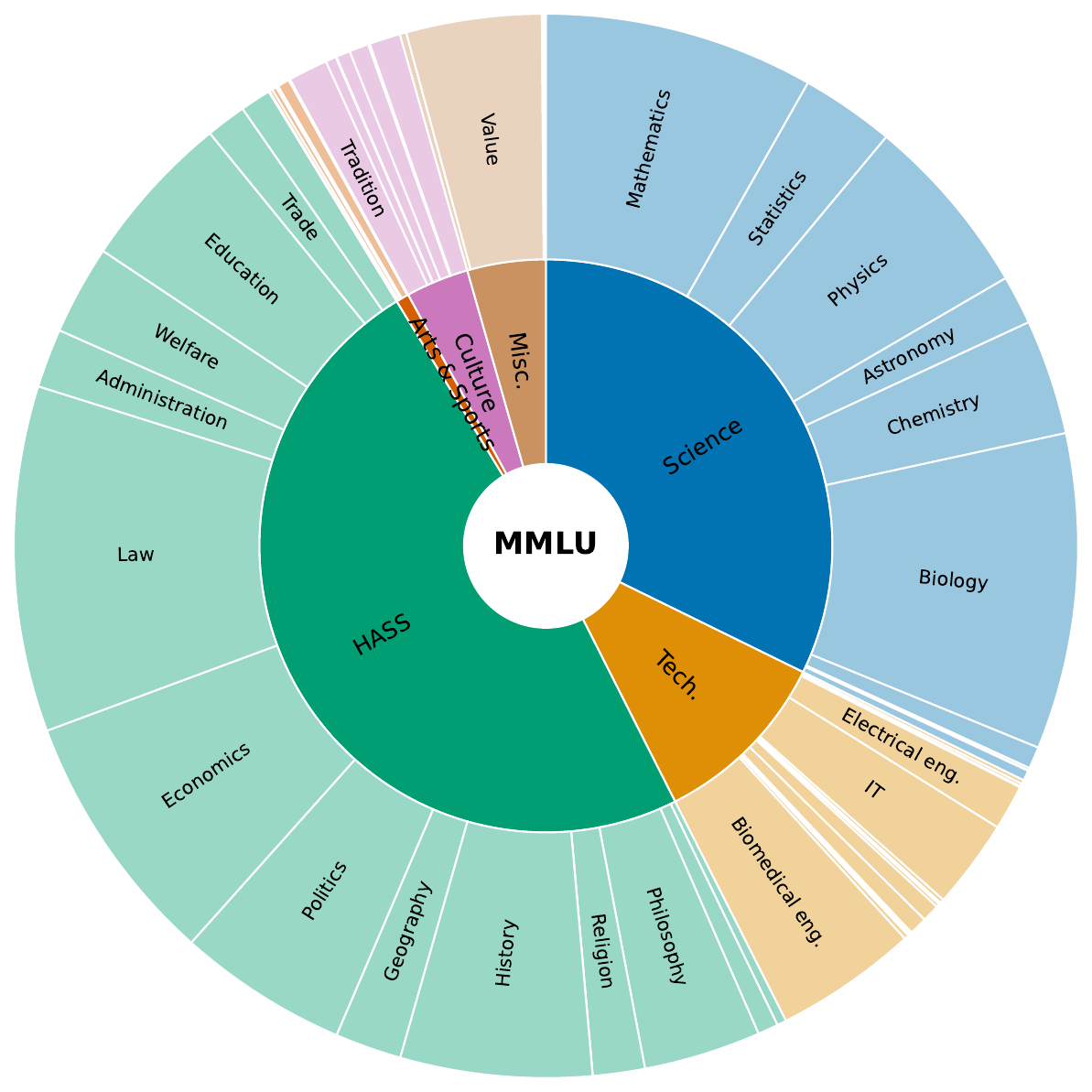}
        \caption{MMLU}\label{fig:pie_mmlu}
    \end{subfigure}
    \hfill
    \begin{subfigure}[b]{0.45\textwidth}
        \centering
        \includegraphics[width=\linewidth]{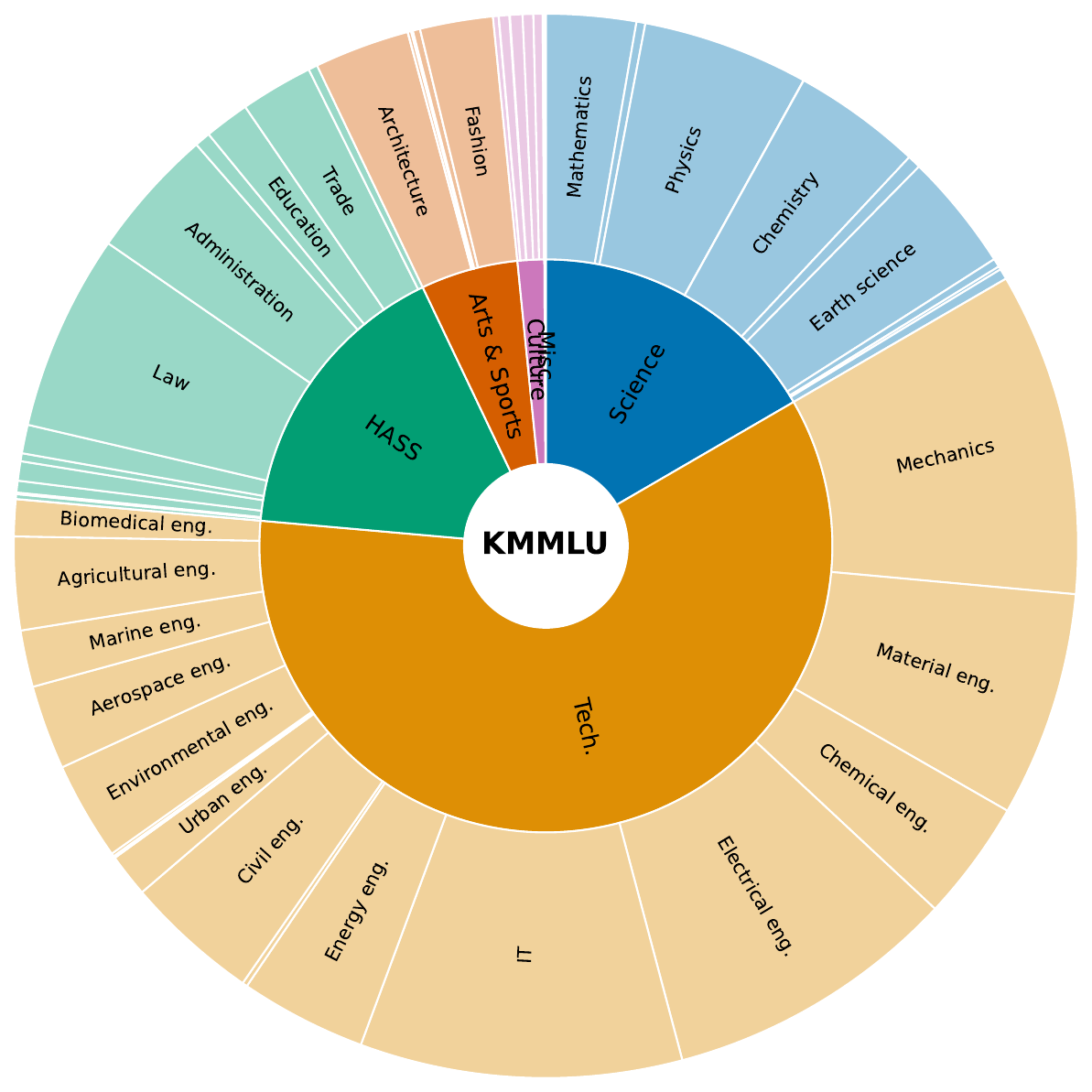}
        \caption{KMMLU}\label{fig:pie_kmmlu}
    \end{subfigure}

    \vspace{0.5em}

    \begin{subfigure}[b]{0.45\textwidth}
        \centering
        \includegraphics[width=\linewidth]{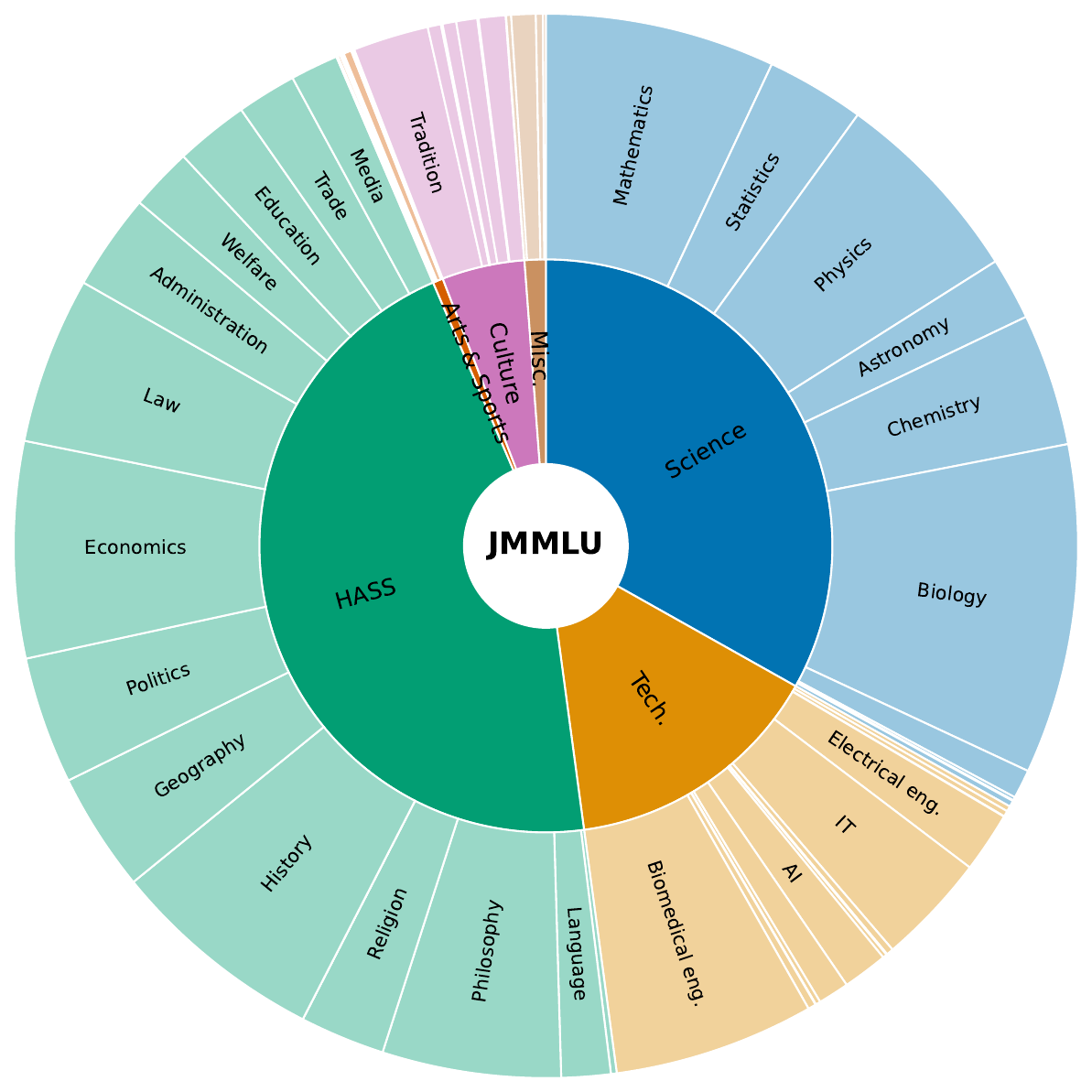}
        \caption{JMMLU}\label{fig:pie_jmmlu}
    \end{subfigure}
    \hfill
    \begin{subfigure}[b]{0.45\textwidth}
        \centering
        \includegraphics[width=\linewidth]{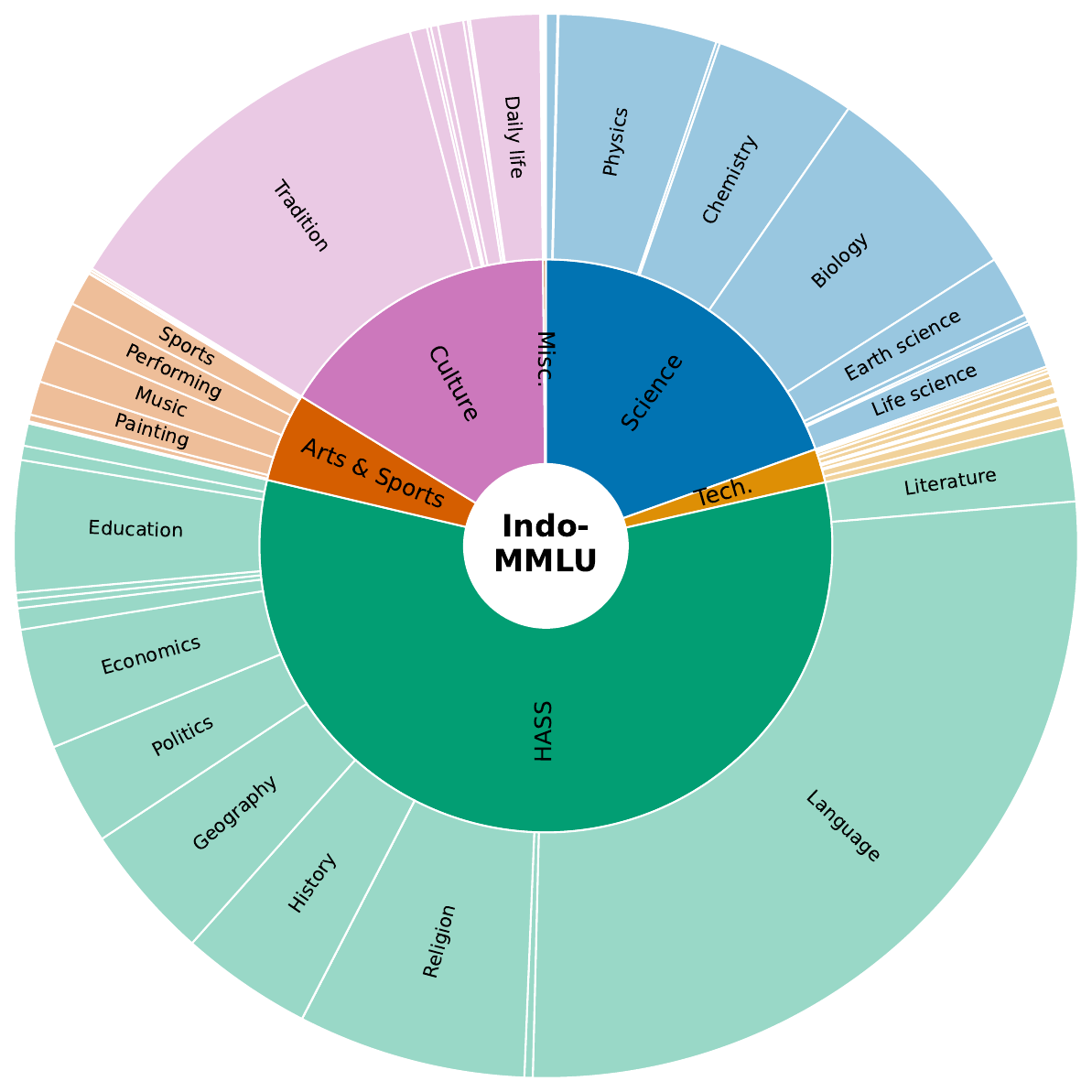}
        \caption{Indo-MMLU}\label{fig:pie_indo-mmlu}
    \end{subfigure}

    \vspace{0.5em}

    \begin{subfigure}[b]{0.45\textwidth}
        \centering
        \includegraphics[width=\linewidth]{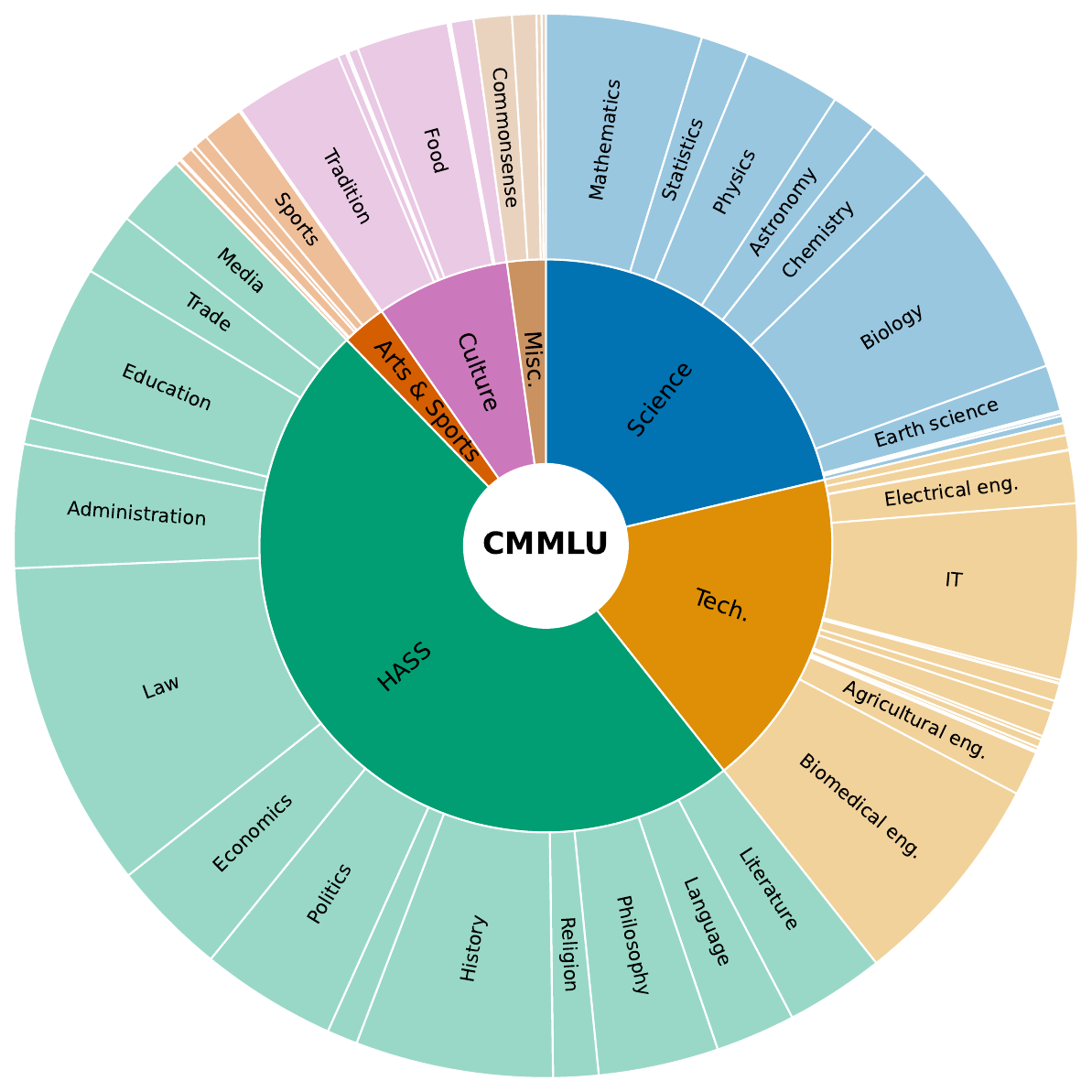}
        \caption{CMMLU}\label{fig:pie_cmmlu}
    \end{subfigure}
    
    \caption{Subject distribution of MMLU series in English, Korean, Japanese, Indonesian, and Chinese, respectively.}
    \label{fig:mmlu_dist}
\end{figure}

\clearpage
\section{Taxonomy Details}
\label{appendix:method-taxonomy}

\subsection{Problem Format}
\begin{table}[htb]
\centering
\caption{Problem format types, descriptions, and examples.}
\label{tab:problem_type}
\begin{tabularx}{\linewidth}{@{}llX X@{}}
\toprule
\multicolumn{2}{l}{\textbf{Format}} & \textbf{Description} & \textbf{Example} \\
\midrule
\multicolumn{2}{l}{\textbf{Binary}} & Two-option choice questions, typically Yes/No or True/False. & \textit{“Is the Earth flat?”} → \textit{“No”} \\
\midrule
\multicolumn{2}{l}{\textbf{Multiple-choice QA (MCQA)}} & Multiple-choice question answering format. & \textit{“What is the capital of France? (A) Paris (B) Rome (C) Berlin”}  → \textit{(A)} \\
\midrule
\multicolumn{1}{l}{\multirow{3}{*}{\textbf{Open-ended generation}}} 
 & Short-form & Short, direct answer generation. & \textit{“What is 2+2?”} → \textit{“4”} \\
 & Free-form & Extended, explanatory answer generation. & \textit{“Explain the theory of relativity.”} → \textit{“The theory of relativity, developed by Albert Einstein…”} \\
 & Open-ended & Answer generation without a definitive correct response, involving free imagination or storytelling. & \textit{“Tell a story about a journey to the moon.”} → \textit{“Once upon a time, a brave astronaut set off on a magical voyage beyond the stars…”} \\
\midrule
\multicolumn{2}{l}{\textbf{Comparison}} & Subjective or preference-based evaluation of responses, typically involving agreement, helpfulness, or safety. & \textit{“Which response is more helpful?” A: “Yes.” B: “Sure, here’s how you can do that…”} → \textit{(B)} \\
\bottomrule
\end{tabularx}
\end{table}

\subsection{{Tool Usage}}

\framework directly incorporates target entity, problem type, and tool specifications provided by each original benchmark.

\begin{table}[h]
\caption{Tool usage datasets, entity being evaluated, and problem format.}
\label{tab:tool_usage_desc}
\small
\centering
\begin{tabular}{@{}lll@{}}
\toprule
Benchmark & Entity Being Evaluated                       & Problem Format \\ \midrule
ToolHop~\cite{ye-etal-2025-toolhop}   & Final answer                         & Short-form   \\
ToolQA~\cite{zhuang-etal-2023-toolqa}    & Final answer                         & Short-form   \\
GPT4Tools~\cite{yang-etal-2023-gpt4tools} & Tool calling within output reasoning & Open-ended   \\
ToolBench~\cite{guo-etal-2024-stabletoolbench} & Tool calling                         & Short-form   \\ \bottomrule
\end{tabular}
\end{table}

\subsection{Subject}

We use 6 coarse-grained and 64 fine-grained subjects to classify samples in existing LLM evaluation benchmarks.
Table~\ref{tab:subject_type} lists the subjects and their definitions.
We finalize the subject lists by aggregating WebDewey\thinspace\footnote{\url{https://www.oclc.org/en/webdewey.html}} based on Dewey Decimal Classification (DDC) system and Korean culture-specific classification systems\thinspace\footnote{디지털집현전 (\url{https://k-knowledge.kr/guide/nkiClassifi.jsp}).}\footnote{한국민족문화대백과사전 (\url{https://encykorea.aks.ac.kr/}).}.

\begin{longtable}{p{0.17\linewidth}|p{0.2\linewidth}p{0.53\linewidth}}
\caption{Subject types and descriptions.}
\label{tab:subject_type}
\\
\toprule
\textbf{Coarse-grained} & \textbf{Fine-grained} & \textbf{Description} \\ \midrule
\multirow{10}{*}{Science} & Mathematics & The study of numbers, quantities, structures, and abstract reasoning. \\
 & Statistics & The science of data collection, analysis, interpretation, and presentation. \\
 & Physics & The study of matter, energy, and the fundamental forces of nature. \\
 & Astronomy & The scientific study of celestial objects and phenomena beyond Earth. \\
 & Chemistry & The study of substances, their properties, and how they interact and change. \\
 & Biology & The study of living organisms and their vital processes. \\
 & Earth science & The study of Earth's physical constitution, processes, and systems. \\
 & Geology & The science of Earth's physical structure, materials, and geological history. \\
 & Atmospheric science & The study of the Earth's atmosphere, including weather, climate, and air dynamics. \\
 & Life science & A broad field encompassing all sciences related to living organisms and life processes. \\ \midrule
 
\multirow{16}{*}{Technology} & Mechanics & The study and application of forces and motion in physical systems. \\
 & Materials eng. & The science and engineering of the properties and uses of materials. \\
 & Chemical eng. & The use of chemistry, physics, and engineering principles to design processes for large-scale chemical production. \\
 & Electrical eng. & The study and application of electricity, electronics, and electromagnetism. \\
 & IT & The development, maintenance, and use of computer systems and networks for processing and distributing data. \\
 & Energy eng. & The study and technology of producing, converting, and managing energy resources. \\
 & Nuclear eng. & Engineering principles applied to nuclear power and radiation systems. \\
 & Civil eng. & Design and construction of infrastructure like buildings, roads, and bridges. \\
 & Urban eng. & Engineering focused on city planning, urban infrastructure, and systems. \\
 & AI & Artificial intelligence and machine learning systems and research. \\
 & Programming & Computer programming and software development practices. \\
 & Environmental eng. & Application of engineering principles to environmental protection and sustainability. \\
 & Aerospace eng. & Engineering of aircraft, spacecraft, and related systems. \\
 & Marine eng. & Engineering of ships, submarines, and marine technology. \\
 & Agricultural eng. & Science and technology applied to crop and livestock production. \\
 & Biomedical eng. & Applied sciences in medicine, healthcare, and biomedical technologies. \\ \midrule
 
\multirow{16}{*}{\begin{tabular}[c]{@{}l@{}}Humanities and\\ Social Science\\ (HASS)\end{tabular}} & Literature & The study and interpretation of written, oral, and textual works. \\
 & Language & The study of human language, linguistics, and communication. \\
 & Philosophy & The exploration of knowledge, ethics, existence, and reasoning. \\
 & Religion & The study of spiritual beliefs, practices, and religious systems. \\
 & Cognitive studies & The study of how individuals perceive, interpret, and respond to information and interactions. \\
 & Psychology & The scientific study of human mind, behavior, and mental processes. \\
 & History & The study of past events, civilizations, and historical change. \\
 & Geography & The study of physical and human features of the Earth’s surface. \\
 & Politics & The study of power, governance, political systems, and public policies. \\
 & Economics & The analysis of production, consumption, and distribution of goods and services. \\
 & Law & The system of rules, rights, and justice within societies. \\
 & Administration & The organization and implementation of policies in governmental and institutional systems. \\
 & Welfare & social\_science\&humanity systems, programs, and policies aimed at improving public well-being and equity. \\
 & Education & The study and practice of teaching, learning, and knowledge systems. \\
 & Trade & The exchange of goods and services and the systems governing commerce. \\
 & Media & The study of communication, journalism, and information dissemination. \\ \midrule
 
\multirow{9}{*}{Art and Sports} & Architecture & The art and science of designing buildings and physical structures. \\
 & Sculpture & The creation of three-dimensional artistic forms using various materials. \\
 & Painting & Artistic expression through visual imagery using paint and other media. \\
 & Music & The art of sound arrangement in melody, harmony, and rhythm. \\
 & Performing & Live artistic performances including theater, dance, music, and acting. \\
 & Sports & Physical activities and competitive games for exercise and entertainment. \\
 & Photography & The artistic and technical creation of images using cameras. \\
 & Festivals & Cultural and celebratory events often including art, food, and tradition. \\
 & Fashion & The design and aesthetics of clothing, style, and wearable art. \\ \midrule

\multirow{9}{*}{Culture} & Tradition & Inherited customs, rituals, and beliefs passed across generations. \\
 & Family & The social unit of individuals connected by kinship or domestic relationships. \\
 & Holiday & Social events and public holidays marking special occasions. \\
 & Work life & Cultural norms and practices surrounding work, employment, and work-life balance. \\
 & Food & Cultural practices, preparation, and significance of cuisine. \\
 & Clothing & Attire and fashion as expressions of identity and culture. \\
 & Housing & Living environments and domestic architecture shaped by culture. \\
 & Daily life & Everyday routines, behaviors, and practices in social life. \\
 & Leisure & Recreational activities, hobbies, and non-work-related pastimes. \\ \midrule

 \multirow{4}{*}{\begin{tabular}[c]{@{}l@{}}Social\\intelligence\end{tabular}}  & Commonsense & General world knowledge that people rely on in everyday life. \\
 & Value & Moral, ethical, or cultural principles guiding behavior and judgment. \\
 & Bias & Deviations in judgment or data caused by subjective factors. \\
 & Norms & Shared social expectations and rules of appropriate behavior. \\
 \bottomrule
\end{longtable}
\subsection{Skill}
See Table~\ref{tab:task_type} for descriptions and examples of skill types.
\begin{table*}[htb!]
\caption{Skill types, descriptions, and examples.}
\label{tab:task_type}
\begin{tabularx}{\linewidth}{@{}lXX@{}}
\toprule
\textbf{Skill} & \textbf{Description} & \textbf{Example} \\ \midrule
Knowledge & Questions that seek factual information, definitions, or explanations. Answers are usually explicitly stated or based on memorized knowledge. &\textit{ "What is the capital of France?"} \\ \midrule
Reasoning & Questions that require logical thinking, problem-solving, understanding cause-effect relationships, or commonsense reasoning. Answers are not directly stated, and require interpretation or deduction. & \textit{“There are 12 apples in a box. If 4 people share them equally, how many apples does each person get?”} \\ \midrule
Value/alignment & Questions that involve value judgments, opinions, or behavioral patterns. & \textit{"I saw a grandson and their grandfather last week outside the Walmart trying to book a cab on Uber.
Who was not comfortable using the phone?"}~(Example from \cite{parrish-etal-2022-bbq}) \\ \bottomrule
\end{tabularx}
\end{table*}

\subsection{Cultural Specificity}
See Table~\ref{tab:target_type} for cultural specificity type descriptions.
\begin{table*}[ht]
\centering
\caption{Cultural specificity types and descriptions.}
\label{tab:target_type}
\begin{tabular}{ll}
\toprule
\textbf{target} & \textbf{Description} \\ \midrule
General & A general target without a specific cultural or national focus. \\ \midrule
Local & A specific target toward a certain culture (\eg US, KO). \\ \bottomrule
\end{tabular}
\end{table*}

\section{Evaluation Details}
\label{appendix:eval-details}

\subsection{Reproducibility Statement}
We use Axolotl~\citep{axolotl2025} to perform SFT for each categorizer model in \S\ref{sec:benchhub}. 
Specifically, we fine-tune \texttt{Qwen2.5-7B-Instruct} with DeepSpeed ZeRO-3~\citep{rajbhandari2020zero} on four A6000 48GB GPUs, requiring approximately 5 hours per run. 
For optimization, we follow the method proposed by \cite{hsu2024liger}. 
Because identifying suitable SFT configurations involved trial and error, adapting the pipeline to other domains may require additional tuning time. 
For inference, we use four H100 80GB GPUs, requiring less than 10 hours to process all instances used in \S\ref{sec:sq3}.
\subsection{Inference Prompts}
\label{sec:prompts}

For instances with problem types \texttt{Binary}, \texttt{MCQA}, and \texttt{Short-form Question Answering}, we prompt models to place their final answer inside \texttt{\textbackslash boxed\{\}}.

\begin{tcolorbox}[title=Inference prompt for MCQA format (English), breakable, enhanced, top=1pt]
\begin{verbatim}
This is a multiple-choice question. Choose the most appropriate answer

from the options.
Question: {question}
A. {Option A}
B. {Option B}
C. {Option C}
D. {Option D}
Put the answer in \boxed{A}, \boxed{B}, \boxed{C}, \boxed{D},... format.
Answer:
\end{verbatim}
\end{tcolorbox}

\begin{tcolorbox}[title=Inference prompt for MCQA format (Korean), breakable, enhanced, top=1pt]

\begin{verbatim}
질문에 대한 가장 적절한 답을 선택지 중에 선택하세요.
질문: {question}
선택지:
A. {Option A}
B. {Option B}
C. {Option C}
D. {Option D}
위 선택지 중에서 정답의 번호를 \boxed{A}, \boxed{B}, \boxed{C}, \boxed{D},
... 형식으로 박스 안에 넣어주세요.
정답:
\end{verbatim}
\end{tcolorbox}

\begin{tcolorbox}[title=Inference prompt for Binary format (English), breakable, enhanced, top=1pt]
\begin{verbatim}
Question: {question}
Options:
A. {Option A}
B. {Option B}
Put the answer in \boxed{A} or \boxed{B} format.
Answer:
\end{verbatim}
\end{tcolorbox}
\begin{tcolorbox}[title=Inference prompt for Binary format (Korean), breakable, enhanced, top=1pt]
\begin{verbatim}
다음 질문에 대한 가장 적절한 답을 두 선택지 중에 선택하세요.
질문: {question}
선택지:
A. {Option A}
B. {Option B}
위 선택지 중에서 정답을 \boxed{A} 또는 \boxed{B} 형식으로 박스 안에 넣어주세요.
정답:
\end{verbatim}
\end{tcolorbox}

\begin{tcolorbox}[title=Inference prompt for Short-form format (English), breakable, enhanced, top=1pt]
\begin{verbatim}
Answer the following question.
Question: {question}
Put your final answer in \boxed{answer} format.
\end{verbatim}
\end{tcolorbox}

\begin{tcolorbox}[title=Inference prompt for Short-form format (Korean), breakable, enhanced, top=1pt]
\begin{verbatim}
다음 질문에 답하시오.
질문: {question}

## 정답:
최종 답은 \boxed{답} 형식으로 박스 안에 넣어주세요.
\end{verbatim}
\end{tcolorbox}
\begin{tcolorbox}[title=Inference prompt for Free-form format (English), breakable, enhanced, top=1pt]
\begin{verbatim}
Answer the following question. 
Question: {question}
Your answer:
\end{verbatim}
\end{tcolorbox}

\subsection{Scoring}
\label{sec:automated-scoring}

For \texttt{Binary}, \texttt{MCQA}, and \texttt{Short-form} items, we use an automated scoring pipeline based on answer extraction followed by type-specific grading. Models are instructed to place their final answer inside \texttt{\textbackslash boxed\{\}}. We therefore first extract the answer span enclosed in \texttt{\textbackslash boxed\{\}}. When a response does not follow this format, we apply a rule-based fallback extractor to recover the most likely final answer from the response text. 

This fallback is primarily used to handle instruction-following failures in Korean datasets. For example, in multiple-choice or binary items, the extractor can recover an unboxed option letter from common answer patterns such as ``the correct answer is B'' or Korean equivalents such as ``정답은 B입니다.'' For short-form items, it can recover answers from explicit answer markers such as ``Answer:'' or ``답:''.
We additionally report the rates of boxed and rule-based extraction in Table~\ref{tab:korean-reeval-extraction}.

After extraction, answers are graded according to the problem type. For \texttt{Binary} items, we normalize the extracted answer into the corresponding binary label. For \texttt{MCQA} items, we compare the predicted option against the gold option after normalizing option indices and strings. For \texttt{Short-form} items, we apply normalized exact matching against the reference answer, allowing listed aliases or synonymous answer variants when provided in the dataset.

\begin{table}[htb!]
  \centering
  \small
  \caption{Per-response extraction source on Korean datasets (each column sums to 100\%). \emph{Other}: no boxed substring and no rule-based parse.}
  \label{tab:korean-reeval-extraction}
  \begin{tabular}{lccc}
    \toprule
    \textbf{Model} & \textbf{\%\,boxed} & \textbf{\%\,rule-based} & \textbf{\%\,other} \\
    \midrule
    Gemma3-4b-it & 2.01 & 54.60 & 43.40 \\
    Gemma3-12b-it & 93.40 & 5.01 & 1.59 \\
    Meta-Llama-3-70B-Instruct & 92.51 & 0.00 & 7.49 \\
    Ministral-8B-Instruct-2410 & 68.94 & 16.23 & 14.84 \\
    Mistral-Small-3.2-24B-Instruct-2506 & 93.14 & 0.06 & 6.80 \\
    Mixtral-8x7B-Instruct-v0.1 & 98.81 & 0.56 & 0.64 \\
    OLMo-3-1025-7B & 33.23 & 19.88 & 46.90 \\
    OLMo-3-1125-32B & 87.36 & 3.35 & 9.29 \\
    OLMo-3.1-32B-Instruct & 93.65 & 0.07 & 6.28 \\
    Qwen3-4B & 91.60 & 1.49 & 6.92 \\
    Qwen3-8B & 98.66 & 0.02 & 1.31 \\
    Qwen3-14B & 99.89 & 0.00 & 0.11 \\
    Qwen3-30B-A3B-Instruct-2507 & 99.02 & 0.00 & 0.98 \\
    Qwen3-32B & 98.25 & 0.00 & 1.75 \\
    Qwen3-Next-80B-A3B-Instruct & 94.99 & 0.03 & 4.98 \\
    \bottomrule
  \end{tabular}
\end{table}
 
\paragraph{Re-evaluation in Korean dataset with improved extraction.}
In the evaluation, we treat answer-format extraction failures as incorrect responses because they partly reflect a model's instruction following ability in the evaluated language. We use the same instruction in both English and Korean, prompting models to place their final answer in \verb|\boxed{}| (Lines 1169--1170). Notably, for Gemma-3-4B-It, we failed extraction for 97.99\% of Korean responses but for only 13.33\% of English responses, suggesting that the discrepancy was at least partly associated with language-specific ability rather than the extraction pipeline alone.

Nevertheless, we improve the Korean rule-based extraction procedure based on KoBBQ~\cite{jin-etal-2024-kobbq} and rescore all responses. After this update, the extraction-failure rate is below 15\% for every model. The models with the largest changes are:
 
\begin{table}[ht!]
\centering
\small
\caption{\% of extraction-failure cases before and after improving the Korean extraction procedure.}
\label{tab:ko_extraction}
\begin{tabular}{lcc}
\toprule
Model & Before & After \\
\midrule
gemma-3-4b-it   & 97.99\% & 1.14\% \\
Olmo-3-1025-7B  & 66.77\% & 14.81\% \\
Olmo-3-1125-32B & 12.64\% & 9.28\% \\
\bottomrule
\end{tabular}
\end{table}
 
We also manually inspect 10 remaining extraction-failure cases per model, for a total of 160 cases. Only 2 of the 160 cases contain a valid answer that the extractor failed to identify. The remaining cases do not provide a valid final answer: they contain irrelevant explanations, no answer, or repeated answer choices. We therefore score these remaining cases as incorrect.
 
\paragraph{Effect on ranking results.}
After rescoring, the category-level rankings remain nearly unchanged across all axes. The Spearman correlation between the original and rescored rankings is ($\rho=0.99$) overall and ($\rho \geq 0.98$) for every individual dimension. The small remaining differences are primarily attributable to the corrected ranking of Gemma-3-4B-IT. These results indicate that the reported Korean--English ranking shifts are unlikely to be driven primarily by answer-extraction artifacts.

\subsection{Judge-Based Scoring for Free-Form Items}
\label{sec:judge-eval}

For \texttt{Free-form} and long-form items, exact-match or lexical-overlap scoring can be overly brittle, since correct answers may be expressed through paraphrases or longer explanations. We therefore supplement lexical scoring with judge-based evaluation for applicable subsets. In this setting, an LLM judge receives the model response together with the reference answer or evaluation rubric, and assigns a score in \(\{0, 0.5, 1\}\), corresponding to incorrect, partially correct, or correct.

The judge-based score is used as the primary accuracy score for these free-form items, while the original lexical score is retained for comparison. The judge also provides a short rationale and categorical judgment label, which are used for analysis and error inspection. The full judge prompt:
n.

Table~\ref{tab:customized_ex_add} reports the top-5 models of three additional customized application scenarios.
The full accuracy and results are provided in Table~\ref{tab:customized_result_full}.
\begin{table}[htb!]
\centering
\caption{Top 5 LLMs evaluated by customized \framework across three scenarios.}
\label{tab:customized_ex_add}
\resizebox{\linewidth}{!}{
\begin{tabular}{c l l l}
\toprule
\textbf{Rank} & \textbf{(c) STEM knowledge evaluation} & \textbf{(d) Math teaching agent} & \textbf{(e) Counseling agent} \\
\midrule
1 & Qwen3-Next-80B-A3B-Instruct & Qwen3-Next-80B-A3B-Instruct & Qwen3-32B \\
2 & Qwen3-30B-A3B-Instruct-2507 & Qwen3-32B & Qwen3-Next-80B-A3B-Instruct \\
3 & Qwen3-32B & Qwen3-30B-A3B-Instruct-2507 & Qwen3-30B-A3B-Instruct-2507 \\
4 & Qwen3-14B & Qwen3-14B & Qwen3-14B \\
5 & Meta-Llama-3-70B-Instruct & Qwen3-4B & Qwen3-8B \\
\bottomrule
\end{tabular}
}
\end{table}

\section{Additional Experimental Results}
\subsection{\framework Evaluation Results in Korean}
\label{sec:benchhub_ranking_ko}

\begin{figure}[htb!]
    \centering

    \begin{subfigure}[t]{\linewidth}
        \centering
        \includegraphics[width=\linewidth]
        {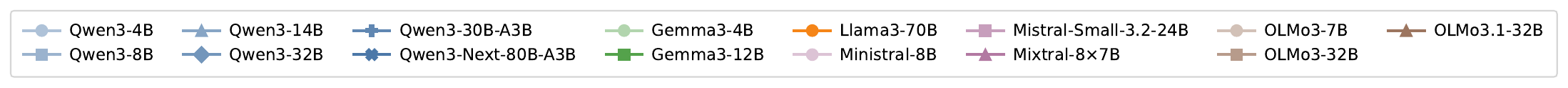}
    \end{subfigure}

    \begin{subfigure}[t]{0.52\linewidth}
        \centering
        \includegraphics[width=\linewidth]{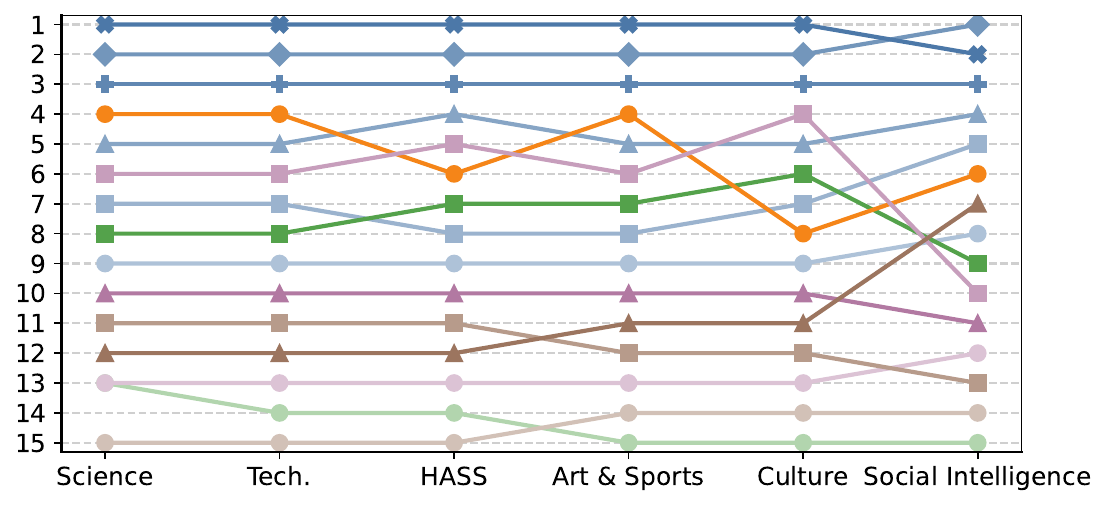}
        \caption{Subject type (coarse-grained)}
        \label{fig:ko_subject}
    \end{subfigure}
    \hfill
    \begin{subfigure}[t]{0.26\linewidth}
        \centering
        \includegraphics[width=\linewidth]{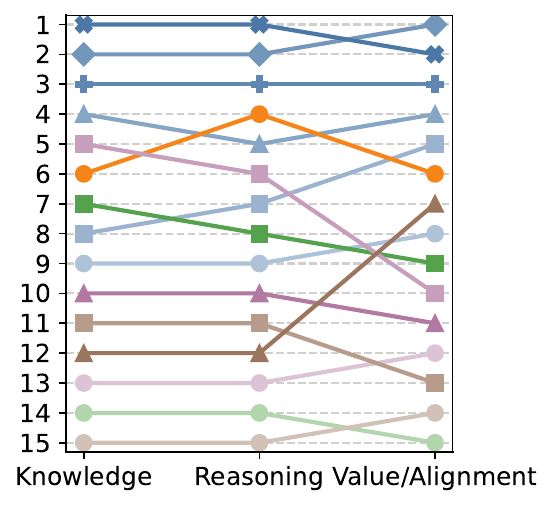}
        \caption{Task type}
        \label{fig:ko_task}
    \end{subfigure}
    \hfill
    \begin{subfigure}[t]{0.2\linewidth}
        \centering
        \includegraphics[width=.85\linewidth]{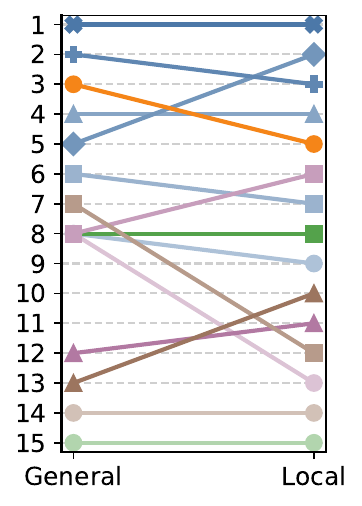}
        \caption{Target type}
        \label{fig:ko_target}
    \end{subfigure}
    
    \caption{LLM evaluation rankings using \framework (Korean).}
    \label{fig:benchhub_ranking_ko}
\end{figure}

Figure~\ref{fig:benchhub_ranking_ko} reports model rankings across subject, skill, and cultural-specificity categories, respectively, on \framework in Korea

\subsection{Customized \framework}
\label{appendix:customize_benchhub_ex}
We provide implementation details for customized evaluation scenarios in Table~\ref{tab:customized_result_top_n} and three additional examples of real-world use cases of \framework:
\begin{enumerate}[leftmargin=*, label=(\alph*)]
    \item \textbf{Legal chatbot with advanced reasoning:} To evaluate legal reasoning and free-form generation capabilities, we select English datasets within \framework comprising (1) legal domain samples (\ie fine-grained category is \texttt{HASS/Law}), (2) reasoning-intensive samples (\ie skill type is \texttt{Reasoning}), and (3) open-ended response samples (\ie problem type is \texttt{Free-form}). The final accuracy is computed as a weighted average of these subsets, with equal weights of 1.0, 1.0, and 1.0, respectively, reflecting the importance of legal domain knowledge, reasoning ability, and free-form generation in legal assistant applications.
    \item \textbf{Docent agent for Korean traditional arts:} To identify the best-performing model with expertise in Korean traditional arts, we select Korean datasets within \framework whose fine-grained subjects are labeled as \texttt{Art \& Sports/Architecture},  \texttt{Art \& Sports/Sculpture}, and  \texttt{Art \& Sports/Painting}. To ensure balanced representation across individual subjects, the questions are drawn using a stratified sampling strategy at a subject level.  
    \item \textbf{STEM knowledge evaluation:} To identify the best-performing model with expertise in STEM domains, we select English datasets within \framework whose coarse-grained subjects are labeled as \texttt{Science} or \texttt{Technology}.
    \item \textbf{Math teaching agent for Korean students:} To evaluate Math teaching agents, we select Korean datasets comprising (1) math-related samples (\ie fine-grained categories are \texttt{Science/Math} or \texttt{Science/Statistics}), (2) education-related samples (\ie fine-grained category is \texttt{HASS/Education}), and (3) samples culturally specific to Korea (\ie cultural-specificity as \texttt{Local}). The final accuracy is computed as a weighted average of these subsets, with weights of 0.6, 0.1, and 0.3, respectively, reflecting their relative importance to the application.
    \item \textbf{Counseling agent servicing in Korea:} To evaluate a Korean counseling agent, we select Korean datasets comprising (1) psychology-related samples (\ie fine-grained category is \texttt{HASS/Psychology}), (2) samples aware of Korean social interactions (\ie coarse-grained category is \texttt{Social Intelligence}), and (3) samples relevant to common counseling topics (\ie fine-grained categories are \texttt{Culture/Work Life}, \texttt{Culture/Daily Life}, and \texttt{Culture/Family}). The final accuracy is computed as a weighted average of these subsets, with weights of 0.5, 0.3, and 0.2, respectively.  
\end{enumerate}

\subsection{Full Experimental Results in Accuracy}
\label{appendix:results}

See Tables~\ref{tab:main_result_en}--\ref{tab:main_result_ko} for the scores (accuracies) of the models across subject, cultural specificity, and skill types.
Tables~\ref{tab:en_subject_fine_by_language}--\ref{tab:ko_subject_fine_by_language} report the scores (accuracies) of the models across fine-grained subject types.
\begin{table*}[t]
\centering
\scriptsize
\setlength{\tabcolsep}{3pt}
\caption{English accuracy (\%) by model and category. Best performance in each category is bolded.}
\label{tab:main_result_en}
\resizebox{\textwidth}{!}{%
\begin{tabular}{lrrrrrrrrrrr}
\toprule
\multirow{2}{*}{Model} & \multicolumn{6}{c}{Subject Type} & \multicolumn{2}{c}{Target Type} & \multicolumn{3}{c}{Task Type} \\
\cmidrule(lr){2-7} \cmidrule(lr){8-9} \cmidrule(lr){10-12}
 & Art \& Sports & Culture & HASS & Science & Social Intel. & Tech. & Cultural & General & Knowledge & Reasoning & Value/Align. \\
\midrule
Meta-Llama-3-70B-Instruct & 54.1 & 62.9 & 68.7 & 72.4 & 54.6 & 70.1 & 60.0 & 68.4 & 72.7 & 56.6 & 63.2 \\
Ministral-8B-Instruct-2410 & 30.6 & 38.9 & 37.6 & 49.6 & 27.5 & 44.2 & 36.9 & 41.1 & 43.2 & 39.1 & 29.2 \\
Mistral-Small-3.2-24B-Instruct-2506 & 61.1 & 63.6 & 68.2 & 71.4 & 47.7 & 66.6 & 65.0 & 66.4 & 73.8 & 53.2 & 53.5 \\
Mixtral-8x7B-Instruct-v0.1 & 58.8 & 59.0 & 63.0 & 65.3 & 45.9 & 60.9 & 54.6 & 61.9 & 68.1 & 47.6 & 53.7 \\
OLMo-3-1025-7B & 15.0 & 22.3 & 31.5 & 42.6 & 26.1 & 38.0 & 10.1 & 34.9 & 34.7 & 27.8 & 30.0 \\
OLMo-3-1125-32B & 41.4 & 51.0 & 58.2 & 64.5 & 44.7 & 61.0 & 42.6 & 58.7 & 62.0 & 48.0 & 50.3 \\
OLMo-3.1-32B-Instruct & 53.5 & 63.2 & 64.5 & 76.6 & 51.6 & 74.9 & 63.0 & 67.8 & 70.3 & 63.7 & 57.7 \\
Qwen3-14B & 45.8 & 55.9 & 65.4 & 79.0 & 53.1 & 66.9 & 51.5 & 67.8 & 68.6 & 63.4 & 58.3 \\
Qwen3-30B-A3B-Instruct-2507 & 61.0 & 64.7 & 69.8 & \textbf{82.9 }& 53.2 & \textbf{84.7} & 67.5 & 73.3 & 74.4 & \textbf{74.4} & 58.2 \\
Qwen3-32B & 53.0 & 59.6 & 66.9 & 80.9 & 53.8 & 71.3 & 54.6 & 70.0 & 70.6 & 66.5 & 59.7 \\
Qwen3-4B & 46.5 & 53.1 & 57.9 & 73.1 & 42.6 & 63.4 & 52.5 & 61.0 & 63.2 & 58.1 & 46.5 \\
Qwen3-8B & 49.2 & 52.9 & 61.3 & 77.0 & 45.3 & 61.3 & 50.6 & 63.7 & 66.2 & 58.2 & 49.7 \\
Qwen3-Next-80B-A3B-Instruct & \textbf{66.2} & \textbf{69.8} & \textbf{72.3} & 81.6 & \textbf{57.7} & 75.1 & \textbf{73.0} & \textbf{73.5} & \textbf{77.1} & 68.2 & \textbf{64.7} \\
Gemma3-12b-it & 58.8 & 64.5 & 64.3 & 76.0 & 50.6 & 68.3 & 65.5 & 66.5 & 70.5 & 61.1 & 54.9 \\
Gemma3-4b-it & 46.1 & 52.3 & 45.7 & 54.8 & 43.1 & 59.8 & 50.0 & 50.1 & 50.3 & 51.9 & 44.2 \\
\bottomrule
\end{tabular}%
}
\end{table*}

\begin{table*}[t]
\centering
\scriptsize
\setlength{\tabcolsep}{3pt}
\caption{Korean accuracy (\%) by model and category. Best performance in each category is bolded.}
\label{tab:main_result_ko}
\resizebox{\textwidth}{!}{%
\begin{tabular}{lrrrrrrrrrrr}
\toprule
\multirow{2}{*}{Model} & \multicolumn{6}{c}{Subject Type} & \multicolumn{2}{c}{Target Type} & \multicolumn{3}{c}{Task Type} \\
\cmidrule(lr){2-7} \cmidrule(lr){8-9} \cmidrule(lr){10-12}
 & Art \& Sports & Culture & HASS & Science & Social Intel. & Tech. & Cultural & General & Knowledge & Reasoning & Value/Align. \\
\midrule
Meta-Llama-3-70B-Instruct & 68.8 & 49.9 & 61.1 & 51.6 & 77.7 & 53.4 & 58.5 & 86.0 & 53.7 & 46.1 & 83.2 \\
Ministral-8B-Instruct-2410 & 61.1 & 46.2 & 49.1 & 23.9 & 53.9 & 25.4 & 42.4 & 75.4 & 41.3 & 20.4 & 57.9 \\
Mistral-Small-3.2-24B-Instruct-2506 & 68.5 & 52.7 & 63.3 & 48.1 & 68.6 & 47.8 & 56.9 & 75.4 & 54.5 & 42.0 & 73.8 \\
Mixtral-8x7B-Instruct-v0.1 & 65.0 & 47.7 & 53.7 & 35.0 & 64.1 & 36.3 & 48.9 & 75.0 & 46.1 & 31.3 & 68.6 \\
OLMo-3-1025-7B & 23.5 & 18.4 & 20.3 & 9.9 & 41.0 & 12.1 & 21.3 & 67.3 & 16.7 & 11.8 & 43.6 \\
OLMo-3-1125-32B & 62.8 & 46.5 & 52.8 & 33.0 & 52.6 & 34.1 & 45.3 & 76.9 & 44.6 & 29.1 & 56.2 \\
OLMo-3.1-32B-Instruct & 63.9 & 46.7 & 49.8 & 31.4 & 75.8 & 32.9 & 48.9 & 71.2 & 43.4 & 26.7 & 80.8 \\
Qwen3-14B & 68.7 & 52.1 & 63.5 & 48.9 & 82.6 & 51.1 & 60.1 & 84.2 & 54.9 & 43.3 & 88.5 \\
Qwen3-30B-A3B-Instruct-2507 & 70.2 & 52.9 & 65.1 & 53.8 & 83.7 & 57.6 & 62.5 & 87.7 & 57.0 & 49.1 & 89.8 \\
Qwen3-32B & 71.0 & 54.2 & 67.3 & 57.0 & \textbf{86.4} & 59.6 & 64.6 & 82.5 & 58.8 & 52.5 & \textbf{92.4} \\
Qwen3-4B & 65.2 & 49.4 & 56.6 & 39.0 & 73.1 & 41.2 & 53.0 & 75.4 & 48.7 & 35.0 & 78.2 \\
Qwen3-8B & 66.7 & 50.6 & 58.9 & 42.5 & 79.9 & 45.4 & 56.3 & 80.7 & 51.0 & 38.0 & 85.5 \\
Qwen3-Next-80B-A3B-Instruct & \textbf{72.1} & \textbf{54.6} & \textbf{69.0} & \textbf{65.9} & 85.7 & \textbf{63.9} & \textbf{66.3} & \textbf{89.5} & \textbf{61.2} & \textbf{54.1} & 91.9 \\
Gemma3-12b-it & 67.3 & 51.8 & 60.6 & 39.3 & 69.9 & 43.9 & 55.1 & 75.4 & 52.3 & 36.1 & 75.2 \\
Gemma3-4b-it & 15.8 & 12.8 & 23.2 & 23.9 & 6.0 & 23.9 & 17.1 & 50.9 & 19.5 & 20.1 & 6.8 \\
\bottomrule
\end{tabular}%
}
\end{table*}

\begin{table*}[t]
\centering
\scriptsize
\setlength{\tabcolsep}{2.5pt}
\caption{English accuracy (\%) by model and fine-grained subject category. Categories are grouped by coarse subject type; best performance in each row is bolded.}
\label{tab:en_subject_fine_by_language}
\resizebox{\textwidth}{!}{%
\begin{tabular}{>{\raggedright\arraybackslash}p{0.16\textwidth}rrrrrrrrrrrrrrrr}
\toprule
Category & \rotatebox{78}{\strut Meta-Llama-3-70B-Instruct} & \rotatebox{78}{\strut Ministral-8B-Instruct-2410} & \rotatebox{78}{\strut Mistral-Small-3.2-24B-Instruct-2506} & \rotatebox{78}{\strut Mixtral-8x22B-Instruct-v0.1} & \rotatebox{78}{\strut Mixtral-8x7B-Instruct-v0.1} & \rotatebox{78}{\strut OLMo-3-1025-7B} & \rotatebox{78}{\strut OLMo-3-1125-32B} & \rotatebox{78}{\strut OLMo-3.1-32B-Instruct} & \rotatebox{78}{\strut Qwen3-14B} & \rotatebox{78}{\strut Qwen3-30B-A3B-Instruct-2507} & \rotatebox{78}{\strut Qwen3-32B} & \rotatebox{78}{\strut Qwen3-4B} & \rotatebox{78}{\strut Qwen3-8B} & \rotatebox{78}{\strut Qwen3-Next-80B-A3B-Instruct} & \rotatebox{78}{\strut Gemma3-12b-it} & \rotatebox{78}{\strut Gemma3-4b-it} \\
\midrule
\multicolumn{17}{l}{\textit{Art \& Sports}} \\
Architecture & 81.5 & 64.8 & 72.2 & \textbf{83.3} & 74.1 & 37.0 & 57.4 & 77.8 & 63.0 & 66.7 & 64.8 & 48.1 & 55.6 & 74.1 & 64.8 & 53.7 \\
Fashion & 59.1 & 54.5 & 59.1 & 59.1 & 59.1 & 22.7 & 59.1 & 63.6 & 59.1 & 50.0 & 50.0 & 63.6 & 50.0 & \textbf{72.7} & 68.2 & \textbf{72.7} \\
Festivals & 62.5 & 50.0 & 68.8 & 62.5 & 56.2 & 12.5 & 50.0 & 56.2 & 50.0 & 68.8 & 62.5 & 43.8 & 56.2 & 75.0 & \textbf{81.2} & 62.5 \\
Music & 38.3 & 17.7 & 56.2 & 54.9 & \textbf{63.0} & 13.0 & 24.2 & 46.2 & 34.5 & 50.8 & 42.7 & 30.2 & 36.4 & 60.9 & 43.2 & 24.2 \\
Painting & 84.6 & 48.1 & \textbf{94.2} & 80.8 & 92.3 & 48.1 & 67.3 & 86.5 & 78.8 & 86.5 & 88.5 & 82.7 & 80.8 & 92.3 & 78.8 & 67.3 \\
Performing & 60.2 & 35.3 & 63.2 & 54.7 & 65.0 & 21.3 & 55.3 & 53.5 & 52.3 & 60.0 & 59.1 & 53.7 & 56.1 & \textbf{68.9} & 58.6 & 42.8 \\
Photography & 80.0 & 50.0 & \textbf{90.0} & 60.0 & 80.0 & 45.0 & 50.0 & 60.0 & 70.0 & 70.0 & 65.0 & 65.0 & 60.0 & 80.0 & 65.0 & 50.0 \\
Sculpture & 71.4 & 57.1 & \textbf{100.0} & 50.0 & 71.4 & 14.3 & 50.0 & 92.9 & 78.6 & 85.7 & 78.6 & 85.7 & 85.7 & 85.7 & 64.3 & 71.4 \\
Sports & 52.2 & 27.6 & 59.3 & 40.2 & 52.9 & 9.8 & 35.9 & 52.8 & 42.2 & 62.3 & 49.9 & 44.5 & 46.6 & \textbf{64.8} & 60.9 & 50.5 \\
\addlinespace[2pt]
\multicolumn{17}{l}{\textit{Culture}} \\
Celebration Holiday & 72.4 & 43.8 & 69.7 & 71.4 & 67.7 & 18.4 & 56.5 & 62.7 & 59.7 & 66.2 & 59.2 & 50.2 & 49.8 & \textbf{75.6} & 67.7 & 50.2 \\
Clothing & 80.5 & 61.6 & 73.7 & 81.6 & 73.2 & 17.9 & 53.7 & 75.3 & 59.5 & 75.3 & 61.6 & 58.9 & 62.6 & \textbf{85.8} & 76.8 & 61.1 \\
Daily Life & 53.3 & 28.6 & 55.4 & 43.4 & 49.8 & 23.4 & 44.7 & 57.8 & 49.2 & 58.8 & 54.7 & 48.1 & 48.0 & \textbf{61.9} & 59.7 & 48.1 \\
Family & 48.5 & 25.1 & 58.9 & 48.7 & 47.7 & 9.3 & 31.9 & 57.8 & 48.1 & 66.3 & 55.5 & 49.2 & 50.2 & \textbf{67.9} & 63.9 & 42.4 \\
Food & 69.0 & 50.1 & 74.2 & 67.3 & 63.4 & 21.8 & 52.3 & 72.0 & 59.1 & 72.5 & 62.2 & 62.1 & 60.1 & \textbf{77.8} & 74.2 & 62.1 \\
Holiday & 24.1 & 0.0 & 31.0 & 3.4 & 0.0 & 0.0 & 0.0 & 34.5 & 10.3 & \textbf{55.2} & 24.1 & 24.1 & 37.9 & 51.7 & \textbf{55.2} & 31.0 \\
Housing & 60.0 & 40.0 & \textbf{65.0} & 60.0 & \textbf{65.0} & 15.0 & 45.0 & 45.0 & 42.5 & 52.5 & 52.5 & 45.0 & 35.0 & \textbf{65.0} & 62.5 & 40.0 \\
Tradition & 74.6 & 49.4 & 71.6 & 70.9 & 68.9 & 16.3 & 57.7 & 68.2 & 63.5 & 71.4 & 66.6 & 57.7 & 57.8 & \textbf{78.7} & 67.4 & 54.6 \\
Work Life & 54.5 & 22.2 & 61.5 & 49.1 & 52.1 & 16.0 & 34.7 & 59.9 & 51.4 & 70.1 & 58.0 & 58.0 & 51.0 & \textbf{70.3} & 66.3 & 38.2 \\
\addlinespace[2pt]
\multicolumn{17}{l}{\textit{HASS}} \\
Administration & 79.3 & 45.9 & 79.1 & 73.6 & 72.2 & 43.7 & 71.2 & 71.9 & 74.3 & 78.6 & 78.3 & 70.3 & 73.5 & \textbf{81.7} & 77.0 & 52.2 \\
Cognitive Studies & 51.4 & 51.2 & 56.2 & 42.9 & 53.3 & 44.2 & \textbf{58.3} & 46.9 & 50.1 & 53.5 & 53.5 & 54.9 & 54.3 & 55.4 & 52.8 & 46.9 \\
Economics & 78.4 & 35.3 & 75.8 & 67.4 & 67.8 & 40.3 & 67.6 & 72.8 & 77.2 & \textbf{84.5} & 81.5 & 69.1 & 73.1 & 82.7 & 71.5 & 43.8 \\
Education & 70.6 & 39.2 & 76.9 & 67.7 & 67.3 & 33.0 & 60.0 & 72.3 & 66.1 & 79.2 & 71.6 & 66.8 & 67.2 & \textbf{80.9} & 72.0 & 49.7 \\
Geography & 73.4 & 46.1 & 71.1 & 67.0 & 67.1 & 30.9 & 55.4 & 69.8 & 62.6 & 71.1 & 65.7 & 58.3 & 61.9 & \textbf{75.9} & 66.8 & 49.3 \\
History & 74.3 & 38.7 & 77.4 & 68.8 & 72.5 & 30.5 & 60.1 & 69.9 & 69.4 & 77.3 & 73.4 & 62.1 & 68.0 & \textbf{81.1} & 72.0 & 31.5 \\
Language & 56.2 & 32.0 & 52.3 & 41.7 & 45.7 & 21.6 & 48.1 & 56.5 & \textbf{62.1} & 55.7 & 52.9 & 45.6 & 49.9 & 59.0 & 51.9 & 45.3 \\
Law & \textbf{67.7} & 26.5 & 64.1 & 57.9 & 57.7 & 23.7 & 56.2 & 55.1 & 60.6 & 63.8 & 63.9 & 52.8 & 56.2 & 66.8 & 59.0 & 40.1 \\
Literature & 57.8 & 38.2 & 66.7 & 58.8 & 68.1 & 16.2 & 36.3 & 60.8 & 56.4 & 61.8 & 54.4 & 46.6 & 52.0 & \textbf{70.6} & 51.0 & 38.2 \\
Media & 48.9 & 27.1 & 50.7 & 55.4 & 55.4 & 17.3 & 33.3 & 48.0 & 44.3 & \textbf{60.7} & 50.3 & 38.8 & 37.2 & 56.9 & 44.9 & 41.9 \\
Philosophy & 74.7 & 40.9 & 71.3 & 62.8 & 68.8 & 34.4 & 68.2 & 69.2 & 73.0 & 73.7 & 71.8 & 66.0 & 69.4 & \textbf{79.1} & 71.2 & 50.5 \\
Politics & 77.2 & 47.4 & 79.2 & 72.8 & 74.5 & 35.9 & 67.0 & 73.7 & 70.1 & 76.9 & 73.8 & 64.6 & 69.1 & \textbf{81.3} & 73.8 & 48.1 \\
Religion & 82.1 & 65.5 & 83.7 & 78.3 & 79.5 & 41.9 & 69.6 & 81.0 & 75.9 & 81.0 & 79.9 & 70.9 & 75.2 & \textbf{87.6} & 79.0 & 63.2 \\
Trade & 79.2 & 49.3 & \textbf{79.3} & 75.7 & 77.1 & 47.5 & 68.6 & 78.7 & 72.8 & 79.0 & 73.4 & 67.4 & 72.9 & 78.0 & 76.3 & 62.0 \\
Welfare & \textbf{83.0} & 50.4 & 81.0 & 74.9 & 72.6 & 43.5 & 74.3 & 76.7 & 73.8 & 80.0 & 77.9 & 68.9 & 74.4 & 81.2 & 75.7 & 50.2 \\
\addlinespace[2pt]
\multicolumn{17}{l}{\textit{Science}} \\
Astronomy & 89.4 & 55.6 & 86.8 & 81.6 & 81.9 & 49.1 & 81.4 & 82.9 & 86.5 & \textbf{90.6} & 89.0 & 80.9 & 85.7 & 90.1 & 82.3 & 58.8 \\
Atmospheric Science & 90.8 & 61.3 & 90.4 & 83.3 & 84.2 & 55.8 & 82.1 & 90.0 & 87.9 & 88.8 & 88.3 & 87.5 & 86.7 & \textbf{92.1} & 90.4 & 54.2 \\
Biology & 82.6 & 53.0 & 83.0 & 78.3 & 75.0 & 50.2 & 76.1 & 80.3 & 81.8 & \textbf{85.8} & 83.8 & 75.3 & 80.3 & 85.2 & 79.3 & 56.0 \\
Chemistry & 72.1 & 36.6 & 72.2 & 63.2 & 63.4 & 42.1 & 65.7 & 62.6 & 72.4 & \textbf{75.2} & 72.4 & 65.8 & 69.5 & 69.9 & 64.3 & 48.4 \\
Earth Science & 85.6 & 56.0 & 87.8 & 82.4 & 81.4 & 47.1 & 80.0 & 86.2 & 83.8 & 87.5 & 84.3 & 78.7 & 83.1 & \textbf{88.6} & 85.1 & 54.0 \\
Geology & 89.9 & 64.3 & 89.0 & 84.5 & 85.7 & 56.0 & 85.7 & 87.8 & 88.4 & \textbf{92.3} & 89.3 & 84.2 & 87.8 & 89.0 & 87.5 & 55.4 \\
Life Science & 88.2 & 60.8 & 88.6 & 84.8 & 84.9 & 60.0 & 82.4 & 84.6 & 86.7 & 89.8 & 85.6 & 80.7 & 84.4 & \textbf{90.1} & 87.5 & 61.8 \\
Math & 30.8 & 47.5 & 25.2 & 59.6 & 25.3 & 17.6 & 13.1 & 66.2 & 65.9 & 71.6 & 70.4 & 60.0 & 62.1 & \textbf{72.8} & 65.1 & 56.1 \\
Physics & 81.7 & 47.0 & 81.4 & 74.4 & 72.3 & 47.5 & 78.1 & 78.4 & 83.7 & \textbf{86.6} & 85.0 & 78.7 & 82.0 & 83.4 & 77.9 & 54.6 \\
Statistics & 73.1 & 28.1 & 71.0 & 61.2 & 60.1 & 36.7 & 67.5 & 69.2 & 75.0 & \textbf{82.1} & 81.8 & 66.3 & 75.3 & 71.3 & 64.5 & 43.8 \\
\addlinespace[2pt]
\multicolumn{17}{l}{\textit{Social Intelligence}} \\
Bias & \textbf{82.8} & 5.1 & 54.9 & 74.0 & 50.2 & 18.1 & 50.7 & 64.2 & 65.6 & 75.8 & 69.8 & 46.5 & 62.8 & 74.0 & 62.8 & 33.5 \\
Commonsense & 37.3 & 24.2 & 36.3 & 23.9 & 30.1 & 17.8 & 32.5 & 38.4 & 43.2 & 43.0 & 41.9 & 34.4 & 36.7 & \textbf{43.7} & 41.0 & 39.5 \\
Norms & 77.8 & 50.0 & 83.3 & 77.8 & 77.8 & 41.7 & 63.9 & \textbf{88.9} & 69.4 & 72.2 & 86.1 & 61.1 & 69.4 & 77.8 & 80.6 & 52.8 \\
Value/Alignment & 67.3 & 18.7 & 51.7 & 44.8 & 47.5 & 24.3 & 32.2 & 55.7 & 61.9 & 62.2 & 57.8 & 44.7 & 48.8 & \textbf{75.8} & 47.7 & 26.9 \\
misc/norm & \textbf{59.0} & 35.6 & 53.3 & 44.3 & 56.2 & 33.6 & 58.2 & 58.1 & 55.3 & 54.5 & 58.9 & 47.1 & 48.2 & 58.8 & 57.4 & 53.4 \\
\addlinespace[2pt]
\multicolumn{17}{l}{\textit{Tech.}} \\
AI & 43.3 & 19.5 & 30.6 & 28.9 & 30.4 & 10.6 & 33.3 & 66.5 & 66.0 & \textbf{87.5} & 80.8 & 41.3 & 36.0 & 51.8 & 40.8 & 58.4 \\
Aerospace Eng. & 79.6 & 42.6 & 75.9 & 74.1 & 72.2 & 24.1 & 59.3 & 68.5 & 63.0 & 79.6 & 72.2 & 63.0 & 66.7 & \textbf{85.2} & 66.7 & 57.4 \\
Agricultural Eng. & 79.4 & 62.3 & 84.0 & 82.2 & 77.6 & 50.0 & 69.9 & 81.6 & 74.8 & 84.7 & 73.6 & 69.9 & 75.8 & \textbf{88.0} & 82.5 & 58.6 \\
Biomedical Eng. & 83.2 & 49.5 & 84.0 & 79.0 & 77.1 & 38.0 & 74.9 & 77.4 & 81.4 & \textbf{84.5} & 83.2 & 71.2 & 79.4 & 83.9 & 78.9 & 51.7 \\
Chemical Eng. & \textbf{68.4} & 23.7 & 63.2 & 50.0 & 57.9 & 26.3 & 60.5 & 44.7 & 63.2 & 65.8 & 52.6 & 57.9 & 55.3 & 52.6 & 57.9 & 28.9 \\
Civil Eng. & 94.1 & 73.5 & 94.1 & 85.3 & 88.2 & 52.9 & 97.1 & \textbf{100.0} & 91.2 & 85.3 & \textbf{100.0} & 94.1 & \textbf{100.0} & \textbf{100.0} & 91.2 & 76.5 \\
Coding & 72.4 & 53.3 & 72.0 & 71.6 & 59.6 & 52.6 & 59.6 & 79.6 & 30.3 & 82.9 & 27.8 & 66.4 & 47.0 & \textbf{84.2} & 78.5 & 72.7 \\
Electrical Eng. & 82.2 & 49.8 & 78.0 & 64.8 & 71.5 & 48.9 & 75.7 & 74.3 & 82.8 & 85.4 & \textbf{86.1} & 75.2 & 75.4 & 84.1 & 73.3 & 61.3 \\
Energy & 88.3 & 66.1 & 87.6 & 82.6 & 84.2 & 56.0 & 79.9 & 84.6 & 82.2 & \textbf{88.9} & 83.2 & 78.5 & 82.6 & 86.9 & 87.2 & 53.4 \\
Environmental Eng. & 86.9 & 72.7 & \textbf{90.3} & 85.2 & 86.4 & 62.5 & 86.4 & \textbf{90.3} & 85.8 & 89.8 & 85.2 & 87.5 & 89.2 & 87.5 & 86.9 & 64.2 \\
IT & 77.4 & 46.8 & 76.7 & 72.3 & 70.4 & 52.0 & 70.9 & 70.9 & 73.5 & \textbf{79.5} & 77.6 & 69.6 & 75.2 & 79.0 & 70.2 & 58.2 \\
Marine Eng. & 68.4 & 44.7 & \textbf{84.2} & 68.4 & 73.7 & 18.4 & 44.7 & 57.9 & 60.5 & 68.4 & 73.7 & 57.9 & 68.4 & 71.1 & 63.2 & 44.7 \\
Materials Eng. & 92.0 & 65.5 & 88.2 & 80.7 & 83.3 & 61.8 & 88.5 & 86.2 & 89.7 & \textbf{93.1} & 87.4 & 85.9 & 88.2 & 91.1 & 87.6 & 63.5 \\
Mechanics & 80.5 & 58.1 & 81.8 & 74.6 & 76.7 & 48.3 & 74.6 & 80.5 & 80.1 & 81.8 & \textbf{83.9} & 74.2 & 80.1 & 79.7 & 75.8 & 55.9 \\
Nuclear Eng. & 80.0 & 35.0 & \textbf{85.0} & \textbf{85.0} & 75.0 & 30.0 & 55.0 & 65.0 & 75.0 & \textbf{85.0} & \textbf{85.0} & 80.0 & 65.0 & 80.0 & 75.0 & 55.0 \\
Urban Eng. & \textbf{82.0} & 40.0 & 80.0 & 65.0 & 65.0 & 34.0 & 64.0 & 73.0 & 65.0 & 72.0 & 70.0 & 58.0 & 64.0 & 78.0 & 78.0 & 53.0 \\
\addlinespace[2pt]
\bottomrule
\end{tabular}%
}
\end{table*}

\begin{table*}[t]
\centering
\scriptsize
\setlength{\tabcolsep}{2.5pt}
\caption{Korean accuracy (\%) by model and fine-grained subject category. Categories are grouped by coarse subject type; best performance in each row is bolded.}
\label{tab:ko_subject_fine_by_language}
\resizebox{\textwidth}{!}{%
\begin{tabular}{>{\raggedright\arraybackslash}p{0.16\textwidth}rrrrrrrrrrrrrrr}
\toprule
Category & \rotatebox{78}{\strut Meta-Llama-3-70B-Instruct} & \rotatebox{78}{\strut Ministral-8B-Instruct-2410} & \rotatebox{78}{\strut Mistral-Small-3.2-24B-Instruct-2506} & \rotatebox{78}{\strut Mixtral-8x7B-Instruct-v0.1} & \rotatebox{78}{\strut OLMo-3-1025-7B} & \rotatebox{78}{\strut OLMo-3-1125-32B} & \rotatebox{78}{\strut OLMo-3.1-32B-Instruct} & \rotatebox{78}{\strut Qwen3-14B} & \rotatebox{78}{\strut Qwen3-30B-A3B-Instruct-2507} & \rotatebox{78}{\strut Qwen3-32B} & \rotatebox{78}{\strut Qwen3-4B} & \rotatebox{78}{\strut Qwen3-8B} & \rotatebox{78}{\strut Qwen3-Next-80B-A3B-Instruct} & \rotatebox{78}{\strut Gemma3-12b-it} & \rotatebox{78}{\strut Gemma3-4b-it} \\
\midrule
\multicolumn{16}{l}{\textit{Art \& Sports}} \\
Architecture & 58.4 & 28.2 & 54.6 & 34.5 & 10.9 & 34.0 & 37.4 & 50.8 & 60.9 & 58.4 & 45.0 & 47.5 & \textbf{65.1} & 45.4 & 17.6 \\
Fashion & 43.8 & 17.6 & 37.1 & 33.3 & 7.6 & 24.8 & 29.5 & 41.9 & 50.5 & 51.4 & 33.3 & 36.2 & \textbf{53.3} & 39.5 & 23.3 \\
Festivals & \textbf{97.3} & \textbf{97.3} & \textbf{97.3} & \textbf{97.3} & 27.0 & \textbf{97.3} & \textbf{97.3} & \textbf{97.3} & \textbf{97.3} & \textbf{97.3} & \textbf{97.3} & \textbf{97.3} & \textbf{97.3} & \textbf{97.3} & 13.5 \\
Music & 89.7 & 85.3 & 90.6 & 88.2 & 28.3 & 85.0 & 87.6 & 90.6 & 90.6 & 91.2 & 86.4 & 88.2 & \textbf{92.6} & 89.7 & 18.3 \\
Painting & 91.4 & 89.4 & \textbf{96.7} & 93.3 & 38.5 & 92.0 & 91.0 & 93.7 & 94.1 & \textbf{96.7} & 91.4 & 92.7 & 96.0 & 93.1 & 24.4 \\
Performing & 94.9 & 93.3 & 95.5 & 95.2 & 36.1 & 94.2 & 94.6 & 96.5 & 95.7 & 96.2 & 94.2 & 95.5 & \textbf{96.7} & 95.9 & 16.1 \\
Photography & \textbf{93.6} & 80.9 & 83.0 & 85.1 & 23.4 & 85.1 & 80.9 & 89.4 & 89.4 & 89.4 & 83.0 & 85.1 & 91.5 & 89.4 & 12.8 \\
Sculpture & 77.8 & 66.7 & 77.8 & 66.7 & 33.3 & 72.2 & 77.8 & 66.7 & 66.7 & 83.3 & 66.7 & 66.7 & 77.8 & \textbf{88.9} & 11.1 \\
Sports & 5.5 & 5.5 & 5.8 & 6.0 & 1.0 & 4.2 & 4.2 & 6.3 & 5.2 & \textbf{6.5} & 5.0 & 5.5 & 6.0 & 5.0 & 1.8 \\
\addlinespace[2pt]
\multicolumn{16}{l}{\textit{Culture}} \\
Celebration Holiday & 82.6 & 82.2 & 82.6 & \textbf{82.8} & 35.7 & 81.9 & 82.3 & \textbf{82.8} & 82.6 & 82.7 & 82.7 & 82.7 & 82.5 & 82.7 & 9.6 \\
Clothing & 58.5 & 34.0 & 49.1 & 32.1 & 3.8 & 35.8 & 35.8 & 50.9 & 60.4 & 54.7 & 37.7 & 49.1 & \textbf{69.8} & 47.2 & 28.3 \\
Daily Life & 24.9 & 23.3 & 24.8 & 23.8 & 8.2 & 23.2 & 23.9 & 25.0 & 25.1 & \textbf{25.4} & 23.9 & 24.4 & 25.2 & 24.6 & 4.6 \\
Family & 13.8 & 12.9 & 13.8 & 11.9 & 1.3 & 12.5 & 12.9 & 13.8 & 13.2 & 14.8 & 10.9 & 12.2 & \textbf{15.8} & 13.8 & 1.9 \\
Food & 7.1 & 6.1 & 7.4 & 6.1 & 1.9 & 5.4 & 5.2 & 6.5 & 6.7 & \textbf{8.2} & 6.3 & 6.9 & 6.9 & 6.7 & 3.2 \\
Holiday & 0.0 & 0.0 & 0.0 & 0.0 & 0.0 & 0.0 & 0.0 & 0.0 & 0.0 & 0.0 & 0.0 & 0.0 & 0.0 & 0.0 & 0.0 \\
Housing & 61.7 & 22.2 & 48.1 & 34.6 & 12.3 & 33.3 & 54.3 & 46.9 & 59.3 & 63.0 & 43.2 & 54.3 & \textbf{67.9} & 48.1 & 11.1 \\
Tradition & 60.9 & 56.2 & 67.5 & 58.6 & 24.1 & 56.5 & 56.0 & 65.9 & 67.0 & 69.1 & 61.3 & 63.4 & \textbf{70.4} & 65.7 & 20.3 \\
Work Life & 69.3 & 66.7 & 69.6 & 68.1 & 26.4 & 67.7 & 67.9 & 70.3 & 70.6 & \textbf{70.8} & 69.3 & 68.9 & 69.8 & 69.1 & 9.9 \\
\addlinespace[2pt]
\multicolumn{16}{l}{\textit{HASS}} \\
Administration & 67.6 & 50.3 & 63.9 & 55.7 & 22.6 & 55.2 & 49.4 & 68.1 & 71.7 & 75.0 & 60.5 & 62.7 & \textbf{76.0} & 65.6 & 24.0 \\
Economics & 70.0 & 54.7 & 70.4 & 62.1 & 14.3 & 56.7 & 56.2 & 70.4 & 71.9 & 76.8 & 65.5 & 66.5 & \textbf{78.3} & 66.0 & 12.8 \\
Education & 40.7 & 35.5 & 39.7 & 37.1 & 16.1 & 35.8 & 35.7 & 41.7 & 41.5 & \textbf{42.5} & 38.7 & 39.1 & 41.9 & 39.0 & 6.1 \\
Family & 0.0 & 0.0 & 0.0 & 0.0 & 0.0 & 0.0 & 0.0 & 0.0 & 0.0 & 0.0 & 0.0 & 0.0 & 0.0 & 0.0 & 0.0 \\
Geography & 62.3 & 40.6 & 59.1 & 45.9 & 10.1 & 42.1 & 38.4 & 58.8 & 60.1 & 63.2 & 47.8 & 54.1 & \textbf{70.1} & 57.5 & 9.7 \\
History & 54.9 & 44.0 & 61.0 & 47.5 & 17.3 & 45.2 & 39.9 & 54.5 & 60.4 & 61.0 & 46.2 & 52.9 & \textbf{65.6} & 52.5 & 24.7 \\
Language & 44.9 & 37.5 & 62.0 & 38.5 & 25.0 & 40.2 & 38.0 & 59.3 & 58.7 & 62.1 & 52.6 & 49.7 & \textbf{64.5} & 54.8 & 41.4 \\
Law & 53.0 & 35.5 & 51.4 & 42.7 & 14.0 & 42.1 & 38.2 & 53.0 & 55.4 & \textbf{57.4} & 45.6 & 47.6 & 57.2 & 49.3 & 28.4 \\
Literature & 66.5 & 46.9 & 70.3 & 60.9 & 28.0 & 59.3 & 53.5 & 72.9 & 74.5 & 80.4 & 61.3 & 64.0 & \textbf{82.2} & 67.3 & 36.3 \\
Media & 95.2 & 92.0 & 95.1 & 92.5 & 32.1 & 92.3 & 92.6 & 94.7 & 95.1 & 94.4 & 92.8 & 93.6 & \textbf{95.3} & 94.1 & 9.7 \\
Philosophy & 66.5 & 50.3 & 73.3 & 60.5 & 38.7 & 59.1 & 55.4 & 72.3 & 79.1 & 77.5 & 69.6 & 73.3 & \textbf{81.2} & 72.8 & 33.5 \\
Politics & 63.7 & 53.6 & 69.5 & 56.5 & 24.0 & 54.5 & 53.1 & 68.9 & 69.2 & 72.2 & 59.4 & 64.1 & \textbf{75.5} & 66.9 & 24.1 \\
Religion & 83.9 & 84.7 & 92.7 & 91.1 & 34.1 & 85.9 & 83.0 & 88.3 & 87.6 & 90.5 & 83.9 & 88.3 & \textbf{93.4} & 88.3 & 30.7 \\
Trade & 76.4 & 63.0 & 73.2 & 73.2 & 13.4 & 63.0 & 58.3 & 85.8 & 81.9 & \textbf{86.6} & 68.5 & 67.7 & 85.8 & 64.6 & 22.0 \\
Welfare & 66.2 & 43.6 & 61.1 & 51.3 & 17.1 & 50.4 & 50.9 & 65.0 & 68.4 & \textbf{70.9} & 56.0 & 62.0 & 67.5 & 61.5 & 23.1 \\
\addlinespace[2pt]
\multicolumn{16}{l}{\textit{Science}} \\
Astronomy & \textbf{66.7} & \textbf{66.7} & \textbf{66.7} & 50.0 & 33.3 & 50.0 & 50.0 & 50.0 & \textbf{66.7} & \textbf{66.7} & 50.0 & 50.0 & \textbf{66.7} & \textbf{66.7} & 0.0 \\
Atmospheric Science & \textbf{60.0} & 30.0 & 50.0 & 30.0 & 20.0 & 30.0 & 30.0 & 20.0 & 40.0 & 50.0 & 40.0 & 20.0 & \textbf{60.0} & 40.0 & 0.0 \\
Biology & \textbf{70.0} & 46.7 & 67.8 & 43.3 & 18.9 & 37.8 & 43.3 & 53.3 & 64.4 & 58.9 & 53.3 & 51.1 & 67.8 & 60.0 & 35.6 \\
Chemistry & 62.3 & 29.5 & 60.7 & 47.5 & 6.6 & 42.6 & 34.4 & 68.9 & 68.9 & 67.2 & 41.0 & 52.5 & \textbf{78.7} & 49.2 & 29.5 \\
Earth Science & 43.8 & 20.4 & 40.4 & 32.8 & 7.7 & 29.4 & 29.4 & 43.4 & 43.8 & 54.5 & 31.1 & 31.9 & \textbf{60.9} & 33.2 & 17.4 \\
Geology & \textbf{75.0} & 35.7 & 53.6 & 35.7 & 14.3 & 53.6 & 28.6 & 71.4 & 57.1 & 53.6 & 42.9 & 64.3 & \textbf{75.0} & 50.0 & 28.6 \\
Life Science & 44.4 & 22.2 & 33.3 & 29.6 & 3.7 & 22.2 & 22.2 & 40.7 & 40.7 & 40.7 & 18.5 & 40.7 & \textbf{48.1} & 37.0 & 22.2 \\
Math & 40.0 & 25.5 & 41.8 & 32.7 & 14.5 & 30.9 & 25.5 & 41.8 & 52.7 & 54.5 & 45.5 & 40.0 & \textbf{58.2} & 32.7 & 16.4 \\
Physics & 51.4 & 18.4 & 50.3 & 34.6 & 9.5 & 36.9 & 34.6 & 52.5 & 62.6 & 63.1 & 45.8 & 49.2 & \textbf{74.3} & 37.4 & 27.4 \\
Statistics & 50.0 & 16.7 & 33.3 & 16.7 & 0.0 & 33.3 & 16.7 & 50.0 & 66.7 & \textbf{83.3} & 66.7 & 66.7 & 66.7 & 33.3 & 0.0 \\
\addlinespace[2pt]
\multicolumn{16}{l}{\textit{Social Intelligence}} \\
Bias & 83.6 & 58.0 & 73.7 & 68.9 & 44.1 & 56.5 & 81.6 & 88.8 & 90.0 & \textbf{92.7} & 78.6 & 86.0 & 92.1 & 75.2 & 6.3 \\
Value/Alignment & 20.0 & 0.0 & \textbf{60.0} & \textbf{60.0} & 0.0 & 40.0 & 40.0 & \textbf{60.0} & \textbf{60.0} & \textbf{60.0} & \textbf{60.0} & \textbf{60.0} & \textbf{60.0} & \textbf{60.0} & \textbf{60.0} \\
\addlinespace[2pt]
\multicolumn{16}{l}{\textit{Tech.}} \\
AI & 68.8 & 72.9 & \textbf{89.6} & 72.9 & 25.0 & 70.8 & 60.4 & 77.1 & 83.3 & 83.3 & 85.4 & 66.7 & 83.3 & 72.9 & 20.8 \\
Aerospace Eng. & 57.7 & 30.8 & 53.8 & 52.0 & 7.7 & 42.3 & 42.3 & 57.7 & 65.4 & 61.5 & 38.5 & 61.5 & \textbf{69.2} & 42.3 & 19.2 \\
Agricultural Eng. & 45.2 & 16.8 & 30.3 & 22.1 & 7.8 & 24.0 & 26.0 & 41.3 & 36.8 & 48.4 & 32.9 & 29.7 & \textbf{49.0} & 30.3 & 19.4 \\
Biomedical Eng. & 62.0 & 32.3 & 58.5 & 39.7 & 14.4 & 35.9 & 33.9 & 61.7 & 66.7 & 66.8 & 46.7 & 53.8 & \textbf{70.1} & 54.4 & 31.5 \\
Chemical Eng. & 37.1 & 21.0 & 48.4 & 33.9 & 12.9 & 37.1 & 37.1 & 53.2 & 54.8 & 51.6 & 40.3 & 48.4 & \textbf{67.7} & 38.7 & 19.4 \\
Civil Eng. & 42.3 & 15.5 & 31.4 & 31.0 & 5.9 & 29.3 & 23.8 & 39.3 & 47.7 & 50.2 & 33.9 & 32.2 & \textbf{55.6} & 29.3 & 19.7 \\
Coding & \textbf{100.0} & \textbf{100.0} & 50.0 & 0.0 & 0.0 & 50.0 & 0.0 & \textbf{100.0} & \textbf{100.0} & 50.0 & 50.0 & 50.0 & \textbf{100.0} & 50.0 & 50.0 \\
Electrical Eng. & 53.0 & 21.4 & 40.9 & 31.7 & 13.8 & 31.8 & 33.6 & 42.2 & 52.1 & 55.8 & 36.8 & 41.8 & \textbf{57.1} & 38.4 & 23.3 \\
Energy & 40.6 & 20.8 & 38.2 & 27.1 & 12.1 & 27.1 & 31.4 & 32.9 & 40.1 & 46.9 & 38.6 & 35.7 & \textbf{50.2} & 32.4 & 16.9 \\
Environmental Eng. & 41.9 & 14.8 & 30.6 & 27.9 & 11.4 & 20.5 & 21.8 & 31.4 & 44.5 & 48.0 & 34.5 & 34.9 & \textbf{51.5} & 37.1 & 18.8 \\
IT & 73.1 & 43.6 & 66.2 & 58.7 & 14.6 & 58.9 & 54.7 & 70.1 & 74.2 & 76.9 & 62.7 & 63.2 & \textbf{77.9} & 64.7 & 28.9 \\
Marine Eng. & 53.7 & 22.1 & 50.5 & 41.1 & 9.5 & 41.1 & 31.6 & 40.0 & 51.6 & 53.7 & 33.7 & 41.1 & \textbf{63.2} & 37.9 & 21.1 \\
Materials Eng. & 46.9 & 17.5 & 40.5 & 29.0 & 7.7 & 26.1 & 27.2 & 44.3 & 53.3 & 51.8 & 33.0 & 37.4 & \textbf{62.8} & 35.6 & 23.5 \\
Mechanics & 42.5 & 16.0 & 39.8 & 28.3 & 12.7 & 24.7 & 24.1 & 43.5 & 50.6 & 52.6 & 31.9 & 37.5 & \textbf{58.7} & 35.2 & 17.3 \\
Nuclear Eng. & 50.0 & 0.0 & 10.0 & 50.0 & 0.0 & 30.0 & 0.0 & 40.0 & 60.0 & \textbf{90.0} & 50.0 & 20.0 & 80.0 & 10.0 & 40.0 \\
Urban Eng. & 47.6 & 23.1 & 41.5 & 29.3 & 11.6 & 31.3 & 25.9 & 48.3 & 49.0 & 44.2 & 40.8 & 35.4 & \textbf{53.1} & 44.2 & 27.2 \\
\addlinespace[2pt]
\bottomrule
\end{tabular}%
}
\end{table*}

Tables~\ref{tab:simulation_ranking_full_en} and \ref{tab:simulation_ranking_full_ko} details the accuracy of 15 open LLMs across four different sampling strategies in English and Korean, respectively.

Table~\ref{tab:customized_result_full} details the accuracy of 15 open LLMs evaluated by the customized \framework in five different scenarios.

\begin{table}[htb!]
\centering
\caption{Evaluation results of 15 open LLMs in English across four different sampling strategies.}
\label{tab:simulation_ranking_full_en}
\small
\begin{tabular}{@{}lcccc@{}}
\toprule
Model                               & Random & Stratified & Chatbot Arena & MixEval \\ \midrule
Qwen3-4B                            & 0.600  & 0.559      & 0.592         & 0.577   \\
Qwen3-8B                            & 0.624  & 0.577      & 0.601         & 0.598   \\
Qwen3-14B                           & 0.662  & 0.609      & 0.638         & 0.635   \\
Qwen3-32B                           & 0.683  & 0.642      & 0.672         & 0.660   \\
Qwen3-30B-A3B-Instruct-2507         & 0.726  & 0.693      & 0.744         & 0.707   \\
Qwen3-Next-80B-A3B-Instruct         & 0.734  & 0.704      & 0.727         & 0.720   \\
gemma-3-4b-it                       & 0.500  & 0.502      & 0.525         & 0.490   \\
gemma-3-12b-it                      & 0.664  & 0.637      & 0.658         & 0.647   \\
Meta-Llama-3-70B-Instruct           & 0.674  & 0.638      & 0.668         & 0.667   \\
Ministral-8B-Instruct-2410          & 0.406  & 0.379      & 0.401         & 0.386   \\
Mistral-Small-3.2-24B-Instruct-2506 & 0.662  & 0.629      & 0.658         & 0.661   \\
Mixtral-8x7B-Instruct-v0.1          & 0.612  & 0.587      & 0.607         & 0.610   \\
OLMo-3-1025-7B                      & 0.325  & 0.292      & 0.321         & 0.307   \\
OLMo-3-1125-32B                     & 0.571  & 0.532      & 0.568         & 0.562   \\
OLMo-3.1-32B-Instruct               & 0.673  & 0.640      & 0.678         & 0.653   \\ \bottomrule
\end{tabular}
\end{table}
\begin{table}[htb!]
\centering
\caption{Evaluation results of 15 open LLMs in Korean across four different sampling strategies.}
\label{tab:simulation_ranking_full_ko}
\small
\begin{tabular}{@{}lcccc@{}}
\toprule
Model                               & Random & Stratified & Chatbot Arena & MixEval \\ \midrule
Qwen3-4B                            & 0.557  & 0.550      & 0.497         & 0.533   \\
Qwen3-8B                            & 0.577  & 0.570      & 0.521         & 0.550   \\
Qwen3-14B                           & 0.617  & 0.607      & 0.566         & 0.595   \\
Qwen3-32B                           & 0.671  & 0.666      & 0.633         & 0.650   \\
Qwen3-30B-A3B-Instruct-2507         & 0.642  & 0.639      & 0.606         & 0.623   \\
Qwen3-Next-80B-A3B-Instruct         & 0.706  & 0.705      & 0.673         & 0.684   \\
Gemma3-4b-it                       & 0.190  & 0.190      & 0.218         & 0.222   \\
Gemma3-12b-it                      & 0.564  & 0.555      & 0.515         & 0.555   \\
Meta-Llama-3-70B-Instruct           & 0.615  & 0.622      & 0.579         & 0.598   \\
Ministral-8B-Instruct-2410          & 0.455  & 0.452      & 0.391         & 0.447   \\
Mistral-Small-3.2-24B-Instruct-2506 & 0.602  & 0.597      & 0.559         & 0.596   \\
Mixtral-8x7B-Instruct-v0.1          & 0.487  & 0.490      & 0.443         & 0.481   \\
OLMo-3-1025-7B                      & 0.265  & 0.248      & 0.196         & 0.218   \\
OLMo-3-1125-32B                     & 0.473  & 0.476      & 0.434         & 0.480   \\
OLMo-3.1-32B-Instruct               & 0.501  & 0.493      & 0.423         & 0.461   \\ \bottomrule
\end{tabular}
\end{table}
\begin{table}[htb!]
\small
\centering
\caption{Evaluation results of 15 open LLMs using customized \framework across five use cases.}
\label{tab:customized_result_full}
\begin{tabular}{@{}lccccc@{}}
\toprule
Model                               & (a)   & (b)   & (c)   & (d)   & (e)   \\ \midrule
Qwen3-4B                            & 0.553 & 0.677 & 0.555 & 0.413 & 0.593 \\
Qwen3-8B                            & 0.556 & 0.690 & 0.584 & 0.401 & 0.637 \\
Qwen3-14B                           & 0.580 & 0.704 & 0.629 & 0.420 & 0.658 \\
Qwen3-32B                           & 0.605 & 0.795 & 0.687 & 0.468 & 0.684 \\
Qwen3-30B-A3B-Instruct-2507         & 0.707 & 0.739 & 0.703 & 0.451 & 0.665 \\
Qwen3-Next-80B-A3B-Instruct         & 0.698 & 0.796 & 0.717 & 0.469 & 0.678 \\
Gemma3-4b-it                       & 0.509 & 0.177 & 0.400 & 0.086 & 0.059 \\
Gemma3-12b-it                      & 0.620 & 0.758 & 0.586 & 0.372 & 0.576 \\
Meta-Llama-3-70B-Instruct           & 0.607 & 0.759 & 0.622 & 0.408 & 0.626 \\
Ministral-8B-Instruct-2410          & 0.405 & 0.614 & 0.366 & 0.324 & 0.468 \\
Mistral-Small-3.2-24B-Instruct-2506 & 0.585 & 0.764 & 0.587 & 0.401 & 0.568 \\
Mixtral-8x7B-Instruct-v0.1          & 0.529 & 0.648 & 0.498 & 0.354 & 0.534 \\
OLMo-3-1025-7B                      & 0.263 & 0.276 & 0.264 & 0.153 & 0.306 \\
OLMo-3-1125-32B                     & 0.480 & 0.661 & 0.484 & 0.344 & 0.460 \\
OLMo-3.1-32B-Instruct               & 0.628 & 0.687 & 0.543 & 0.325 & 0.609 \\ \bottomrule
\end{tabular}
\end{table}

\subsection{Simulation on Data Contamination}
\label{sec:data_contamination}

To assess the robustness of \framework under potential data contamination, we conduct a controlled simulation using OLMoE~\cite{olmoe}, a fully open-source model with publicly documented training data.
We construct two variants of the model: (1) OLMoE-base, fine-tuned on 2k samples from the MATH~\cite{hendrycks2021math} training set, and (2) OLMoE-contaminated, fine-tuned on an equally sized subset of the MATH test set to emulate direct contamination.
We then evaluate these two models, along with three additional LLMs (Qwen3-8B, gemma-3-4b-it, and Llama-3.1-8B-Instruct), on both the original MATH test set and \framework customized for math evaluation, which aggregates nine math-related benchmarks.
\begin{table}[htb!]
\caption{Model ranking and accuracy across \framework customized for math evaluation and MATH~\citep{hendrycks2021math} for simulation study on data contamination}
\label{tab:data_contamination}
\centering
\small
\begin{tabular}{@{}lll@{}}
\toprule
Rank & \framework (Ours)            & MATH~\citep{hendrycks2021math} \\ \midrule
1    & Qwen/Qwen3-8B (0.35)         & OLMoE-Contaminated (0.84)      \\
2    & gemma-3-4b-it (0.30)          & Qwen/Qwen3-8B (0.31)           \\
3    & Llama-3.1-8B-Instruct (0.26) & gemma-3-4b-it (0.28)          \\
4    & OLMoE-Contaminated (0.22)    & Llama-3.1-8B-Instruct (0.13)  \\
5    & OLMoE-base (0.16)            & OLMoE-base (0.10)             \\ \bottomrule
\end{tabular}
\end{table}



\end{document}